\documentclass[12pt]{article}
\usepackage{apacite}
\usepackage{amsmath}
\usepackage{times}
\usepackage{graphicx}
\usepackage{color}
\usepackage{multirow}
\usepackage[authoryear]{natbib}
\usepackage{rotating}
\usepackage{bbm}
\usepackage{latexsym}
\usepackage{amsmath}
\usepackage{latexsym}
\usepackage{graphicx}
\usepackage{caption}
\usepackage{subfig}
\usepackage{multirow}
\usepackage{comment}
\usepackage{graphicx}
\usepackage{amsmath}
\usepackage{latexsym}
\usepackage{graphicx}
\usepackage{caption}
\usepackage{subfig}
\usepackage{multirow}
\usepackage{comment}
\usepackage{color}
\usepackage{amssymb}
\usepackage{tikz}
\usepackage{bm}
\DeclareMathOperator*{\argmax}{argmax}
\DeclareMathOperator*{\argmin}{argmin}
\usepackage{times}
\usepackage[ruled,linesnumbered]{algorithm2e}
\usepackage[utf8]{inputenc}
\usepackage[english]{babel}
\usepackage{lineno}
\usepackage[utf8]{inputenc}
\usepackage[english]{babel}
 
\usepackage[utf8]{inputenc}
\usepackage[english]{babel}
 
\usepackage{amsthm}
 \usepackage{url}
\usepackage{ marvosym }
 \usepackage{natbib}
\usepackage{lipsum}

\usepackage[hang,flushmargin]{footmisc}

\usepackage[utf8]{inputenc}
\usepackage[english]{babel}
 
\newtheorem{theorem}{Theorem}
\newtheorem{corollary}{Corollary}[theorem]
\newtheorem{lemma}{Lemma}
\usepackage{amsthm,amssymb}

\usepackage{amsthm}
 \usepackage{algorithm2e,setspace}

\usepackage{setspace}

\usepackage{thmtools}
\declaretheoremstyle[headfont=\normalfont]{normalhead}
\declaretheorem[style=normalhead]{definition}
\newtheorem{assumption}{Assumption}

\newtheorem{remark}{Remark}
\newcommand{\captionfonts}{\normalsize}

\makeatletter  
\long\def\@makecaption#1#2{%
  \vskip\abovecaptionskip
  \sbox\@tempboxa{{\captionfonts #1: #2}}%
  \ifdim \wd\@tempboxa >\hsize
    {\captionfonts #1: #2\par}
  \else
    \hbox to\hsize{\hfil\box\@tempboxa\hfil}%
  \fi
  \vskip\belowcaptionskip}
\makeatother   

\begin{document}
\hspace{13.9cm}1

\ \vspace{20mm}\\

{\LARGE \bf Distribution Matching for  Machine Teaching \\}
 \begin{center}
{ \large  Xiaofeng Cao and  Ivor W.\ Tsang }\\
{$^{\displaystyle  }$Australian Artificial Intelligence Institute, University
of Technology  Sydney, Australia.\\ 
Email: xiaofeng.cao@uts.edu.au, ivor.tsang@uts.edu.au}\\
\end{center}
%


\thispagestyle{empty}
\markboth{}{NC instructions}
\ \vspace{-0mm}\\
You may need to know the following questions before reading our manuscript. 

\textbf{Q1: What is machine teaching?} \\
 Machine teaching is an inverse problem of machine learning that aims at 
 steering the student learner towards its target hypothesis, in which the teacher has already
known the student's learning parameters. In simple terms, machine teaching can help to find the best training data for student learners automatically.

\textbf{Q2: What is our research problem?} \\
Previous machine teaching studies on machine teaching focused on balancing the teaching risk  and cost to find those best teaching examples deriving the student model. This optimization solver is in general ineffective when the student learner   does not disclose any cue (i.e. a black-box) of  the learning parameters. 
To supervise such a teaching scenario,  this paper 
presents a  distribution matching-based machine teaching strategy.

\textbf{Q3: Why it is significant?} \\
Our study will help machine learning to find the best teaching examples on a black-box setting, then the training examples will be closed-form. In this paper, we present a distribution matching perspective to resolve this issue from optimization.

\textbf{Q4: What is our novelty?} \\
Technically, our strategy can be expressed as a  cost-controlled  optimization process    that
 finds the optimal teaching examples without further exploring in  the parameter distribution of the student learner.

\textbf{Q5: What is the related study?} \\
The  most relevant machine learning study to machine teaching is active learning. 
  Its key assumption is that a student learner who frequently interacts with a teacher (annotator)  would do better or no worse than other passive student learners who  randomly solicit the training examples. Generally, active learning forwardly updates the current training model  into its target, while the target  is always agnostic.

 \newpage
 {\LARGE \bf Distribution Matching for  Machine Teaching \\}
  \begin{center}
{ \large  Xiaofeng Cao and  Ivor W.\ Tsang }\\
{$^{\displaystyle  }$Australian Artificial Intelligence Institute, University
of Technology  Sydney, Australia. \\ 
Email: xiaofeng.cao@uts.edu.au, ivor.tsang@uts.edu.au}\\
\end{center}

\begin{abstract}
 Machine teaching is an inverse problem of machine learning that aims at 
 steering the student learner towards its target hypothesis, in which the teacher has already
known the student's learning parameters.
Previous studies on machine teaching focused on balancing the teaching risk  and cost to find those best teaching examples deriving the student model. This optimization solver is in general ineffective when the student learner   does not disclose any cue of  the learning parameters. 
To supervise such a teaching scenario,  this paper 
presents a  distribution matching-based machine teaching strategy. Specifically,   this strategy backwardly and iteratively performs the halving operation on the teaching cost to find a desired teaching set. Technically, our strategy can be expressed as a  cost-controlled  optimization process    that
 finds the optimal teaching examples without further exploring in  the parameter distribution of the student learner. Then, given any a limited teaching cost, the training examples will be  closed-form.  Theoretical analysis and experiment results demonstrate this strategy.  
\end{abstract}

\section{Introduction}
{Machine learning \citep{mitchell1997machine}  is the study of artificial intelligent algorithms that improve automatically through model construction and parameter experience. Highly informative or representative training examples accelerate the convergence of the learning model. However, how to control the machine learning paradigm if there is a teacher who has already known  the learning parameters of the student and wants better training examples  
to improve its generalization? This inverse  question of machine learning was studied by machine teaching \citep{zhu2018overview}, which  explores the optimal training data via driving the student
learner to its target hypothesis.  It  has been shown many promising paradigms ranging from a teaching scheme to a student learner  \citep{DBLP:conf/icml/LiuDLLRS18} such as a curriculum to human education system in curriculum learning \citep{DBLP:journals/corr/MatiisenOCS17},  an iterative query algorithm to an 
annotator in active learning \citep{DBLP:conf/icml/Dasgupta0PZ19}, etc. The key assumption is that the teacher knows the target  parameters of the student model e.g. a specified hyperplane in  SVM classifier,  geometric properties  of  clustering centers,  etc. Typically, a 
machine teacher  interacts with  its student learner by exploring those  teaching examples that 
 minimize the \emph{parameter  disagreement} (difference) of the current training model and its desired \citep{gao2017preference} \citep{shinohara1991teachability}. }

\begin{figure}[!t]
\centering
\includegraphics[scale=0.32]{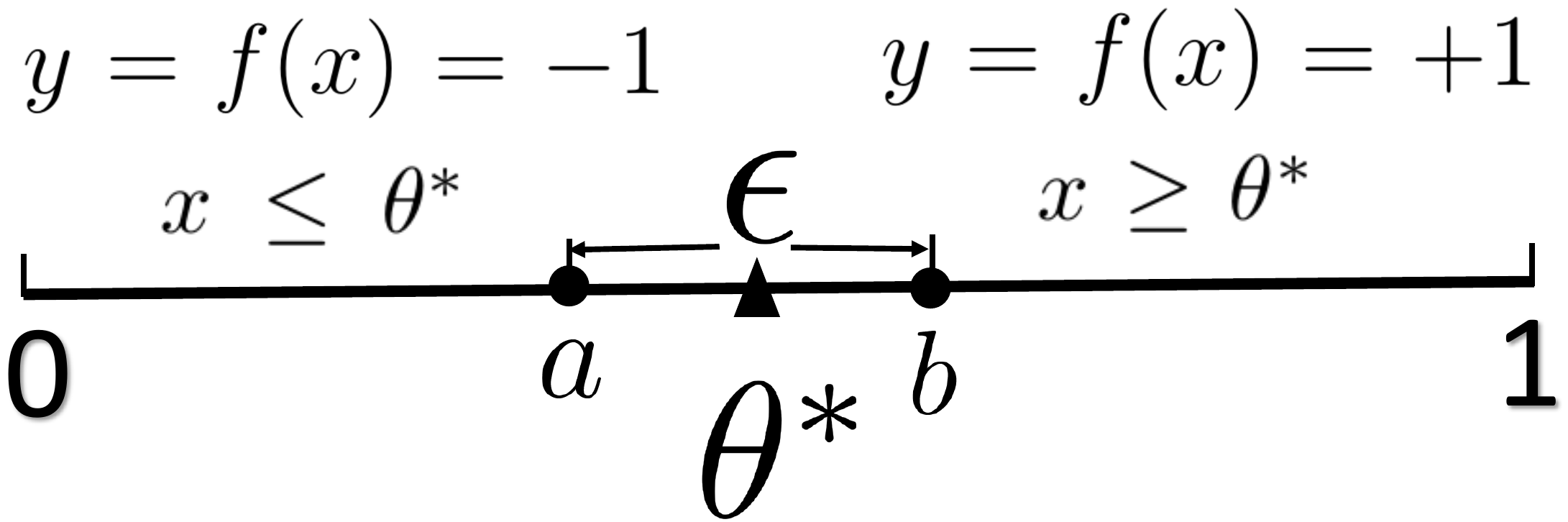}
\caption{{Label complexity of a student learner. Given an error $\epsilon$,  the passive learner  needs  $\mathcal{O}(1/\epsilon)$ data,  and the active learner  needs  $\mathcal{O}({\rm log}{1/\epsilon})$;  the machine teacher only needs two symmetrical
examples $a$ and $b$ (black points) around $\theta^*$ to minimize the parameter disagreements: $\|\hat \theta-\theta^* \|$, where the target parameter $\hat \theta$ is denoted by $\hat \theta= \frac{a+b}{2}$, where $\|\cdot \|$ denotes an effective metric over the parameter space. } }
\end{figure}

In simple terms, machine teaching can help to find the best training data for student learners automatically. With machine teaching, those   teaching examples can be  closed-form to supervise the training model of a student learner.  As far as we know, the  most relevant machine learning study to machine teaching is active learning \citep{dasgupta2008general}. 
  Its key assumption is that a student learner who frequently interacts with a teacher (annotator)  would do better or no worse than other passive student learners who  randomly solicit the training examples. Generally, active learning forwardly updates the current training model  into its target, while the target  is always agnostic. We next employ a threshold  classifier  to further explain   active learning and machine teaching following the survey of  \citep{zhu2018overview}.
  
  \par {Considering an interval [-1,1] with positive and negative labels over a uniform distribution 
   $\mathop{P}$ (see Figure~1), the parameterized threshold classifier at $\theta^*=0$ stipulates the classification function $f$ for any $(x,y)\sim \mathop{P}$: $y=f(x)=+1, {\rm s.t.}\ x\geq \theta^*$, $y=f(x)=-1, {\rm s.t.} \ x\leq \theta^*$, where $y$ denotes the label of $x$. Drawing i.i.d. $n$ samples from  $\mathop{P}$, a passive (random) learner will make $n$ times of querying from $f(x)$  that yields a generalization error of $\|\theta-\theta^*\|=\mathcal{O}(1/n)$ due to an average uniform spacing $1/n$, where $\theta$ yields a passive learner. Given an error $\epsilon$, 
  a passive learner  needs  $\mathcal{O}(1/\epsilon)$ data sent to the classifier. Specifically, the passive learner at least receives 10,000 samples from $\mathop{P}$ to obtain a desired error 0.0001. For an active learner who usually employs binary search, the learner halves the remaining interval and removes the data over it, thereby receives  around  $\mathcal{O}({\rm log}{1/\epsilon})$ samples to obtain an error  $\epsilon$ because  $\|\theta-\theta^*\|=\mathcal{O}(1/2^n)$.  Specifically, the active  learner at least receives 13 samples from $\mathop{P}$ to obtain a desired error 0.0001. However, a machine teacher who  knows $\theta^*$ only needs two teaching examples to obtain $\epsilon$: $(\theta^*-\epsilon/2,-1)$, and $(\theta^*+\epsilon/2,+1)$. Then, for any student learner, they will easily achieve better performance using those teaching examples.}

 From a machine learning perspective, Eq.~(1) firstly presents a general strategy on active learning: optimizing a model parameter  $\theta$ with current training data $D_0 \subset D$, subsequently updated by queries $\hat D_q$:
\begin{equation}
\begin{split}
 &\min_{\theta\in \Theta} R(\theta, D_0\cup \hat D_q)+\Omega(\theta), \\
 &{\rm s.t.}\   \hat D_q\in \argmax_{D_q\subset  D\backslash D_0} \Big|R(\theta, D_0)-R(\theta, D_0\cup D_q)\Big|,\\
 \end{split}
\end{equation}
where $R(\cdot, \cdot)$ denotes the empirical risk function,  $\Omega(\cdot)$ denotes the regularization constraint, $D$ denotes the full training data,  $D_q$ denotes the queries from $D\backslash D_0$ using active learning which maximizes the risk disagreement, and $\Theta$ denotes the parameter  space. With the minimizer of the regulated learning risk, model parameter $\theta$ derives the update on $D_0$ via  merging $\hat D_q$.

Given a teacher who has already known  the learning parameters, machine teaching inversely  optimizes the training data by estimating parameter disagreements over the generalization models.
\cite{zhu2015machine} proposed a more general machine teaching formula based on Eq.~(1),
\begin{equation}
\begin{split}
 &\min_{D\subset\mathbb{D}}\mathcal{L}(\hat\theta, \theta^*)+\eta \mathcal{M}(D),\\
&\text{s.t.} \   \hat\theta\in \text{arg min}_{\theta\in \Theta} R(\theta,D)+\Omega(\theta), \\
\end{split}
\end{equation}
where $\mathbb{D}$ denotes the complete or whole search space of $D$ that covers all candidate training subsets,  $\mathcal{L}(\cdot,\cdot)$ w.r.t. parameter disagreement denotes the teaching risk, $\mathcal{M}(\cdot)$ denotes the teaching cost of the input set, and $\eta$ denotes a balance coefficient. Typically, $\mathcal{L}(\hat\theta, \theta^*)$ can be simply defined as the indication disagreement $\mathbb{E}_{x\sim P }\mathbbm{1}_{(\hat\theta(x)\neq \theta^*(x))}$, where $\mathbbm{1}$ denotes the indication function, and $\hat \theta$ denotes the desired parameter. Teaching cost can be simply defined as the cardinality of $D$:  $\mathcal{M}(D)=\|D\|_0$, in which $\mathcal{M}(\mathbb{D})=2^{\|D\|_0}$ ($\|\cdot\|_0$ denotes $L_0$ norm)\footnote{$2^{\|D\|_0}=
\binom{\|D\|_0}{1}+\binom{\|D\|_0}{2}+...+\binom{\|D\|_0}{\|D\|_0}$, where $\binom{\|D\|_0}{k} $ 
denotes the operation of selecting $k$ teaching  examples from the full training set $D$.}.

 In  this paper, we consider one challenging problem: how to   teach a student  learner  which does not disclose any cue (also termed as a black-box learner) from  the  distribution of the parameters, i.e.  estimating  $\mathcal{L}(\hat\theta, \theta^*)$ is  inefficient due to improper  parameter disagreement  or inestimable parameter space e.g.  huge amounts of parameters in a large neural network. Our main proposal to solving Eq.~(2) in such scenario is  optimizing an approximated distribution for teaching.
Specifically, we 
  1)  approximate   $R(\theta,D)$ into a surrogate $R(\theta,D')$ by shrinking $D$ into its surrogate $D'$ with smooth  boundary, and 2) then transform $\mathcal{L}(\hat\theta, \theta^*)$ as a distribution metric optimized in $D'$  that assumes ${D}$  is w.r.t. $\hat\theta$  and 
$\mathbbm{D}$  is w.r.t. $\theta^*$.

\par Our analysis based on importance sampling \citep{beygelzimer2008importance} shows that  approximating the teaching risk by its surrogate is still  guaranteed safely to  yield a tighter  learning  bound on label complexity (number of sampled data to achieve a desired error).  
To implement the  optimization of distribution matching-based machine teaching, we employ the hyperbolic metric \citep{ganea2018hyperbolic} that hierarchically  ranks the transformed teaching risks. A Poincaré  measure $ \mathbbm{h}(\cdot,\cdot)$ \citep{sarkar2011low} is then utilized due to its effectiveness in scattering those ranked features \citep{tay2018hyperbolic}
\citep{tran2020hyperml}
\begin{equation}
\begin{split}
 \mathcal{L}(\hat\theta, \theta^*)\approx \mathbbm{h}(D , \mathbbm{D}).
\end{split}
\end{equation}

The main technical question  of this paper is as follows: how to generalize Eq.~(3) with surrogate $R(\theta,D')$. A distribution matching-based machine teaching algorithm that transfers the disagreement  estimation of parameters   into hypotheses, thereby approximating the hypothesis to distribution, is then presented by narrowing the assumption of Eq.~(3). Assume $\theta^*$ is generalized from the optimal hypothesis $h^*$, i.e. $R(\theta^*,D)=R(h^*,D)$, $\hat \theta $ is generalized from the  hypothesis $\hat h $, i.e. $R(\hat \theta, D)=R(\hat h,D)$,  let $D'\subset \mathbb{D}$ be the optimal 
surrogate with respect to $h^*$, let $\hat D\subset \mathbb{D}$ be the desired training set with respect to $\hat h$, with the proposal of Eq.~(3), we further  have 
\begin{equation}
\min_{\hat \theta\in \Theta} \|\hat\theta-\theta^*\|=\min_{\hat h\in \mathcal{H}} \|\hat h-h^*\|\approx \min_{\hat D\in \mathbb{D}} \|\hat D-D'\|_\mathbbm{h},
\end{equation}
where $\|\hat D-D'\|_\mathbbm{h}:=\mathbbm{h}(\hat D, D')$. In optimization of distribution matching-based machine teaching, minimizing $\|\hat D-D'\|_\mathbbm{h}$ is solved by controlling the teaching cost, i.e. 
  backwardly and iteratively halving  $\mathcal{M}(D')$.
Then, the final update on $D'$ is defined as the teaching set $\hat D$.

 Typical machine teaching mainly  studies  the optimal teaching set when the student learner is a white-box whose underlying parameter distributions are known, such as  linear learners \citep{liu2016teaching}, sequential learner \citep{lessard2019optimal}, Bayesian learners of the exponential family \citep{zhu2013machine}, etc. Theoretical discussion was the main trend in past decades. A complete teaching theory such as teaching dimension and teaching complexity were proposed \citep{khan2011humans}.
This paper focuses more on a black-box learner   which further leads to ineffective estimation on parameter disagreement for teaching risk.
We thus simplify the introduction of theoretical progress, and present related active learning work.
Finally, we contribute this work  from the following highlights.
\begin{itemize}
    \item  We introduce the idea of distribution matching for machine teaching a   student learner who  does not disclose any cue of  the learning parameters, i.e. a black-box.
    
    \item  We present a  cost-controlled  optimization process    that
 finds the optimal teaching examples without further exploring in  the parameter distribution. 
 
 \item We presents a  distribution matching-based machine teaching strategy  by backwardly and iteratively performing the halving operation on the teaching cost.
\end{itemize}

The organization of this paper is outlined    as  follows.  The  related work  is  listed  in  Section~2.  Section~3 presents the main theoretical results.  Section~4  presents distribution matching-based machine teaching.
Section~5  presents   experiments and Section~6  discusses the classifier perturbations to the assumption of this work,  followed by the conclusion in Section~7.

\section{Related Work}
\subsection{Machine Teaching}
When the training set of a model consists of plenty data from an underlying distribution, a teacher always desires to pick up some teaching examples from the original data set to supervise its student learner \citep{zhu2018overview}. In this teaching process, the teacher can be generalized as a human expert, a learning   algorithm or system. Two factors are studied in the teaching process: how to fix  the size of the teaching set and whether the teacher knows the full knowledge of the parameter distribution of the student model? 

\par Early works answered the first  problem by studying the teaching complexity.  
 Introducing a novel concept named teaching dimension \citep{khan2011humans},  the complexity of teaching is characterized by the
 minimum number of teaching examples must reveal to uniquely identify any hypothesis chosen from the hypothesis class.   If the teaching examples are selected independently from the original data pool, the learner can cooperate with a teacher who  supervises a teaching set.  Recently, machine teaching \citep{mei2015using} studied  the teaching model when the learner was a convex minimizer, such as   least square regression, and simple support vector machine. In \citep{DBLP:conf/icml/LiuDLLRS18}, they presented   {a theoretical interpretation against} multiple-teacher teaching rather than a single teacher.
  Under the  joint teaching for conjugate Bayesian learners,  
 Zhu et al. \citep{zhu2013machine} proposed a new concept called class teaching dimension. 
 In this definition, the teacher independently picks up some learners   as a  representation for the whole class of learners. Indeed, it shrinks the original teaching dimension by the representation features of the learner class. 
 
 \par In real-world applications, a variety of learning tasks  involved with machine teaching were studied. For instance, in  curriculum learning \citep{bengio2009curriculum}, human can intelligently annotate or recognize examples when the teaching examples  are not randomly presented but
organized in a meaningful order. Those orders help the teachers to optimize  a group of teaching set for the subsequent learning tasks.  {To improve the generalization of an active learner, \cite{DBLP:conf/icml/Dasgupta0PZ19} proposed a black-box teaching scheme  to shrink training sets for any family of classifier by merely serving up the
relevant examples beforehand, and does not need to observe the 
feedback from the learner, where ``shrink'' can be deemed as a typical distribution matching-based machine teaching strategy.  }

\subsection{Active Learning}
Active learning \citep{cohn1994improving}  adopts the same sampling goal as machine teaching to find the optimal training data, but forwardly updates  the
models.   
In this  task, the learners are given access to interactively query the labels of a group of unlabeled data. The learning goal is that
the queries can substantially improve the performance of a learning model within a given annotation budget.
Theoretically, the learners always try to maintain a version space \citep{beygelzimer2010agnostic} which covers a series of candidate  hypotheses and shrinks its size via querying as few as possible data.  However, the  version space-based learning theory  unfortunately  has drawbacks of computational intractability \citep{dasgupta2008general}, i.e.,
 guaranteeing that only hypotheses from this space are
returned is intractable for nonlinear classifiers.  

To develop a new strategy  which addresses the above limitations, \cite{Agnostic} and \cite{hanneke2007bound} constructed   learning algorithms    to predict which data may significantly affect the subsequent hypothesis, thereby giving different weight coefficients. The convergence guarantees, adopted from a PAC-style\footnote{ PAC: Probably approximately correct. In computational learning theory,  the  learner   must select a generalization function i.e. the hypothesis  from a certain class of possible functions (also called hypothesis class). The goal is that, with a high \textbf{P}robability, the selected function will have low generalization error to be \textbf{A}pproximately \textbf{C}orrect. }, is rigorous and tighter than the generalized bounds of any supervised learning algorithms. The other technique, termed as importance-weighted active learning   \citep{beygelzimer2008importance}, provides an unbiased 
sampling approach with the loss-weighting and  has more practical use in observing the error and label complexity  change.

\par In practical tasks,  traditional active learning methods such as   pool-based AL  \citep{tong2001support} samples the data that reduces the error rate in a  descried  change  by repeatedly visiting the unlabeled data pool. Usually, the learner is given access to the  hypothesis class easily, i.e. can favorably observe the hypothesis updates by estimating the error disagreements (differences). 
However, when supervising a black-box learner,  obtaining precise details of the  prior labels and list of classifier parameters are not available, and only the queries on labels are accessible. This makes many of the traditional  strategies, which estimate the error disagreements,  are not applicable, or at the least can not work well \citep{rubens2011active}.   To reduce the dependence on a single classifier, query by committee algorithm \citep{seung1992query} 
uses a set of classifiers to evaluate the error rate changes  and selects the data which  maximize the disagreement
among the committee members. However, estimating the error rate changes  in settings of single  or multiple classifiers  will cost
 expensively on time and space complexities.

\section{Main Theoretical Results}
\par The main purpose of our theoretical study is to reconstruct the original distribution   by its surrogate with a smooth boundary.
  Section~3.1 introduces importance sampling  that used for approximating  $R(\theta, D)$ and Section~3.2 provides the safety guarantee  for its surrogate. Section~3.3 presents the   label complexity bounds of minimizing $R(\theta, D')$.   Section~3.4 generalizes the approximation of $D$ to $D'$ in hyperbolic geometry. Section~3.5 presents a case study of the approximation.
Proof sketches of Theorems~1 and 2 are  presented in Appendix.

\subsection{Importance Sampling}
Importance  sampling  \citep{beygelzimer2008importance} uses importance weighting to correct sampling bias and rigorously observe the on-line error change for a machine learning model.  In this section, we find a surrogate $D'$ to approximate $R(\theta,D)$ by employing the importance sampling algorithm,  
which further eliminates the   noisy perturbations around the boundary of the distribution.     

In importance sampling, the machine learning algorithm assigns an unlabeled data $x_t \in D$ with a  probability  $p_t$ to query its label $y_t$. The underlying rule is: if $x_t$  is selected for   querying at $t$-time of sampling, its weight is set to $\frac{1}{p_t}$.  Let $f(\cdot)$ denote the mapping loss function   from $ D $ to $\mathcal{Y}$,   given a classification hypothesis $h: D  \times \mathcal{Y}\rightarrow \mathbbm{R}$, where $\mathcal{Y}$ denotes the label space of  $D$, {let  ${\rm err}_T(h)$ be  the expected error loss over $D$} of a hypothesis $h$ at query time $T$, the learning risk with $T$ times of sampling from $D$ is defined as 
\begin{equation}
\begin{split}
{R(h, D, T):=} \ {\rm err}_T(h)=  \frac{1}{T}\sum_{t=1}^{T}\frac{q_t}{p_t}f(h(x_t), y_t),
\end{split}
\end{equation}
where   $q_t$ denotes a Bernoulli distribution with   $q_t \in \{0,1\}$,  $y_t \in \mathcal{Y}$, and it denotes the true label of $x_t$. Importance sampling 
 uses the probability weights to  eliminate the sampling bias with rigorous label
complexity bounds  {to accelerate the convergence of the minimization on $R(h,D)$,  where label complexity denotes  the number of the sampled data to achieve a desired error.}  We  next  exploit the idea of importance sampling to    present the label complexity for the approximated surrogate and its safety guarantee.

\subsection{Safety Guarantee for Surrogate}
{In importance sampling,  learning in surrogate $D'$  can keep consistent properties for the machine learning model but eliminates  the noisy perturbations from the boundary of the distribution i.e. its conceptual version space.}   Theoretically, a desired safety guarantee \citep{beygelzimer2008importance} expects that the performance of a machine learning algorithm   keeps  a provably consistency on its inherent optimal hypothesis.

Given an agnostic distribution ${P}$  maintaining a training set $D$ for sampling, we define a subset  $D'$ with a smooth boundary as its surrogate.
 \begin{definition} Surrogate $D'$ of $D$. 
Given a finite hypothesis class $\mathcal{H}$ {with finite VC dimension \footnote{Vapnik–Chervonenkis  dimension. It is a measure of the capacity such as complexity of a space of functions that can be learned by a  classification learning algorithm. In VC theory \citep{vapnik2013nature}, the VC bound  is defined as the cardinality of the largest set of input training data that a learning algorithm can shatter.} bound $d$} that is uniquely  associated with $D$. Let $D'$ be a  surrogate   of $D$ and $\mathcal{H}'$ be the shrunk hypothesis class over $D'$.  
Assume that $d_\mathcal{H}(D)$ and $d_{\mathcal{H}'}(D')$  {are the hypothesis diameters (maximum hypothesis disagreement)} of $\mathcal{H}$ and $\mathcal{H}'$, respectively, for any probability $\delta$, surrogate $D'$ is one subset from $D$ with a smooth boundary that satisfies  
\begin{equation}
\begin{split}
 &d_\mathcal{H}(D)- d_{\mathcal{H}'}(D')\leq  \sqrt{2\Big({\rm In}\  d+{\rm In} \frac{2}{\delta}\Big)}, \\
 &{\rm  s.t. } \  D'\subset D, \|D'\|_0=n'<\|D\|_0=n,  \\
 &d_\mathcal{H}(D):= {\rm max} \Bigg\{ \Big|{\rm err}_D(h^+, x_t^+)-{\rm err}_{D}(h^-, x_t^-)\Big|_\mathcal{H} \Bigg\},     \\
 &d_{\mathcal{H}'}(D'):={\rm max} \Bigg\{\Big|{\rm err}_{D'}(h^+, x_t^+)-{\rm err}_{D'}(h^-, x_t^-)\Big|_{\mathcal{H}'} \Bigg\}, \\
\end{split}
\end{equation} 
where   ${\rm err}_D(h^+, x_t^+)$    denotes the error of   a subsequent hypothesis $h^+$   over 
 $D$ after adding and annotating $x_t$ with a positive label,   ${\rm err}_D(h^-, x_t^-)$ follows the annotation assumption of a negative label, ${\rm err}_{D'}(h^+, x_t^+)$ and ${\rm err}_{D'}(h^-, x_t^-)$ also follow a surrogate $D'$,  and  $x_t$ is the sampled data from $D$ at $t$-time. 
\end{definition}
 
\begin{definition}
 Safety guarantee. Given   $\mathbb{E}_{\theta \in \Theta}\ R(\theta, D')$ be the expected empirical risk over surrogate $D'$, let $d$ be the  finite VC dimension bound  that is uniquely  associated with $D$, for any probability $\delta$,    if 
\begin{equation}{\rm  Pr}\Bigg \{   | R(\theta, D)   - \mathbb{E}_{\theta \in \Theta}R(\theta, D') |\!\leq\! \! \sqrt{2\Big({\rm In}\  d+{\rm In} \frac{2}{\delta}\Big)}\Bigg \}\approx 1,\end{equation}
any machine learning model  that minimizes $R(\theta, D)$ is guaranteed safely on minimizing  $R(\theta, D')$. 
\end{definition}
 \begin{assumption}
 With importance sampling, assume that $\theta\in \Theta$ is respected to $h\in \mathcal{H}$, recalling Eq.~(4),  approximating $R(\theta, D)$ into $R(\theta, D')$ is equivalent to approximating $R(h, D)$ into $R(h, D,T)$,  where $R(h, D, T)$ denotes  error risk of   $T$ times of importance sampling   w.r.t. Eq.~(5) and $R(h, D)=\mathbb{E}_{(x,y)\sim {D}} f(h(x),y)$ without importance sampling.
\end{assumption}

Theorem~1 observes the  ground-truth risk disagreement  and its expectation, where the risk disagreement is over the full training data and its surrogate. 
\begin{theorem}
With Assumption~1, given the training set ${D}$, for all finite hypothesis class $\mathcal{H}$ with a VC dimension bound $d$, for any probability $\delta>0$ and $p_t>\phi$, if a  learning algorithm samples $T$ times to obtain a  surrogate of $D$,   let   $\textbf{\rm R}$ be the ground-truth  risk disagreement of the surrogate and its full training data that stipulates $\textbf{\rm R}=|R(h, D, T)- R(h, D)|$,  $\hat{\textbf{\rm R}}$ be the expected  risk disagreement that stipulates $\hat{\textbf{\rm R}}=\mathbb{E}_{h\in \mathcal{H}} |R(h, D, T)- R(h, D)|$,  with Definition~2, the generalization probability bound of  achieving a safe surrogate is
\begin{equation}
\begin{split}
{\rm  Pr}\Bigg \{   | \textbf{\rm R}  -\hat{\textbf{\rm R}} |\!\leq\! \! \sqrt{2\Big({\rm In}\  d+{\rm In} \frac{2}{\delta}\Big)}\Bigg \}\!\!\leq\!\exp\!\Bigg ({\frac{-4({\rm In} \ d+{\rm In} \frac{2}{\delta}) }{T  \phi^{-2}}}\Bigg ).
\end{split}
\end{equation} 

\end{theorem}

In brief, Theorem~1 shows that  there exists nearly consistent hypothesis diameters between the full training data and its surrogate, where the diameter of surrogate is over its expectation.

\begin{corollary}  In Theorem~1,  $\textbf{\rm R}$ and  $\hat{\textbf{\rm R}}$   denote the   \emph{maximum} and \emph{expected}  risk disagreement of the $T$ times  importance sampling and full training data, respectively. For a given hypothesis class $\mathcal{H}$ which covers all feasible hypotheses, the maximum error disagreement is close to the hypothesis diameter \citep{tosh2017diameter} of  $\mathcal{H}$. If the expected hypothesis distance of a sub hypothesis class over $D'$  is close to it, we say sampling in $D'$  yields consistency as sampling in $D$. Therefore, with Definition~1, Theorem~1 has another equivalent form 
\begin{equation}
{\rm  Pr}\Bigg \{   |  d_\mathcal{H}(D)- \mathbb{E}_{D'\subset D}   d_\mathcal{H'}(D')|\!\leq\! \! \sqrt{2\Big({\rm In}\  d+{\rm In} \frac{2}{\delta}\Big)}\Bigg \}\!\!   \approx 1.
\end{equation} 
 \end{corollary}
{Specifically, the probability bound of Eq.~(8) approximates 1. With Assumption~1, approximating $R(\theta,D)$ into $R(\theta,D')$ achieves safety guarantee for any  $\theta \in \Theta$ over $D'$.}

\subsection{Label Complexity Bound for Minimizing $R(\theta,D')$}
We follow \citep{Agnostic} to present the label complexity of  minimizing $R(\theta,D')$. 
\begin{assumption}
Let $n$ denote the sample amount in  $D$, its VC dimension bound $d$ approximates to $2^n$. By using importance sampling,  $D'$ is with a  VC bound $2^T$.
\end{assumption}

With Assumption~2, an upper bound of the label complexity of minimizing $R(\theta,D')$ is presented.
\begin{theorem}
 Given the slope asymmetry $K_f$ that bounds the   loss function $f(h(x),y)$ w.r.t. Eq.~(5) for any hypothesis $h$ over $D'$: $K_f= \mathop{{\rm sup}}\limits_{x_t', x_t\in D'} \left |\frac{{\rm max} \ {\rm err}_{D'}(h(x_t), \mathcal{Y})-{\rm err}_{D'}(h(x_t'), \mathcal{Y}) }    {{\rm min} \ {\rm err}_{D'}(h(x_t), \mathcal{Y})- {\rm err}_{D'}(h(x_t'), \mathcal{Y}) } \right|$, considering a disagreement coefficient $\vartheta= \mathbb{E}_{x_t\in \mathcal{D}}   \mathop{{\rm sup}}\limits_{h\in B(h^*,r)} \left \{  \frac{ \ell(h(x_t),\mathcal{Y})-\ell(h^*(x_t),\mathcal{Y})   }{r}       \right\}$,  
 if the learning algorithm uses  $\vartheta$
 to smooth those data of $D$ with smaller hypothesis disagreements than $r$, with a probability $1-\delta$, at $t$-time, {minimizing $R(\theta,D')$ into  $R(\theta^*,D')$, i.e. 
   updating the current hypothesis $h$ into the optimal hypothesis $h^*$} in surrogate $D'$, costs at most  $4\vartheta\times K_f\times \Bigg(R(h^*,D')+2\sqrt{\frac{8}{t-1} {\rm In} \Big( \frac{2(t^2-t) \frac{d}{2^{n-T}}}{\delta}\Big)} \Bigg) $.
\end{theorem}

Note that $\vartheta$ is an  error disagreement parameter that used to perform the importance sampling. Any hypothesis $h$ holding a  hypothesis disagreement  smaller  than $r$ in terms of  $\vartheta$, will be considered as a null hypothesis  which presents insignificant influence for updating the current model, thereby being
smoothed from the candidate hypothesis class. 
More related analysis based on this class of  error disagreement parameters can refer to Hanneke's work e.g. \citep{hanneke2007bound} \citep{hanneke2014theory}.

Note that $K_f$ is a constant that satisfies $K_f\geq 1$.  Based on the importance sampling of \citep{beygelzimer2008importance}, $K_f$ affects the label complexity bound due to its ``sensitivity". For example,  given a 0-1 loss for $f(h(x),y)$, $K_f$ will be 1. However, for a hinge loss, $K_f$ will be $\infty$. Therefore, for a sensitive loss function, the learning algorithm will require a large number of importance sampling 
times to obtain a desired hypothesis, then may lead to many ineffective queries. In other words, the sensitive loss function usually presents a coarse estimation on hypothesis disagreements. We here present a  lemma  to improve the generalization of $K_f$.
\begin{lemma}
Let $h$ be generalized as a logistic hypothesis that stipulates  $f(h(x),y):={\rm In}(1+e^{xy}) $, assume that the label space $\mathcal{Y}\in [-1,+1]$,  if $x\in[-M,M]$,  $K_f$  can be as large as $(1+e^M)$.
\end{lemma}

\subsection{Approximating $D$ into $D'$  using Poincaré Distance }
\par Poincaré distance \citep{ganea2018hyperbolic} of hyperbolic geometry  has presented an effective improvement in latent hierarchical  tasks compared to  Euclidean distance ($\|\cdot \|_2$) such as ranking features \citep{tay2018hyperbolic} \citep{tran2020hyperml}, embeddings \citep{nickel2018learning},  non-linear gradient descending \citep{nitta2017hyperbolic}, etc. To approximate $D$ into $D'$, we need to rank   one property of all its members  based   specified estimations such as clustering property, density characteristics, geometric structure, etc.   Poincaré distance thus is introduced to implement the ranking of the approximation progress. 

\par Let $\mathcal{B}^{\bm{d}}=\{ x \in \mathbbm{R}^{\bm{d}}, \|x\|_2 < 1 \}$ be an  open $\bm{d}$-dimensional unit  Poincar\'e sphere ($\|\cdot\|_2$ denotes the $L^2$ norm), $u$ and $v$ be any two vectors in the sphere, i.e. $u, v \in \mathcal{B}^{\bm{d}}$, the Poincar\'e    distance between  them is defined as
\begin{equation}
\begin{split}
\mathbbm{h}(u,v)={\rm arccosh}\Bigg(1+2 \frac{\|u-v\|^2 }{(1-\|u\|^2)(1-\|v\|^2) }\Bigg).
\end{split}
\end{equation}

\par Based on the work of \cite{cao2018multidimensional},  noisy perturbations around the boundary usually are characterized with low  density observations. We thus estimate the  density of the data  constrained within a fixed   hypersphere  
\begin{equation}
\begin{split}
&\mathbb{S}:=\{ v, \forall v\in D, \ {\rm s.t.}\  \psi(u,v)=1  \},\\
&{\rm s.t. }\  \psi(u,v)=  \left\{
             \begin{array}{lr}
             0, &  \mathbbm{h}(u,v)>\mathbbm{r},  \\
            1, & 0 \leq  \mathbbm{h}(u,v)\leq  \mathbbm{r},  \\
             \end{array}
\right. 
\end{split}
\end{equation}
where  $\mathbbm{r}$ denotes the radius of  the hypersphere $\mathbb{S}$ centered with $u$. A more general equation that applies  hypersphere $\mathbb{S}$ to observe the density on $u$ is presented
\begin{equation}
\begin{split}
f_\mathbb{S}(u)=\frac{1}{\|\mathbb{S} \|_0} \sum_{v \in \mathbb{S}} \frac{1}{\sqrt{2\pi}^{\bm{d}} \mathbbm{r}^{\bm{d}}} {\rm exp} \Bigg [-\frac{1}{2}\left(\frac{\mathbbm{h}(u,v)}{\mathbbm{r}} \right)^2   \Bigg].
\end{split}
\end{equation}

   \begin{figure*}[!t]
\subfloat[\emph{digit}]{
\label{fig:improved_subfig_b}
\begin{minipage}[t]{0.49\textwidth}
\centering
\includegraphics[width=2.9in,height=2.31in]{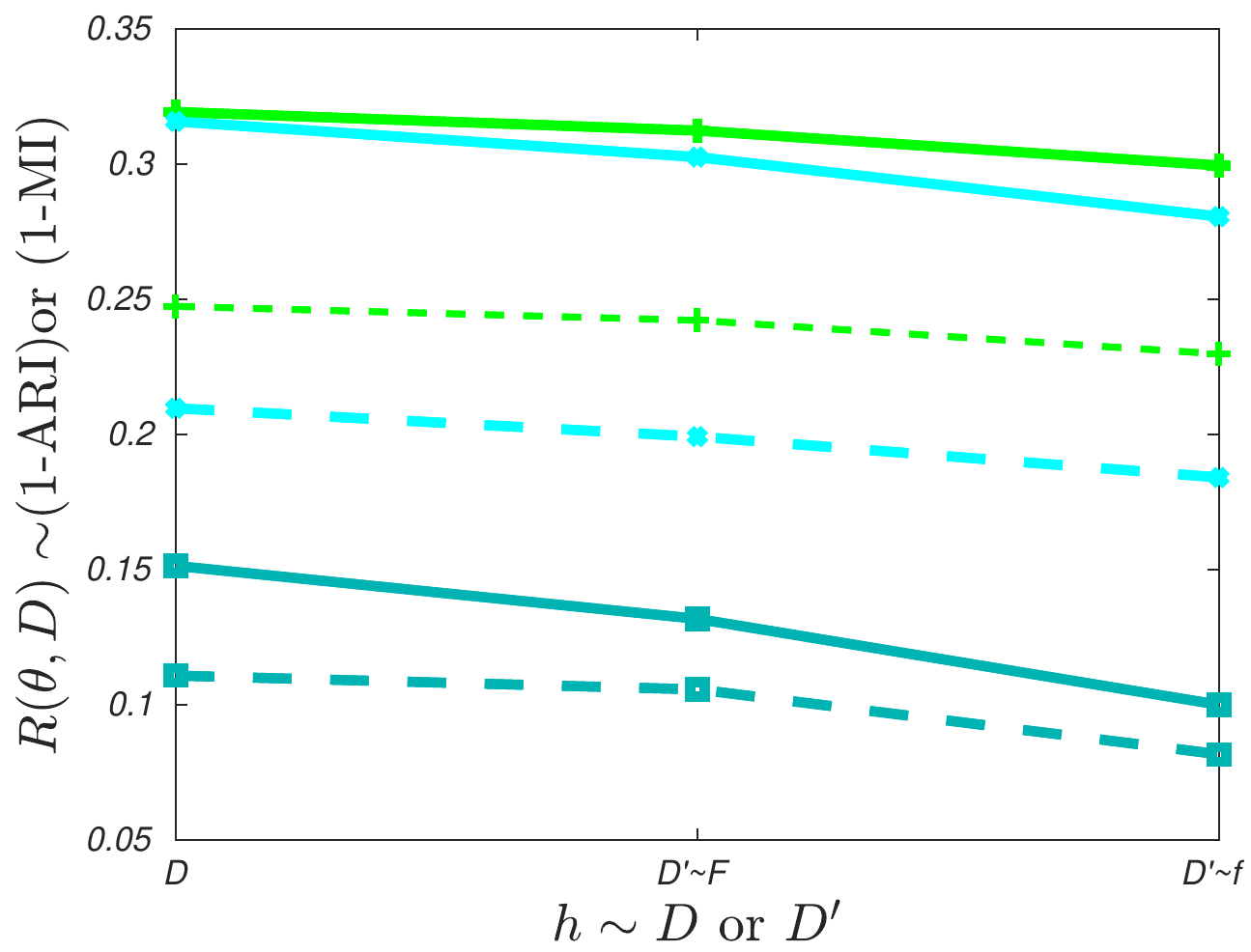}
\end{minipage}
}
\subfloat[ \emph{USPS}  ]{
\label{fig:improved_subfig_b}
\begin{minipage}[t]{0.49\textwidth}
\centering
\includegraphics[width=2.9in,height=2.31in]{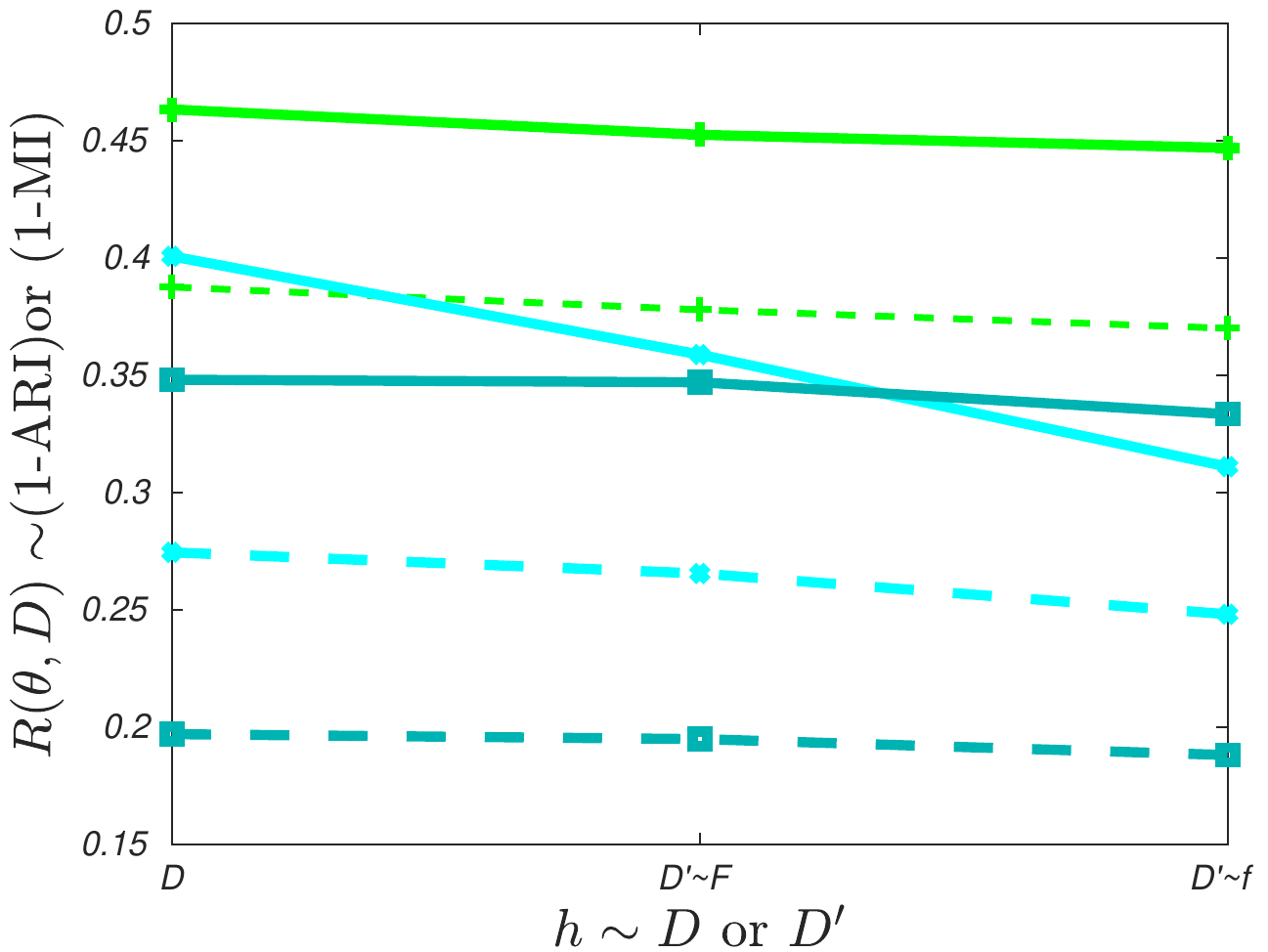}
\end{minipage}
}\\
\subfloat[\emph{FashionMnist}]{
\label{fig:improved_subfig_b}
\begin{minipage}[t]{0.49\textwidth}
\centering
\includegraphics[width=2.9in,height=2.31in]{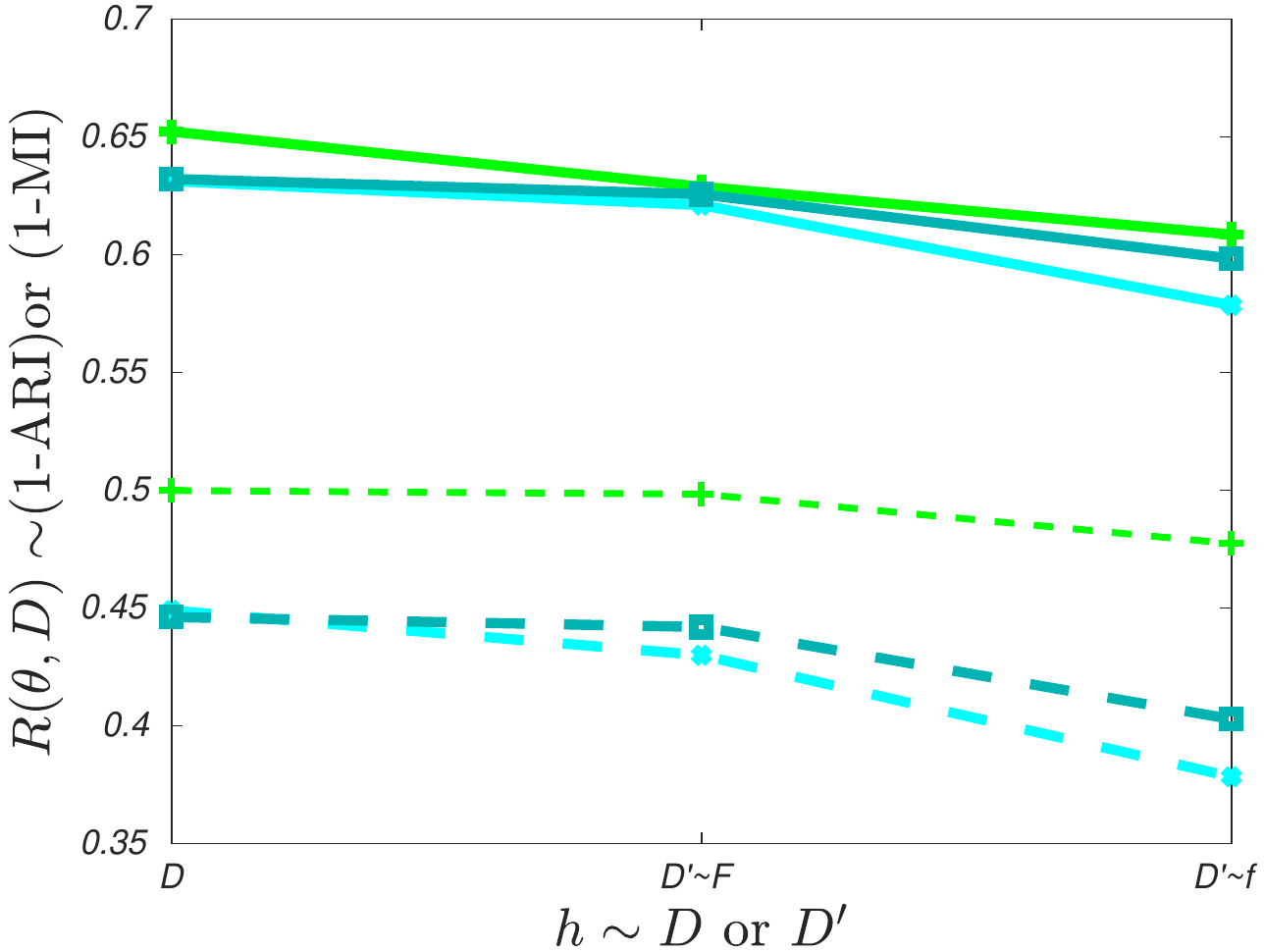}
\end{minipage}
}
\subfloat{
\label{fig:improved_subfig_b}
\begin{minipage}[t]{0.49\textwidth}
\centering
\includegraphics[width=1.83in,height=1.71in]{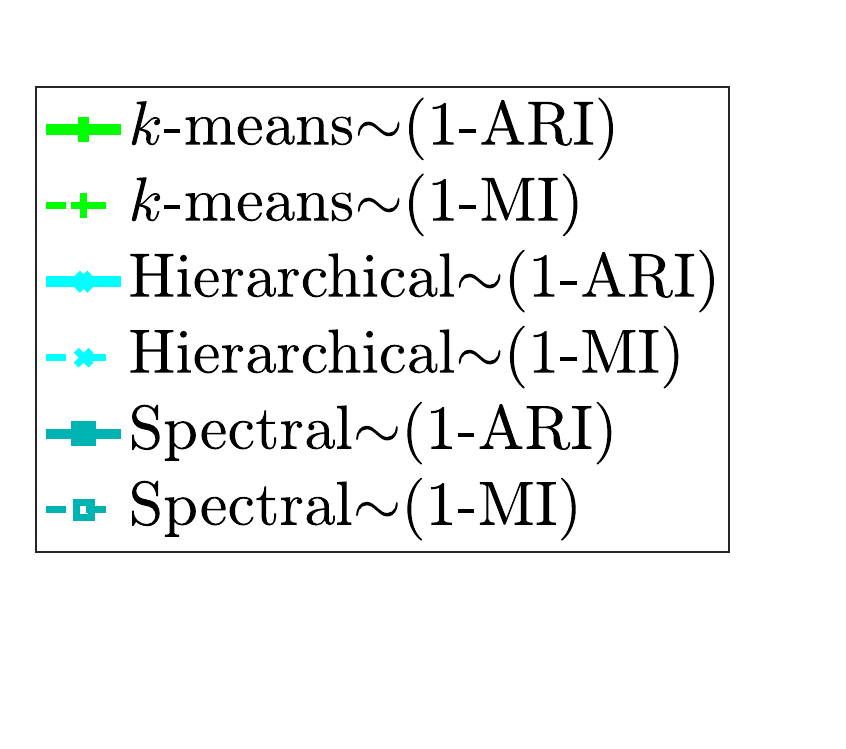}
\end{minipage}
}

\caption{Case study of clustering on surrogate with a smooth boundary  to improve clustering.   $R(\theta,D)$ is generalized as (1-ARI) or (1-MI) coefficients. } 
\end{figure*}
\subsection{Case Study:  Improving Clustering on Surrogate}
This case study collects three real-world data sets and then compares the  clustering performance of three typical clustering baselines in $D'$, where $f_\mathbb{S}(u)$ is used to approximate $D$ into $D'$. To show the advantages of Poincaré distance in ranking,   
$f_\mathbb{S}(u)$  is generalized into $F_\mathbb{S}(u)$ characterized with Euclidean distance, which is further used to compare Eq.~(12):
   \begin{equation}
\begin{split}
F_\mathbb{S}(u)=\frac{1}{\|\mathbb{S} \|_0} \sum_{v \in \mathbb{S}} \frac{1}{\sqrt{2\pi}^m \mathbbm{r}^{\bm{d}}} {\rm exp} \Bigg [-\frac{1}{2}\left(\frac{\|u-v\|_2)}{\mathbbm{r}} \right)^2   \Bigg].
\end{split}
\end{equation}

\par Datasets of the case study are \emph{digit, USPS,} and \emph{FashionMnist}, where all the features of the data are scaled within an numerical  unit of $10^{-5}$  to satisfy $\|x_i\|_2<1$, $\forall x_i\in D$.  The sizes of these data sets are $3823\times 65$, $9,298\times 257$,  and $70,000\times 784$, respectively.    $R(\theta,D)$    is generalized as  (1-adjusted rand index (ARI))\citep{hubert1985comparing} and (1-mutual information (MI))\citep{vinh2010information} coefficients. 
  
  \par Figure~2 presents the $R(\theta,D)$ values that yield 1) minimizing $R(h,D)$ by clustering baselines, 2) minimizing $R(h,D')$ by clustering baselines with $F_\mathbb{S}(u)$, and minimizing $R(h,D')$ by clustering baselines with $f_\mathbb{S}(u)$, where the clustering baselines are generalized as $k$-means, hierarchical,   and   spectral clustering algorithms,  the parameters $\mathbbm{r}$ of Eqs.~(11)  and (12) are defined as 0.4, and $\|D\|_0-\|D'\|_0=0.05n$, i.e. approximate $D$ into $D'$ by eliminating perturbations  from $0.05n$ boundary examples. Specifically, 
  kernel function of spectral clustering is set as  RBF, driving a kernel parameter as 0.1 to construct an affinity matrix, where a $k$-means clustering is used  to  assign labels in the embedding space of the kernel.
  
\par As the shown in Figure~2, the solid lines    yield (1-ARI), and dash lines   yield (1-MI). $h\sim D$ denotes  performing clustering in $D$, referring  $R(\theta,D)$  values on y-axis.   $h\sim D'$ includes  $D'\sim F$  and $D'\sim f$, where $D'\sim F$
denotes that $D'$ is approximated by $F_\mathbb{S}(u)$ w.r.t. Eq.~(13), and  $D'\sim f$ denotes that $D'$ is approximated by $f_\mathbb{S}(u)$ w.r.t. Eq.~(12).
 It is intuitively that clustering in $D'$ yields $R(h,D')\leq R(h,D)$. Moreover,  approximating $D$ into $D'$ using $f_\mathbb{S}(u)$ ($D'\sim f$) achieves lower empirical risks than that of $F_\mathbb{S}(u)$ ($D'\sim F$) due to its hierarchical metric on ranking. This further demonstrates that  density observations employing Poincaré distance can yield more accurate surrogate than Euclidean distance on eliminating the noisy perturbations around the 
boundary.

\section{Distribution Matching-based Machine Teaching}
Section~4.1 presents the assumption of the distribution matching-based machine teaching. Section~4.2  presents the detailed optimization scheme by
generalizing $\mathbbm{h}(D, \mathbbm{D})$ in hyperbolic geometry. Section~4.3 describes the distribution matching-based machine teaching algorithm.

\subsection{Assumption}
With the effectiveness of approximating $R(\theta,D)$ in hyperbolic geometry, the generalization of $\mathbbm{h}(D, \mathbbm{D})$ also follows this non-Euclidean structure. Recalling Eq.~(3), we here present a more  formal  assumption against teaching a black-box learner: transfer the disagreement  estimation of parameters   into hypotheses, thereby approximating the hypothesis to distribution.

\begin{assumption}
Assume that $\theta^*$ is generalized from the optimal hypothesis $h^*$, i.e. $R(\theta^*,D)=R(h^*,D)$, $\hat \theta $ is generalized from the  hypothesis $\hat h $, i.e. $R(\hat \theta, D)=R(\hat h,D)$,  let $D'\subset \mathbb{D}$ be the optimal 
surrogate with respect to $h^*$, let $\hat D\subset \mathbb{D}$ be the desired training set with respect to $\hat h$, with the proposal of Eq.~(3), we further  have  
\begin{equation}
\min_{\hat \theta\in \Theta} \|\hat\theta-\theta^*\|=\min_{\hat h\in \mathcal{H}} \|\hat h-h^*\|\approx \min_{\hat D\in \mathbb{D}} \|\hat D-D'\|_\mathbbm{h},
\end{equation}
where $\|\hat D-D'\|_\mathbbm{h}:=\mathbbm{h}(\hat D, D')$. 
\end{assumption}
Another expression of Eq.~(14) is  $\mathcal{L}(\hat\theta, \theta^*)=\min_{\hat \theta\in \Theta} \|\hat\theta-\theta^*\|\approx \mathbbm{h}(D , \mathbbm{D})$. We thus have the following remark.
\begin{remark}
With Assumption 3, the  optimization of distribution matching-based machine teaching over  $\mathbbm{h}(D, \mathbbm{D})$  is generalized into  $\min_{\hat D\in \mathbb{D}} \|\hat D-D'\|_\mathbbm{h}$.
We thus iteratively halve $\mathcal{M}(D')$ i.e. $\|D'\|_0$, which linearly  reduces the teaching cost. With iterative halving, $\mathcal{M}(D')$ varies from  $n'/2$ into $n'/4, n'/8, n'/16,...,n'/{2^l}$, where
$l$ denotes the halving frequency,   $n'/{2^i}=\lceil n'/{2^i} \rceil$, and the
remaining examples after the $l$th halving are the final teaching test if the learner does not control the output $\mathcal{M}(D')$.  Specifically, the halving process is   implemented with the  Poincaré distance of hyperbolic geometry. With iterative halving on $D'$, the final update on $D'$ is defined as the teaching set $\hat D$.
\end{remark}

\subsection{Cost-controlled Optimization}
With Remark~1,  the optimization of distribution matching-based machine teaching over generalized  $\min_{\hat D\in \mathbb{D}} \|\hat D-D'\|_\mathbbm{h}$ is solved by controlling the teaching cost of $D'$, that is performing
a continuous algorithmic halving  on $D'$, where the final update on $D'$ is the desired target $\hat D$.

The algorithm begins by  generalizing $R(\theta,D)$. Let $D’=\{\mathcal{D}_1,\mathcal{D}_2, ... , \mathcal{D}_{n'} \}\sim P$,  $R(\theta, \mathcal{D}_i):=\theta \mathcal{D}_i+\varepsilon_i$, where $\varepsilon_i$ 
denotes a constant  constraint on $\mathcal{D}_i$, machine teaching with a black-box  is to optimize $\theta^*$ 
\begin{equation}
\theta^*=\argmin_{\theta\in \Theta}\Bigg\{\|\theta-\theta^*\|:= \sum_{i=1}^{n'}\Big\|\theta \mathcal{D}_i  -R(\theta, \mathcal{D}_i) \Big\|^2             \Bigg\}.
\end{equation}
Recalling Eq.~(1), we add a regularization constraint $\Omega(\theta):=\eta\|\theta\|^2$ to  Eq.~(15)
\begin{equation}
\begin{split}
\theta^*=\argmin_{\theta\in \Theta}
\Bigg\{\|\theta-\theta^*\|+ \Omega(\theta):= \sum_{i=1}^{n'}\Big\|\theta \mathcal{D}_i  -R(\theta, \mathcal{D}_i)\Big\|^2+  \eta\|\theta\|^2             \Bigg\}.
\end{split}
\end{equation}
Based on Assumption~3, estimating the parameter disagreement can be transferred into distribution disagreement.
We next introduce the Poincaré distance $\mathbbm{h}(\cdot,\cdot)$ that stipulates $R(\theta, \mathcal{D}_i):= \sum_{j=1}^{n'/2} \alpha_j  \mathbbm{h}(\mathcal{D}_j, \mathcal{D}_i)$, then Eq.~(16) is equivalent to
\begin{equation}
\begin{split}
\min_\alpha \sum_{i=1}^{n'/2}\Bigg\| \sum_{j=1}^{n'/2}\alpha_j  \mathbbm{h}(\mathcal{D}_j, \mathcal{D}_i)\!&-\!\theta \mathcal{D}_i)\Bigg\|^2\! 
 +\! \sum_{i=1}^{n'/2}  \sum_{j=1}^{n'/2}\alpha_i\alpha_j  \mathbbm{h}(\mathcal{D}_i, \mathcal{D}_j),\\
\end{split}
\end{equation}
where $\alpha=[\alpha_1, \alpha_2, ... , \alpha_{n'/2}]$.
 Let $\textbf{H}_{ij}:= \mathbbm{h}(\mathcal{D}_i, \mathcal{D}_j)$, $[\textbf{H}_{\hat {D'}D'}]_{ij}:= \mathbbm{h}(\hat{\mathcal{D}}_i, \mathcal{D}_j)$, where $\hat{D'}\subset D',\ {\rm s.t.} \  \| \hat{D'} \|_0=n'/2$, with Definition~3.1 in   \citep{yu2006active},  Eq.~(17) is transferred as  $\mathbbm{h}(\hat {D'} ,  D')$ that can be solved by transductive optimization 
\begin{equation}
\begin{split}
 \mathbbm{h}(\hat {D'} ,  D'):= \min_{\hat{D'}\subset D'} \textbf{Tr}\Bigg[\textbf{H}_{\hat {D'}D'}(\textbf{H}_{  {D'}D'}+\eta \textbf{C})^{-1}\textbf{H}_{ {D'}\hat {D'}}\Bigg],
\end{split}
\end{equation}
where $\textbf{C}\varpropto    {D'}  {D'}^T$. To optimize $\hat{D'}$, let $\mathcal{D}$ be the last selected teaching example, $\hat{D'}_i$ subsequently is obtained by
\begin{equation}
\begin{split}
\hat{D'}_i:=\argmin_{\mathcal{D}\in D' } \textbf{Tr}\Bigg[\textbf{H}_{\hat {D'}\mathcal{D}}(\textbf{H}_{ \mathcal{D}\mathcal{D}}+\eta \textbf{C})^{-1}\textbf{H}_{ \mathcal{D}\hat {D'}}\Bigg],
\end{split}
\end{equation}
where $ \textbf{H}=\textbf{H}-\textbf{H}_{\hat {D'}\mathcal{D}}(\textbf{H}_{ \mathcal{D}\mathcal{D}}+\eta \textbf{C})^{-1}\textbf{H}_{ \mathcal{D}\hat {D'}}$.

\begin{algorithm}[t]
 \caption{Distribution Matching-based Machine   Teaching Algorithm}
   \textbf{Input:} Full training set $D$, $0\leftarrow j$.\\
     Estimating    $f_\mathbb{S}(\mathcal{D}_i)$ for any $\mathcal{D}_i$:\\
$\!\mathcal{F}(i)  \leftarrow   f_\mathbb{S}(\mathcal{D}_i)=\frac{1}{\|\mathbb{S} \|_0} \sum_{v \in \mathbb{S}} \frac{1}{\sqrt{2\pi}^{\bm{d}} r^{\bm{d}}} {\rm exp} \Bigg [\!-\frac{1}{2}\left(\frac{\mathbbm{h}(\mathcal{D}_i,v)}{r} \right)^2   \Bigg] $.\\
       Keep top $n'$ examples with large  $\|\mathcal{F}\|_1$ to obtain $D'$.\\
     \While {$j\leq l$}{ 
       \For{$i=1,2,...,\frac{n'} {2^l}$  }{
  $ \hat{D'}_i=\argmin_{\mathcal{D} \in D'} \textbf{Tr}\Bigg[\textbf{H}_{\hat {D'}\mathcal{D}}(\textbf{H}_{ \mathcal{D}\mathcal{D}}+\eta \textbf{C})^{-1}\textbf{H}_{ \mathcal{D}\hat {D'}}\Bigg]$,\\
  s.t.  $\textbf{H}=\textbf{H}-\textbf{H}_{\hat {D'}\mathcal{D}}(\textbf{H}_{ \mathcal{D}\mathcal{D}}+\eta \textbf{C})^{-1}\textbf{H}_{ \mathcal{D}\hat {D'}}$.
     \\
       }
      
     $D' \leftarrow \hat{D'}.$\\
        $j  \leftarrow j+1.$
        }

 \textbf{Output:} the final teaching set   $\mathcal{M}( \hat{D'})$; if the student learner   controls the out $\mathcal{M}( \hat{D'})$, performing $k$-medoids on $ \hat{D'}$ to satisfy the request.
\end{algorithm}

\subsection{Distribution Matching-based Machine   Teaching Algorithm}
Our distribution matching-based machine teaching  algorithm is presented in Algorithm~1. Here, $l$ denotes the frequency of performing the halving operation on $D'$ with a default constraint of   $\mathcal{M}(D')=n'=0.95\|D\|_0$. Lines~2 to 4 approximate $R(\theta,D)$ into $R(\theta,D')$ by shrinking $D$ into its surrogate $D'$. Lines 5 to 12 perform the iterative halving process on $D'$. The final update on $D'$ after $l$ times halving is the machine teaching set  $\hat{D'}$.   If the student learner   controls the output  $\mathcal{M}(\hat D')$,  $k$-medoids is performed on the final update of $\hat D'$ to satisfy the student learner's request.

\section{Experiments}
Typical  machine teaching algorithms  estimate the parameter disagreement of models to generalize  the teaching risk, where  the teacher  knows the desired  parameter, i.e. the learner is a white-box.   When teaching a black-box learner, parameter estimations 
 may be inefficient  due to improper  parameter disagreement  or inestimable parameter space.  We thus select a series of supervised and unsupervised machine learning baselines, which can be generalized as white-box teaching, to compare our distribution-based machine   teaching   algorithm. 

To solve Eq.~(2) of general machine teaching, there exists three conditions which can simply its optimization process: 1) control $\mathcal{M}(D)$ with Eq.~(1)'s solver of  active learning, 2) reduce the search space for limited risk minimization, and 3) fix  $\mathcal{M}(D)$ with unsupervised machine learning.  To realize these conditions, three groups of experiments  are presented: 
\begin{itemize}
    \item regulating $\mathcal{M}(D)$ to minimize the  risk disagreement of $\|R(\theta,D)-R(\theta^*,D) \|$, i.e. supervised way;
    \item   reducing the search space of $\mathbbm{D}$ to observe the perturbations to typical machine learning and our machine teaching algorithms;  
    \item  minimizing $\|R(\theta,D)-R(\theta^*,D) \|$ with quantitative  $\mathcal{M}(D)$, i.e. unsupervised way.
\end{itemize}

 Data sets. The data sets used in the first two groups of experiments are the full training data of   \emph{Adult},   \emph{Phishing}, \emph{Satimage}, and \emph{MNIST} data sets, where $\|R(\theta,D)-R(\theta^*,D) \|$ is over those training data.  The sizes of these data sets  are 11,055$\times$ 68,  1,605 $\times$ 14,   4,435$\times$36, and  60,000$\times$ 780, respectively.  The data sets used in the third group of experiment are CIFAR10 and CIFAR100, where $\|R(\theta,D)-R(\theta^*,D) \|$ is over their test data.  The sizes of the two data sets  are all 60, 000 with 32$\times$32 pixels.

 Baselines.  Four  supervised learning algorithms  that regulate  $\mathcal{M}(D)$ to minimize $\|R(\theta,D)-R(\theta^*,D) \|$  are selected including expected error reduction (ERR) \citep{roy2001toward},  Pre-clustering \citep{dasgupta2008hierarchical},  transductive experimental design (TED) \citep{yu2006active} and self-paced active learning (SPAL) \citep{tang2019self}.  Specifically, they are active learning algorithms.
 Three typical unsupervised machine learning algorithms that minimize $\|R(\theta,D)-R(\theta^*,D) \|$ with quantitative  $\mathcal{M}(D)$ are selected: $k$-medoids, hierarchical, and spectral clustering. Those baselines are finally used in experiment of teaching a deep   neural network. A case study of teaching on Gaussian data is firstly presented before the  experiments. Note that distribution-based machine   teaching is denoted as DM-based machine   teaching  in all experimental figures.
  \begin{figure*}[!t]
\subfloat[Gaussian dataset $D$]{
\label{fig:improved_subfig_b}
\begin{minipage}[t]{0.49\textwidth}
\centering
\includegraphics[width=2.9in,height=2.31in]{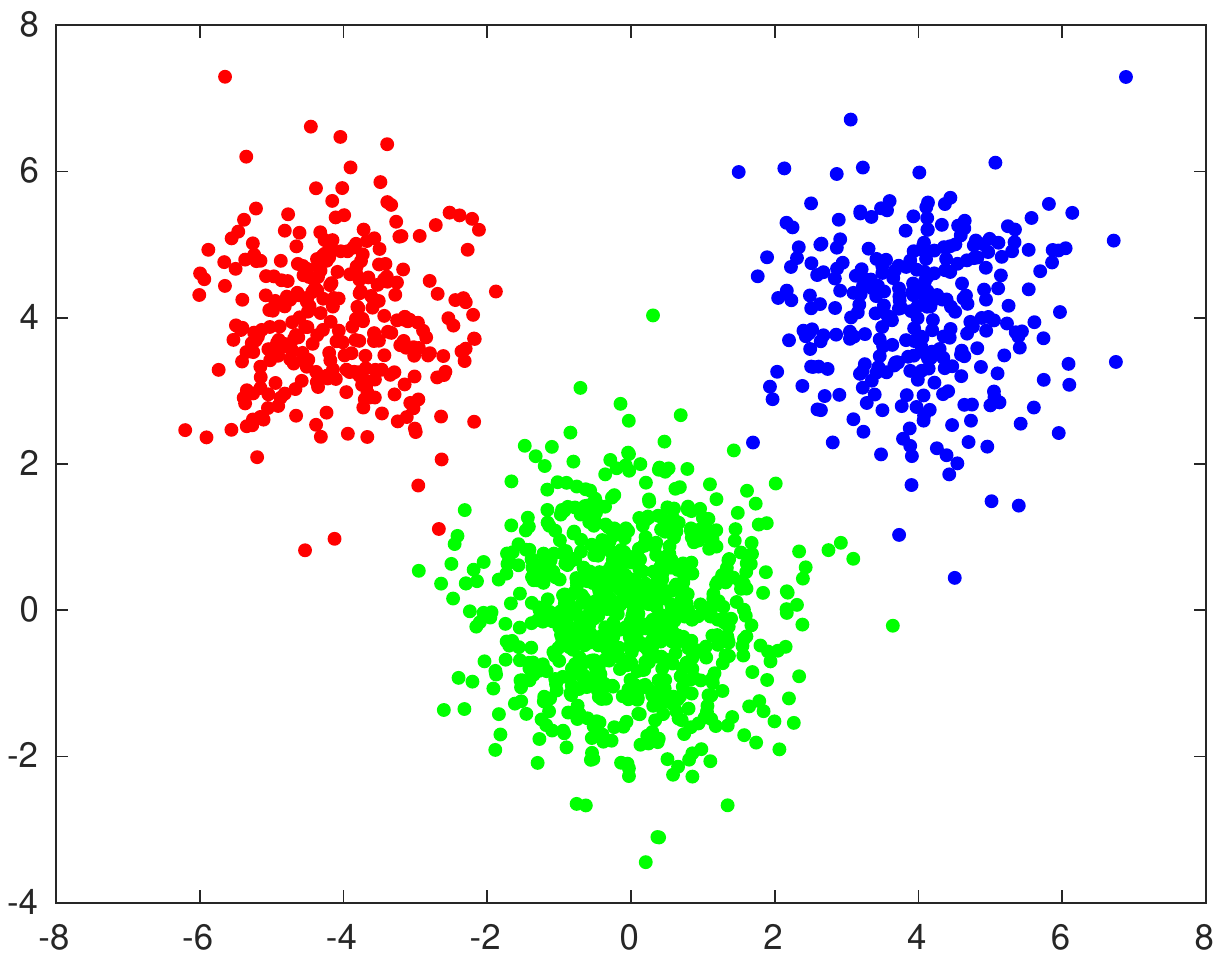}
\end{minipage}
}
\subfloat[ Surrogate $D'$ (blue examples) of $D$  ]{
\label{fig:improved_subfig_b}
\begin{minipage}[t]{0.49\textwidth}
\centering
\includegraphics[width=2.9in,height=2.31in]{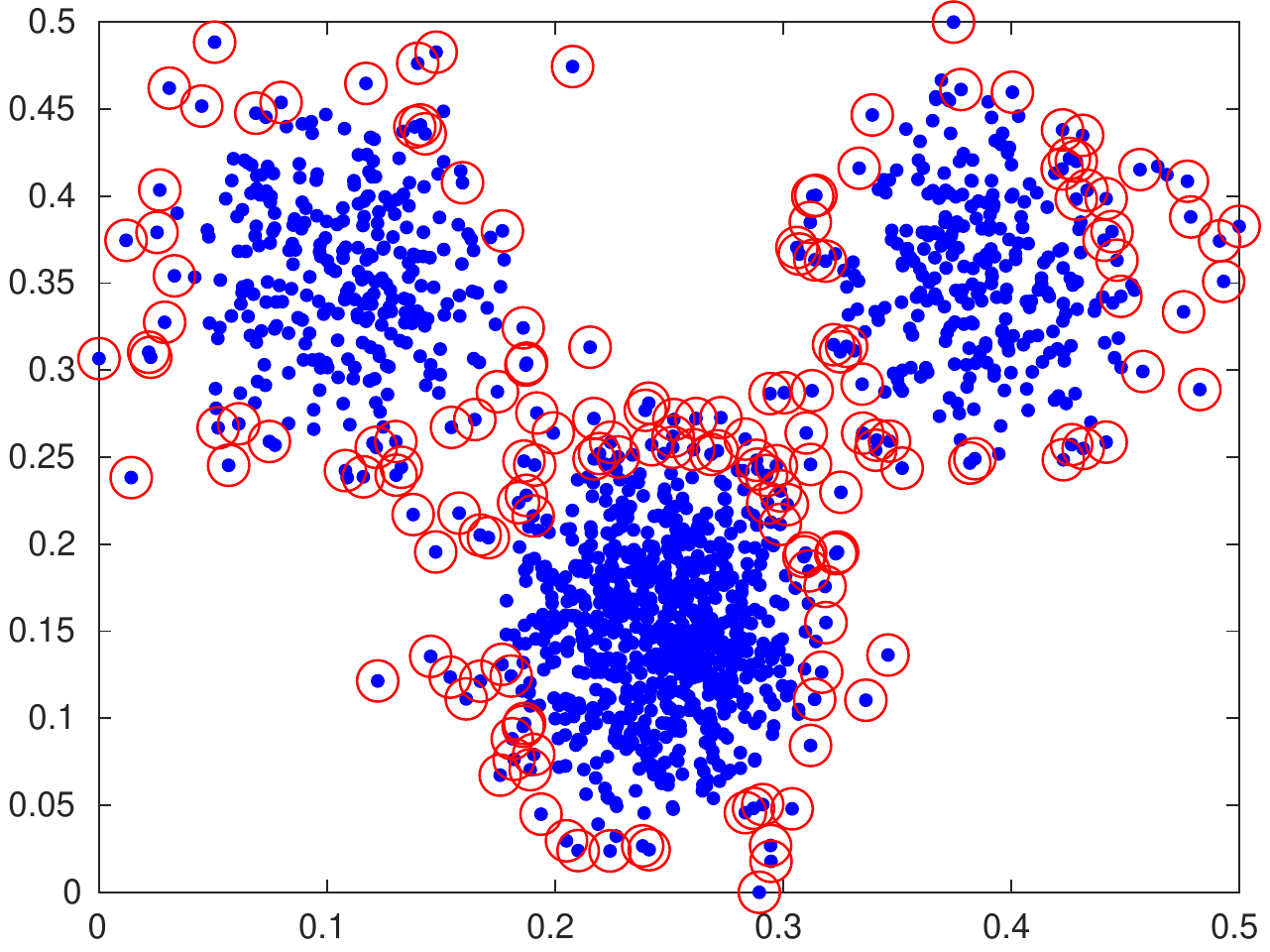}
\end{minipage}
}\\  
\subfloat[$\mathcal{M}(D')=615$]{
\label{fig:improved_subfig_b}
\begin{minipage}[t]{0.32\textwidth}
\centering
\includegraphics[width=1.90in,height=1.51in]{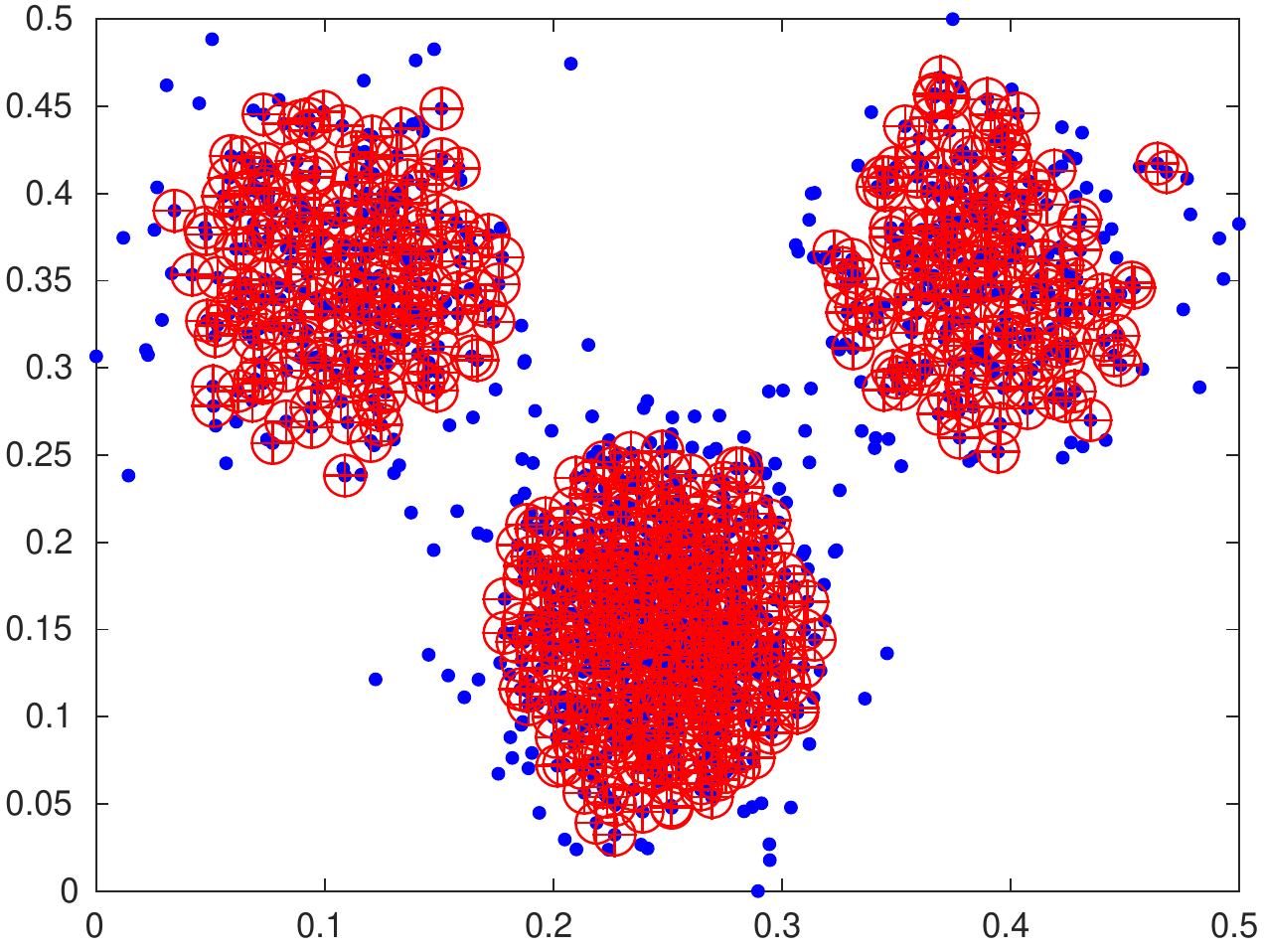}
\end{minipage}
}
\subfloat[$\mathcal{M}(D')=307$]{
\label{fig:improved_subfig_b}
\begin{minipage}[t]{0.32\textwidth}
\centering
\includegraphics[width=1.90in,height=1.51in]{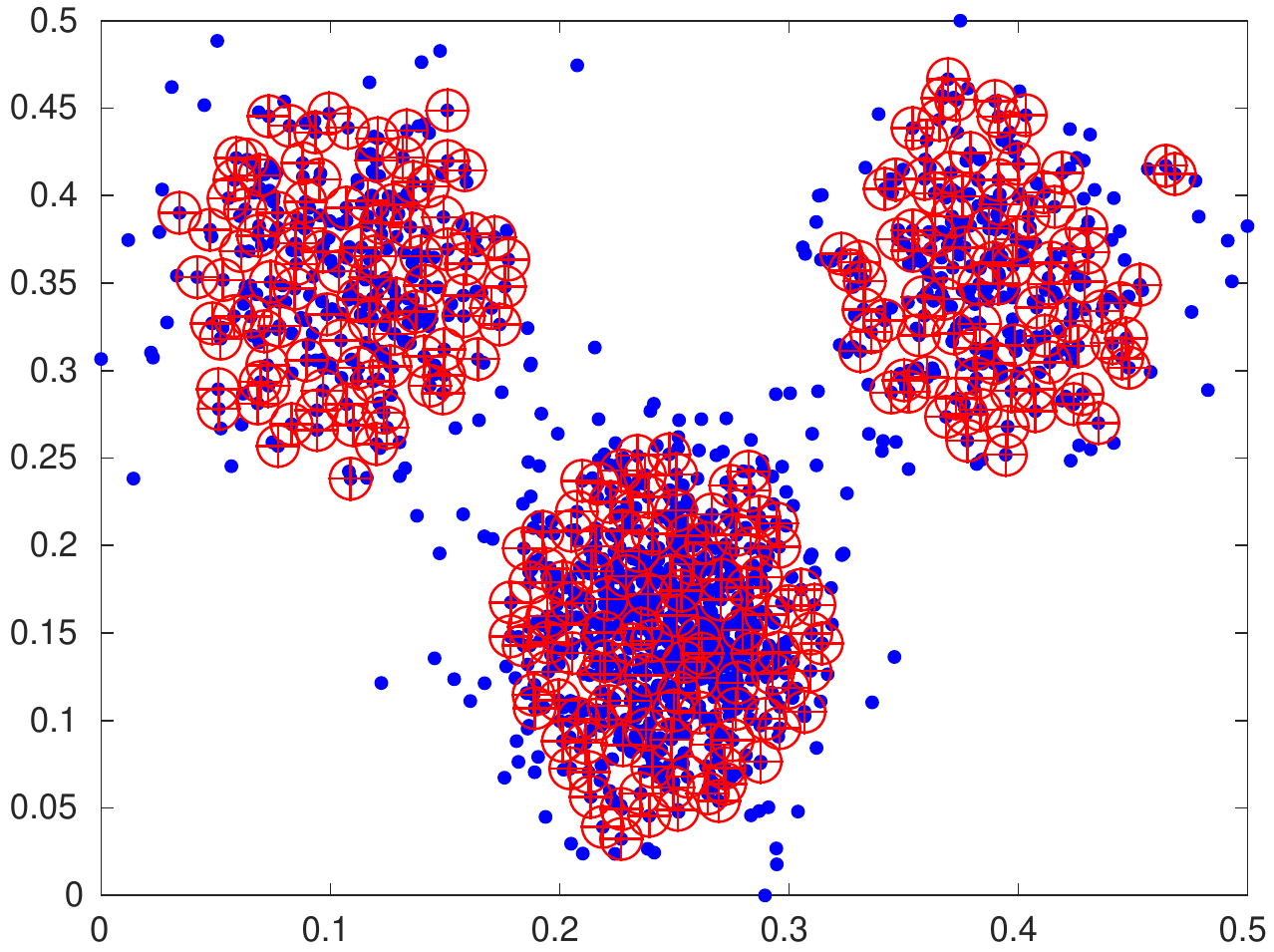}
\end{minipage}
}
\subfloat[$\mathcal{M}(D')=153$]{
\label{fig:improved_subfig_b}
\begin{minipage}[t]{0.32\textwidth}
\centering
\includegraphics[width=1.90in,height=1.51in]{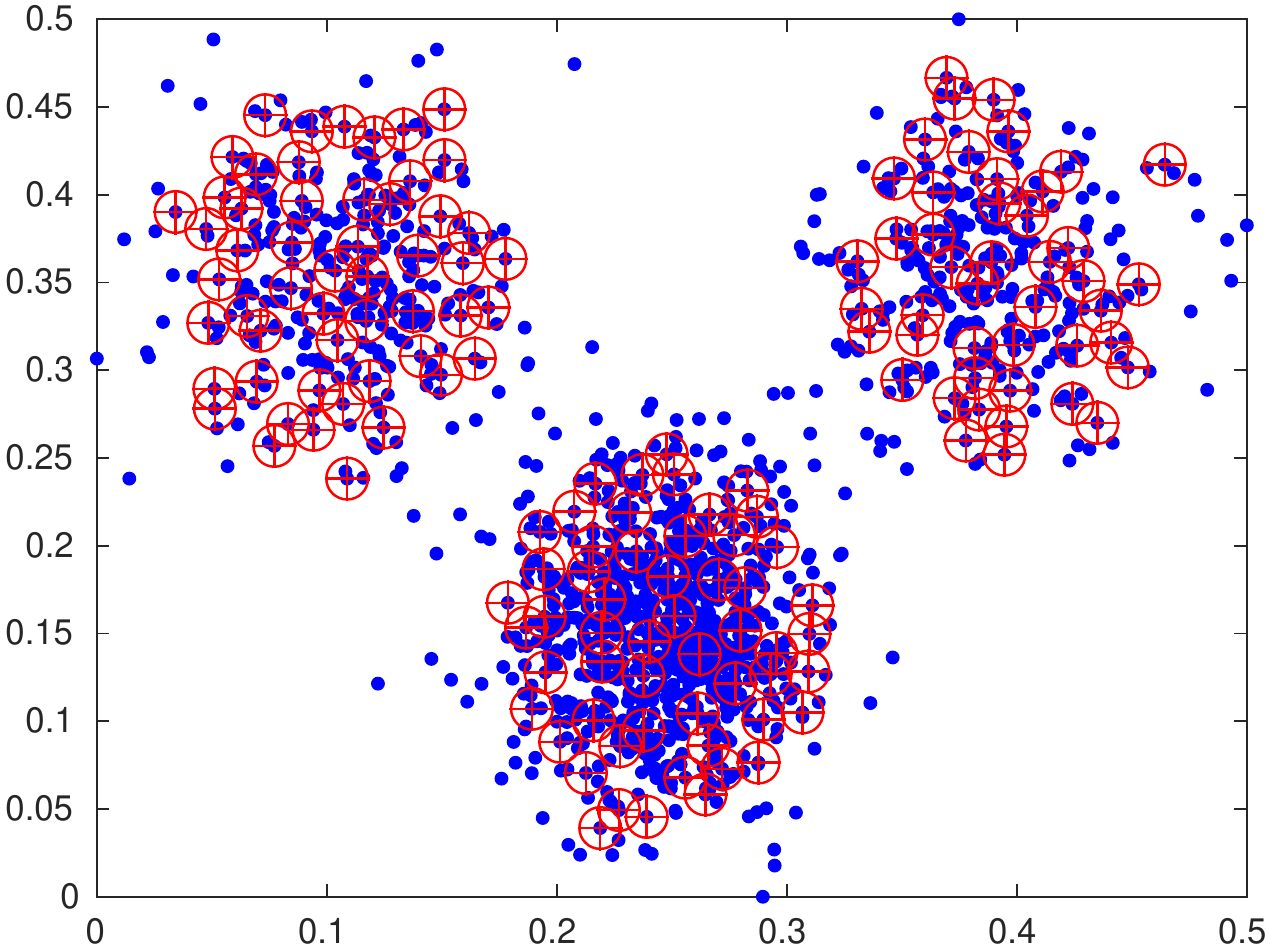}
\end{minipage}
}\\
\subfloat[ $\mathcal{M}(D')=76$ ]{
\label{fig:improved_subfig_b}
\begin{minipage}[t]{0.32\textwidth}
\centering
\includegraphics[width=1.90in,height=1.51in]{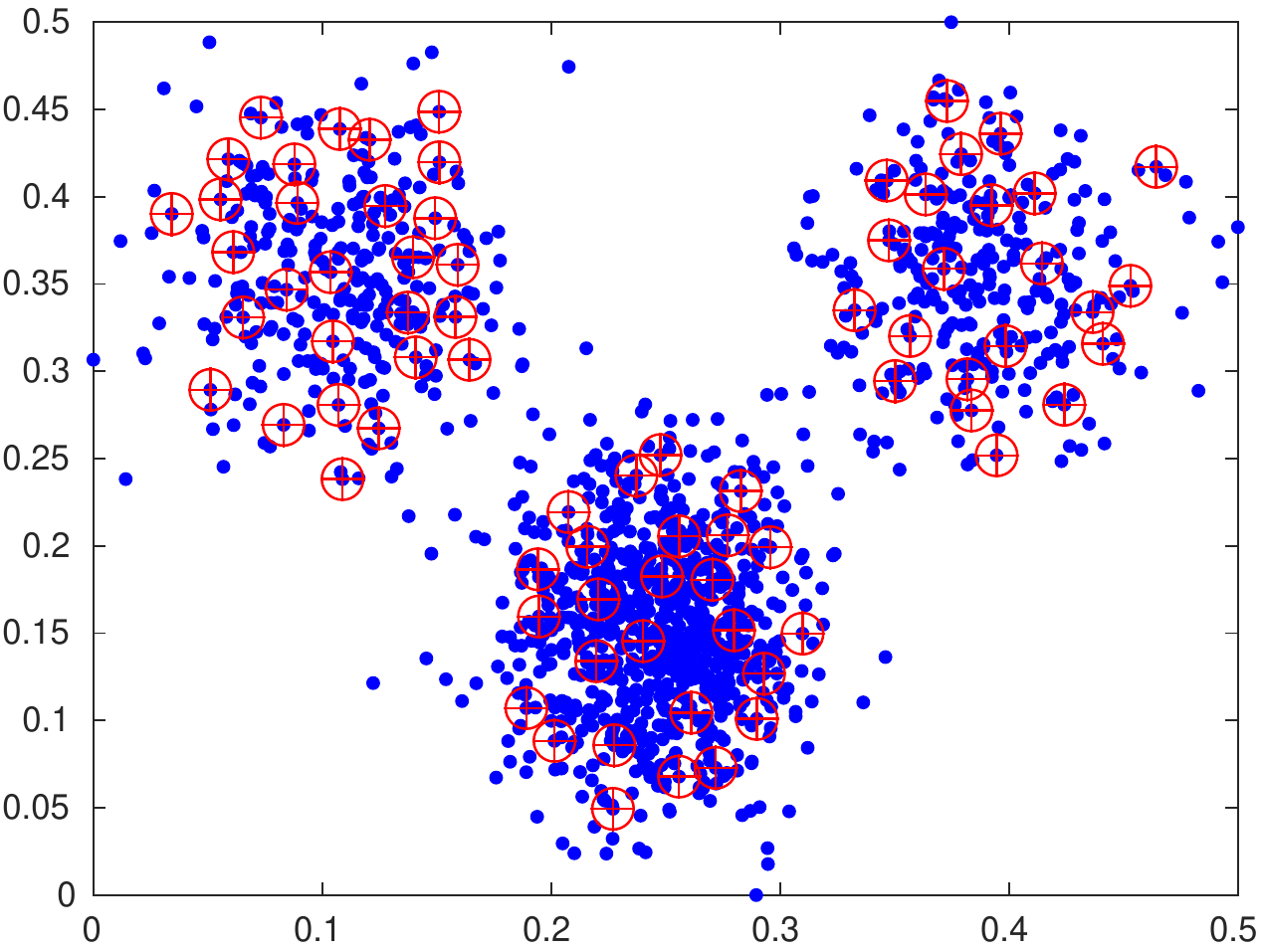}
\end{minipage}
}
\subfloat[$\mathcal{M}(D')=38$]{
\label{fig:improved_subfig_b}
\begin{minipage}[t]{0.32\textwidth}
\centering
\includegraphics[width=1.90in,height=1.51in]{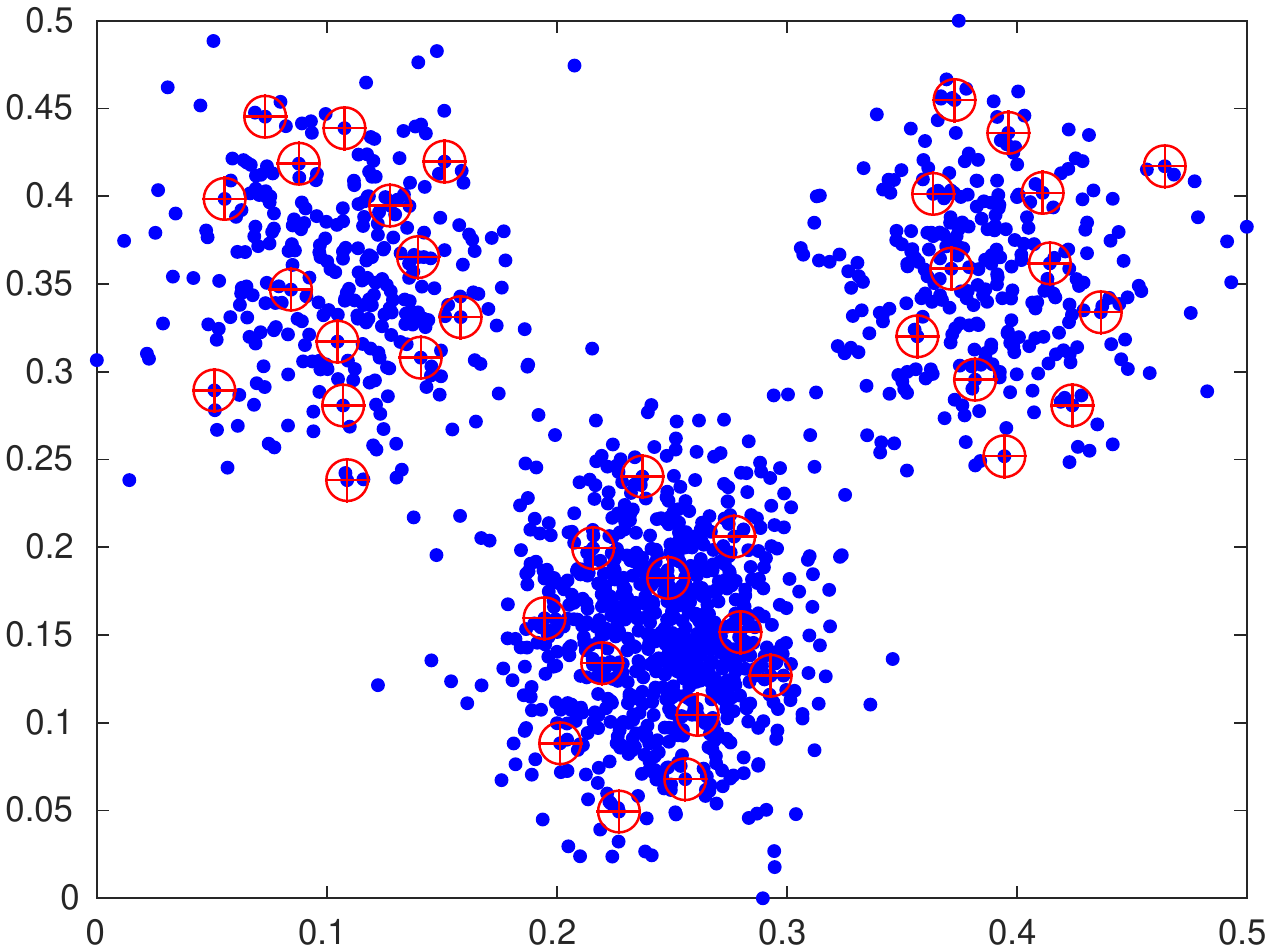}
\end{minipage}
}
\subfloat[$\mathcal{M}(D')=19$]{
\label{fig:improved_subfig_b}
\begin{minipage}[t]{0.32\textwidth}
\centering
\includegraphics[width=1.90in,height=1.51in]{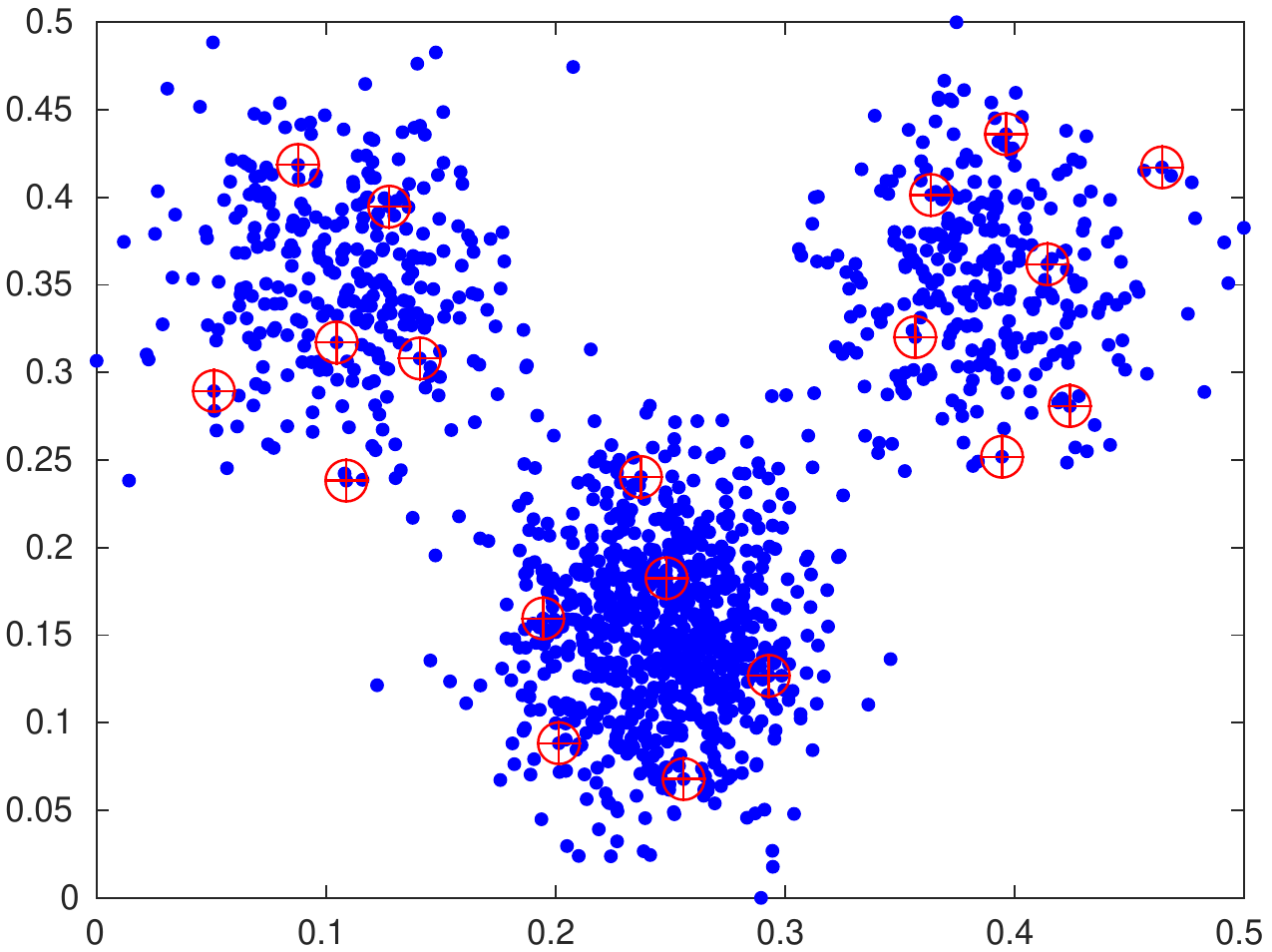}
\end{minipage}
}
\caption{{ 2D visualizations of distribution matching-based machine teaching on a Gaussian dataset $D$. In (b), circled data are boundary examples. In (c) to (h), circled data with `+' are teaching sets with different $\mathcal{M}(D')$}}
\end{figure*}

\subsection{Case Study: Teaching on a 2D Gaussian Dataset}
Figure 3 presents a case study of distribution matching-based machine teaching on a 2D Gaussian dataset. The 2D visualizations dynamically show  the iterative halving process on surrogate $D'$: 1) Figure 3(a) draws the full training Gaussian data $D$, where $\mathcal{M}(D)=1400$; 2) Figure 3(b) draws the surrogate of $D$, where the circled 170 data are boundary examples, the remaining blue points are the data of $D'$, and the parameter settings are $r=0.4$, $\eta=1.0e-4$; 3) Figures 3(c) to 3(h) show  the iterative halving process, where $\mathcal{M}(D')$ is continuously halved. The presented teaching sets with different $\mathcal{M}(D') $ properly match the distribution of $D$ without noisy perturbations around boundary. 

\par Specifically, all the teaching examples are distributed inside the clusters with high densities due to the smooth boundary of the surrogate  $D'$ (w.r.t. Lines 1 to 4 of Algorithm 1). The iterative halving (w.r.t. Lines 5 to 12) is performed on the last update of $D'$, which keeps consistent distribution properties as its previous. Therefore, all the teaching sets with different $\mathcal{M}(D') $ yield consistent distributions as the original distribution of $D$.

\begin{figure*}[!t]
\subfloat[\emph{Adult}]{
\label{fig:improved_subfig_b}
\begin{minipage}[t]{0.49\textwidth}
\centering
\includegraphics[width=2.9in,height=2.31in]{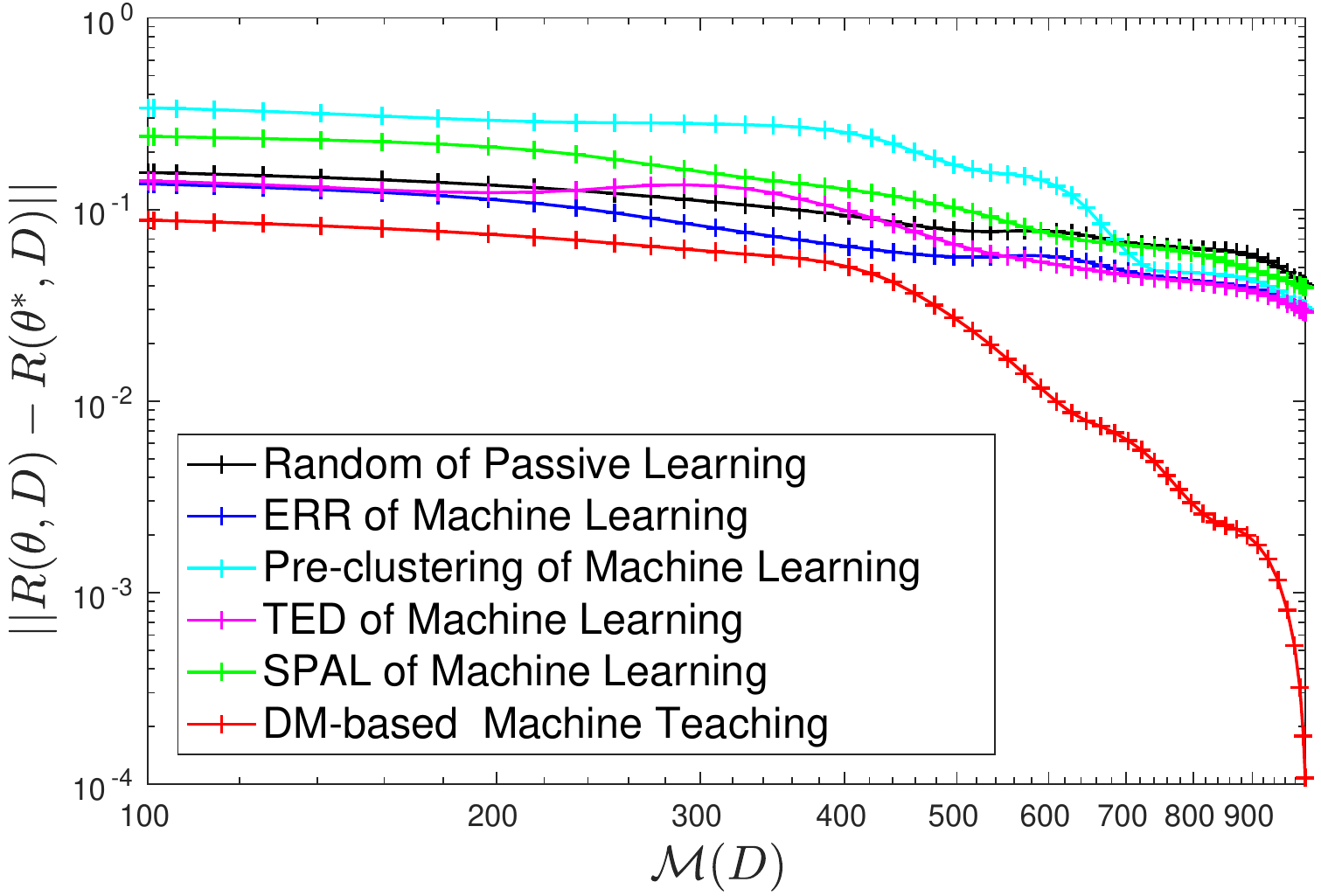}
\end{minipage}
}
\subfloat[ \emph{Phishing}  ]{
\label{fig:improved_subfig_b}
\begin{minipage}[t]{0.49\textwidth}
\centering
\includegraphics[width=2.68in,height=2.34in]{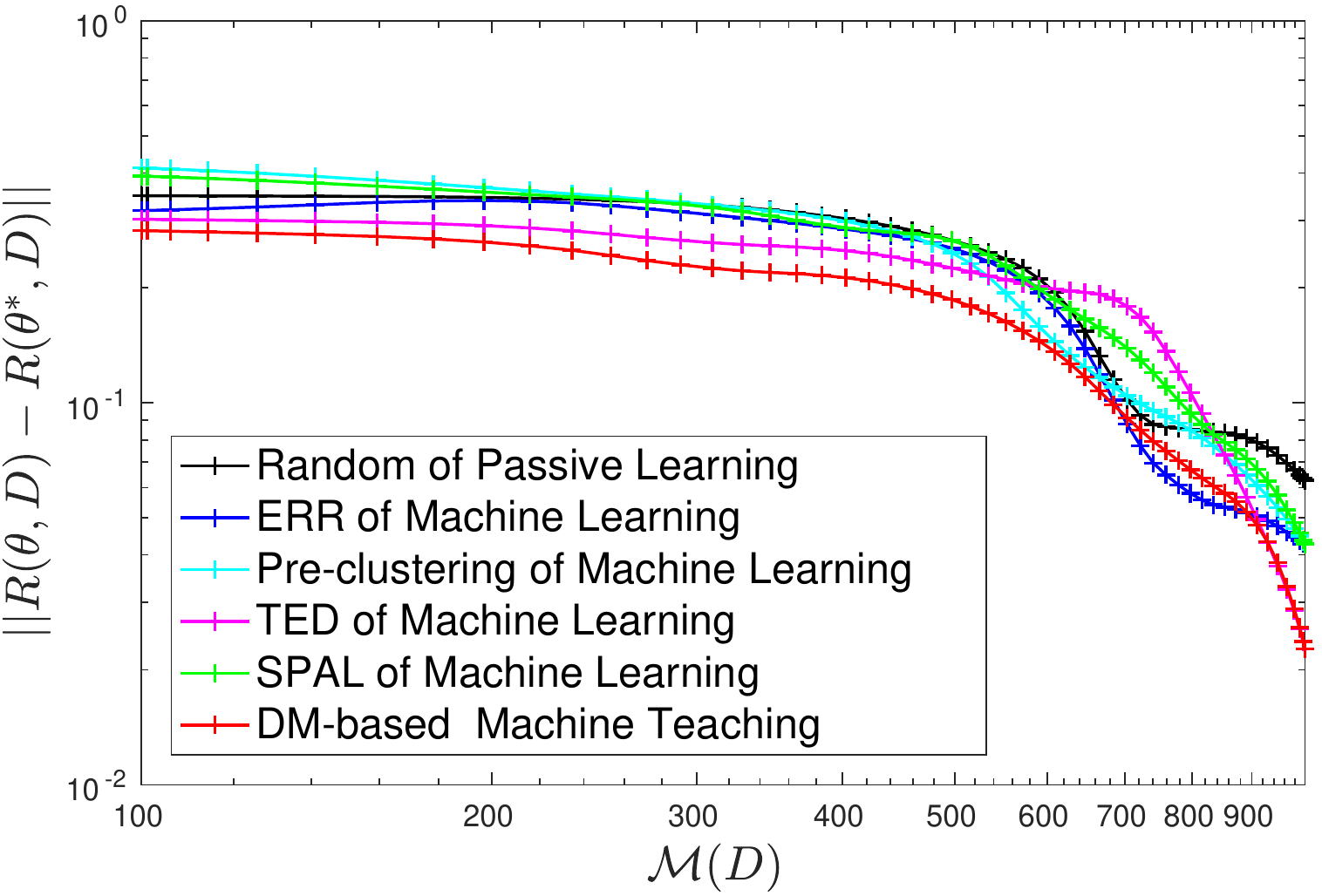}
\end{minipage}
}\\
\subfloat[\emph{Satimage}]{
\label{fig:improved_subfig_b}
\begin{minipage}[t]{0.49\textwidth}
\centering
\includegraphics[width=2.9in,height=2.31in]{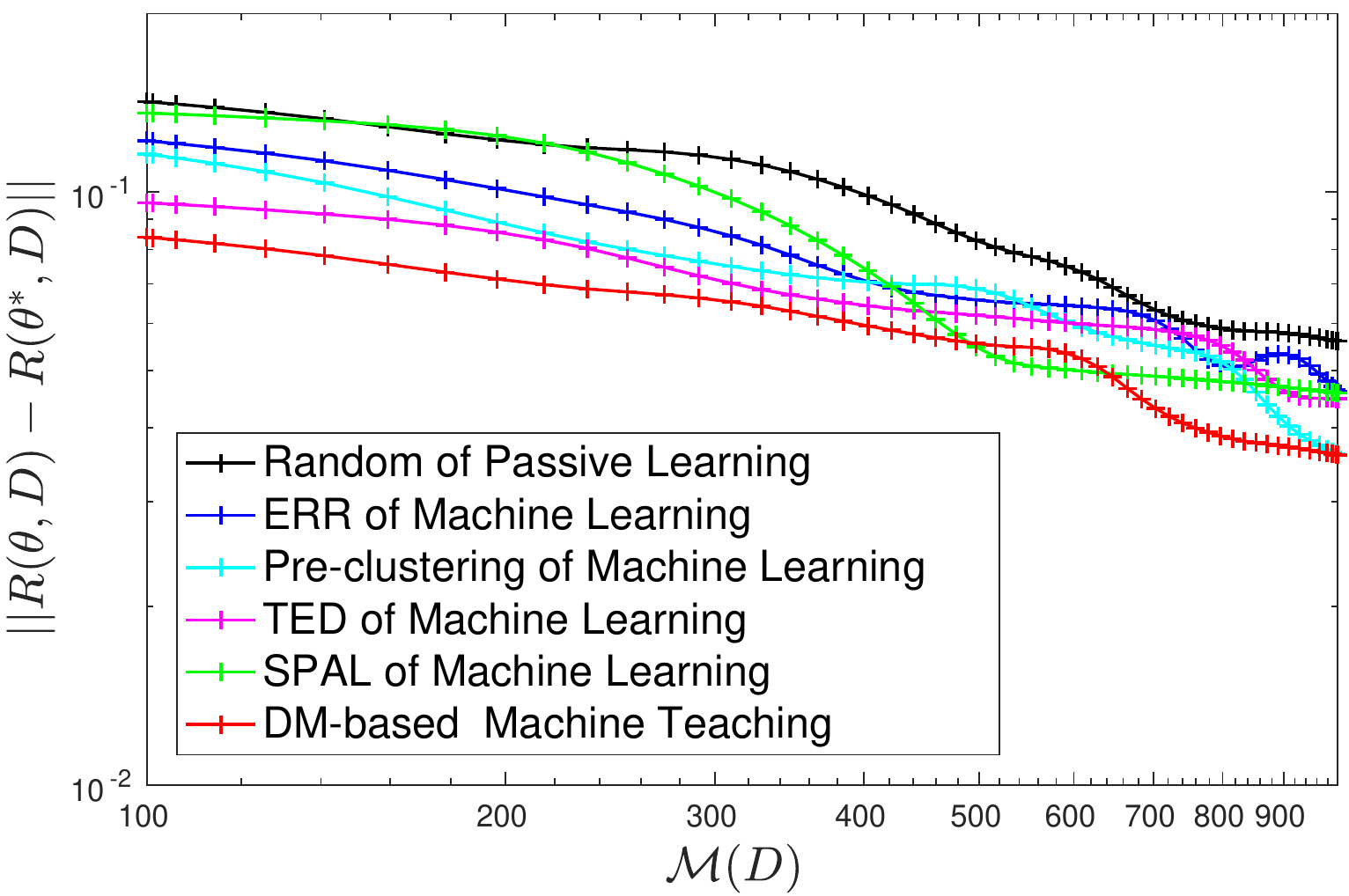}
\end{minipage}
}
\subfloat[\emph{MNIST}]{
\label{fig:improved_subfig_b}
\begin{minipage}[t]{0.49\textwidth}
\centering
\includegraphics[width=2.68in,height=2.34in]{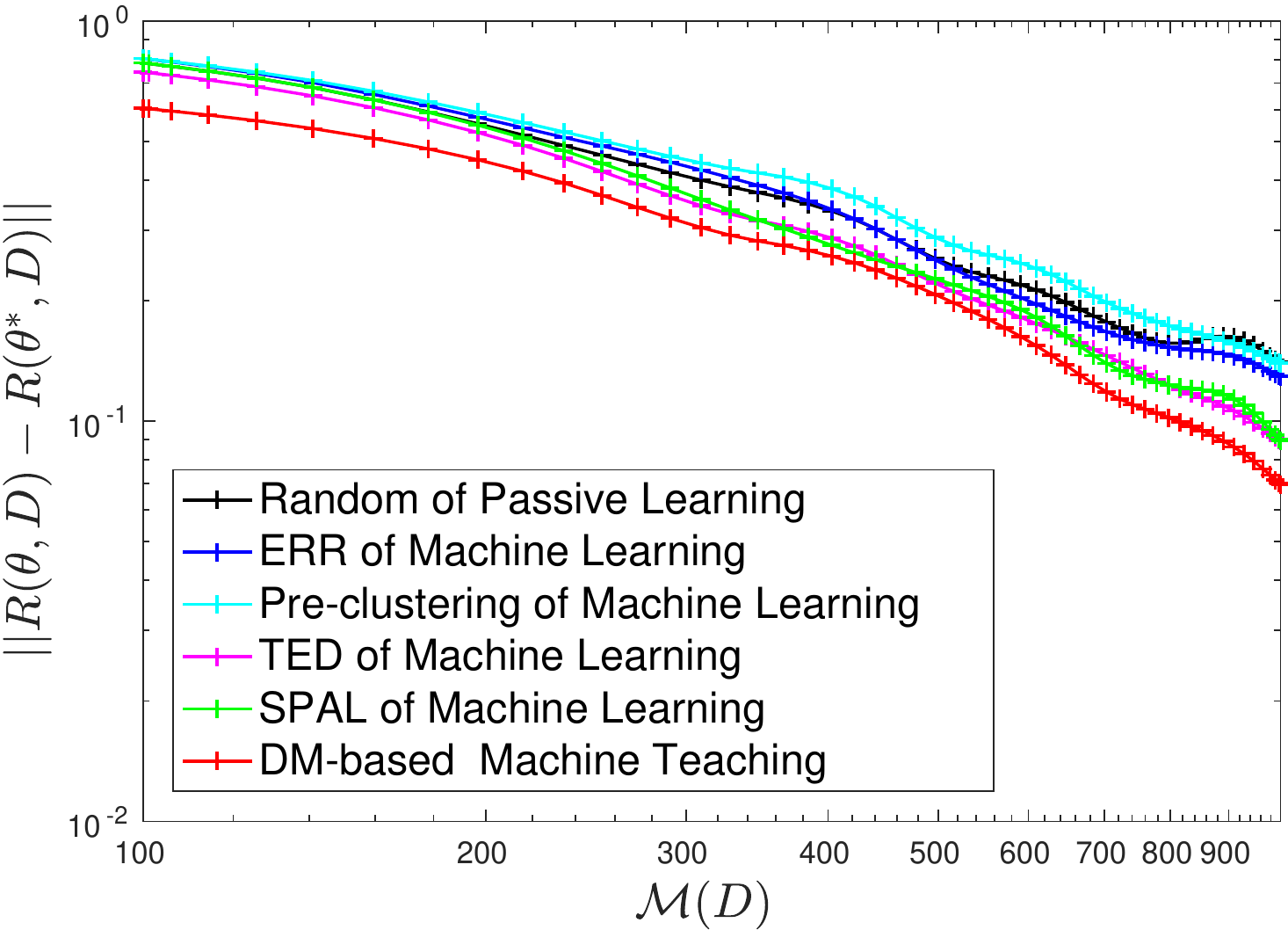}
\end{minipage}
}

\caption{Regulating $\mathcal{M}(D)$ to minimize $\|R(\theta,D)-R(\theta^*,D)\|$   on \emph{Adult, Phishing, Satimage}, and \emph{MNIST} data sets. }
\end{figure*}

\subsection{Regulating $\mathcal{M}(D)$ to Minimize $\|R(\theta,D)-R(\theta^*,D) \|$}
Regulating $\mathcal{M}(D)$  is  important for  both machine teacher and student learners due to over-fitting or computational  overhead.  	With Assumption 3,  the goal  of machine teaching is to minimize $\|R(\theta,D)-R(\theta^*,D) \|$, where $\theta^*$ is with respect to $D$.  The experimental datasets are \emph{Adult, Phishing, Satimage,} and the \emph{MNIST}. We assume that $h$ is generated from a SVM classifier with a RBF kernel.  That means, $R(\theta^*,D)=0.1200$ on \emph{Adult}, $R(\theta^*,D)=0.1141$ on \emph{Phishing},  $R(\theta^*,D)=0.0875$ on \emph{Satimage}, and $R(\theta^*,D)=0.0009$ on \emph{MNIST}, where each  $h^*$ is over the full training data.

\par The compared machine learning baselines are typical active learning algorithms including ERR,  Pre-clustering,  TED and SPAL, where ERR maximizes the expected error reduction over a SVM classifier,  Pre-clustering employs   the Hierarchical clustering  and the pruning budget is set from 100 to 1000 with a step of 100,  TED  uses a hyperparameter $\sigma$=1.8 (kernel bandwidth parameter) to generate the kernel matrix and vary $\lambda$ (kernel ridge regression) from 0.01 to 1 with a step of 0.01, and SPAL sets the paced learning parameter from 0.01 to 1 with a step of 0.01, etc. 
   Before running those  machine learning baselines, we randomly select 10 data  from each dataset to train an  initial hypothesis for them.  For our distribution matching-based machine teaching algorithm,   $\mathcal{M}(D')=n'$, $n'/n \in [0.85, 0.95], $ $r=[0.1, 0.5]$ $\eta=10e-4$, and $l$ is constrained by $\mathcal{M}(D')$ that satisfies $\frac{n'}{2^l}\geq \mathcal{M}(D')$. 

\par Figure 4 draws the learning curves of regulating  $\mathcal{M}(D)$   to minimize  $\|R(\theta,D)-R(\theta^*,D) \|$ into an expected risk across the best parameter candidates of each baseline.
 From  the test results in Figure  4, we  find that machine teaching algorithm can regulate $\mathcal{M}(D)$  better than the machine learning baselines, i.e. spend fewer training data to obtain an expected learning risk. Especially at the beginning of those curves, $\mathcal{M}(D)$ of machine teaching is much smaller than that of the  machine learning baselines.  
\begin{figure*}[!t]
\subfloat[\emph{Adult}]{
\label{fig:improved_subfig_b}
\begin{minipage}[t]{0.49\textwidth}
\centering
\includegraphics[width=2.9in,height=2.31in]{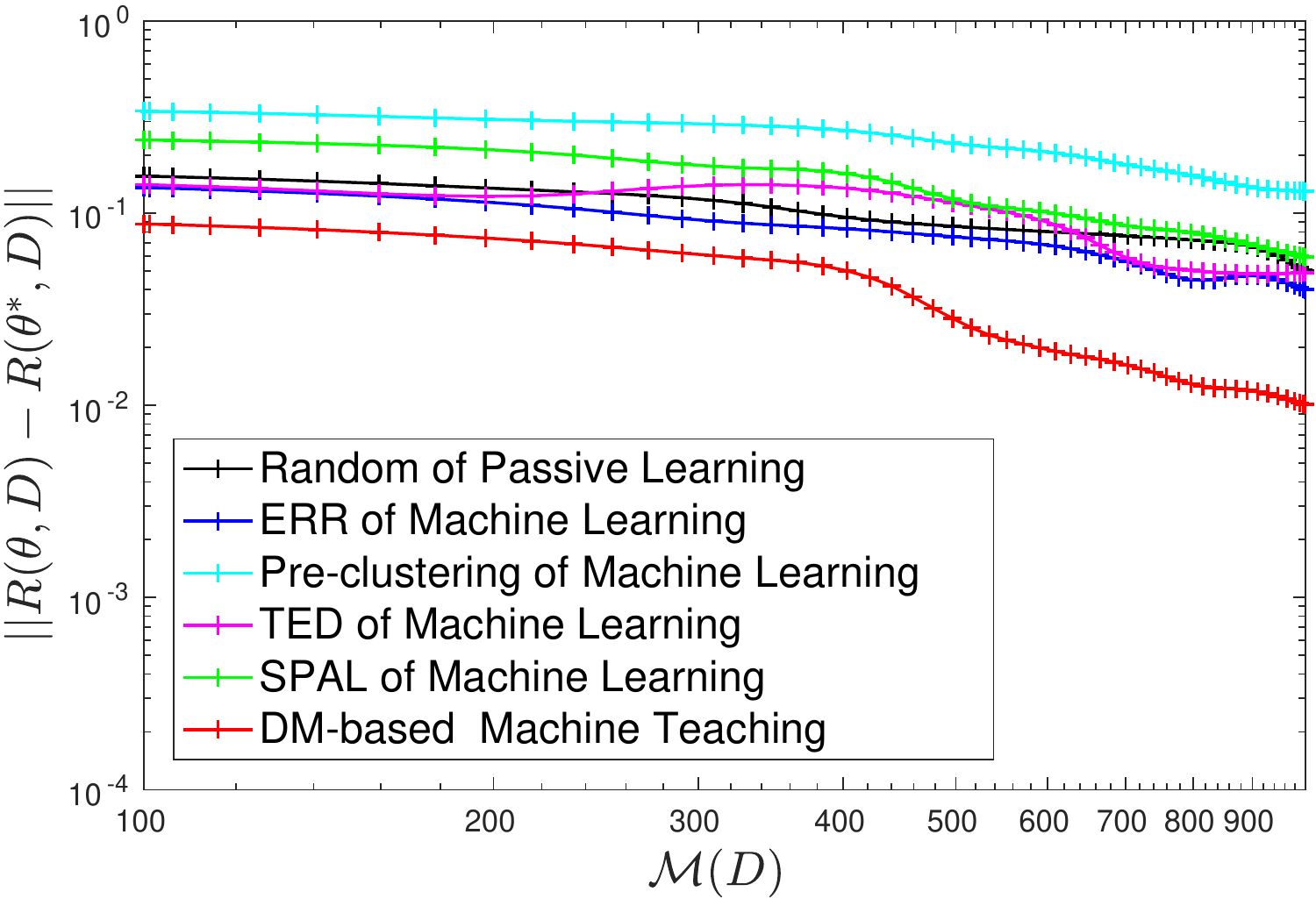}
\end{minipage}
}
\subfloat[ \emph{Phishing}  ]{
\label{fig:improved_subfig_b}
\begin{minipage}[t]{0.49\textwidth}
\centering
\includegraphics[width=2.9in,height=2.31in]{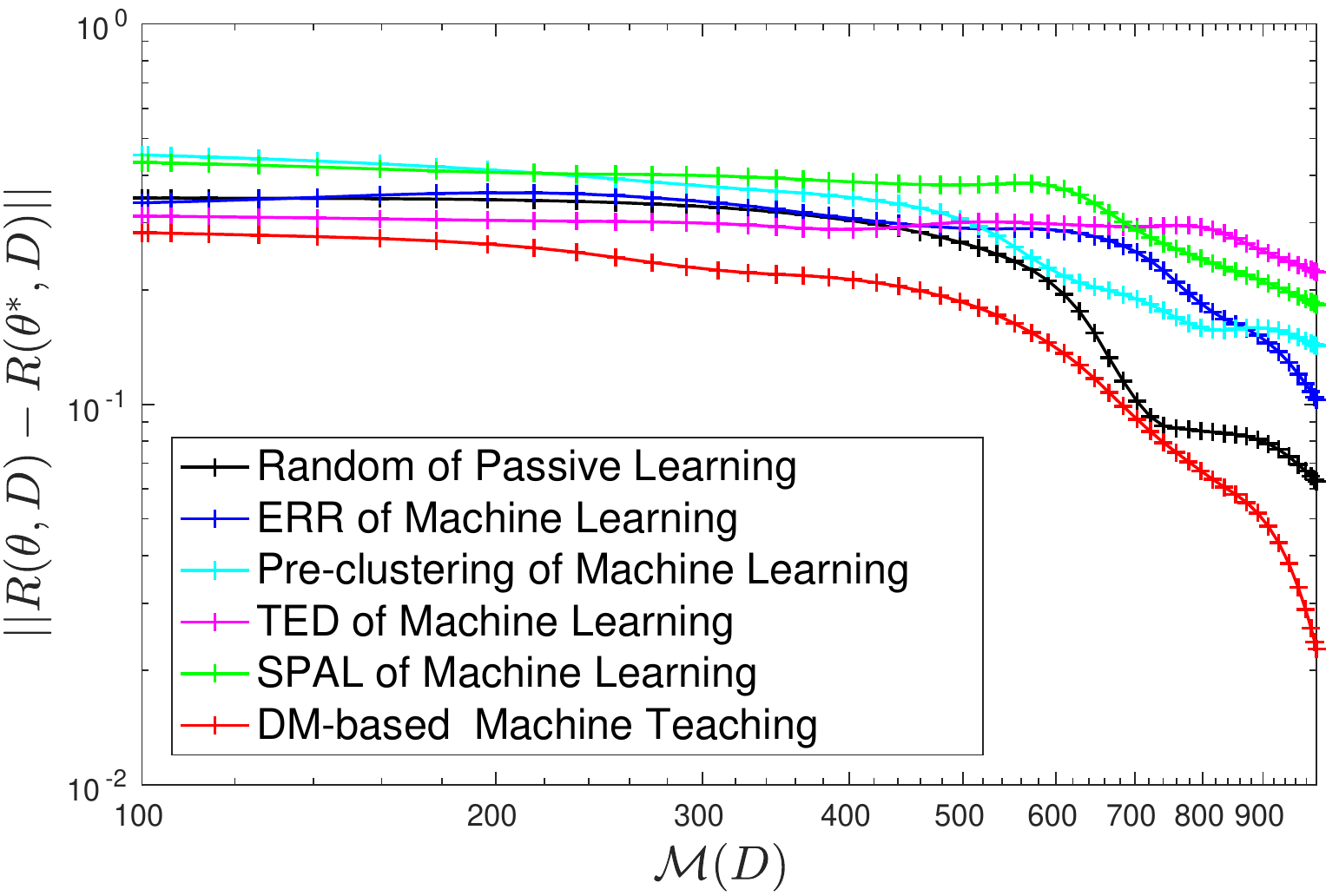}
\end{minipage}
}\\
\subfloat[\emph{Satimage}]{
\label{fig:improved_subfig_b}
\begin{minipage}[t]{0.49\textwidth}
\centering
\includegraphics[width=2.9in,height=2.31in]{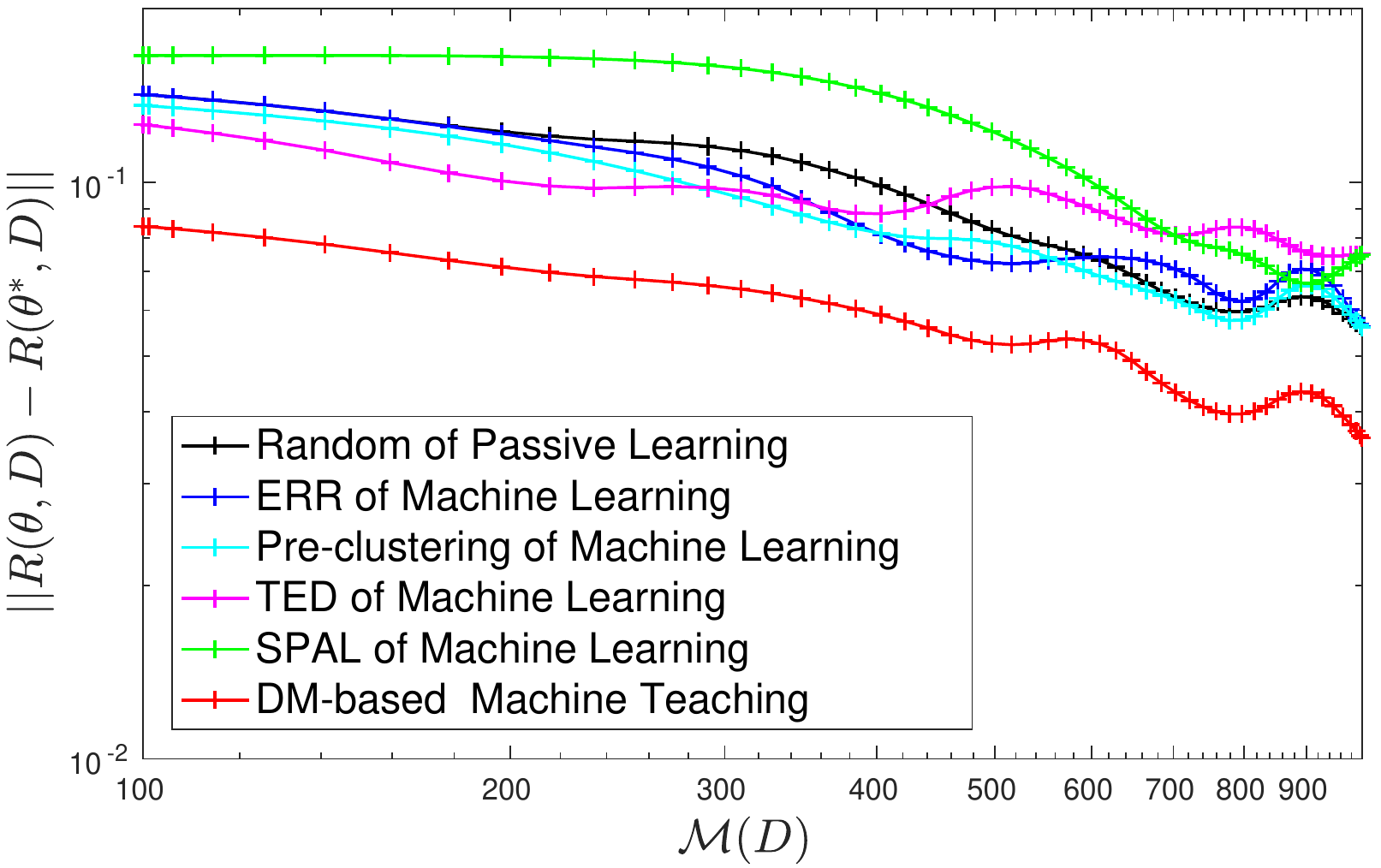}
\end{minipage}
}
\subfloat[\emph{MNIST}]{
\label{fig:improved_subfig_b}
\begin{minipage}[t]{0.49\textwidth}
\centering
\includegraphics[width=2.9in,height=2.31in]{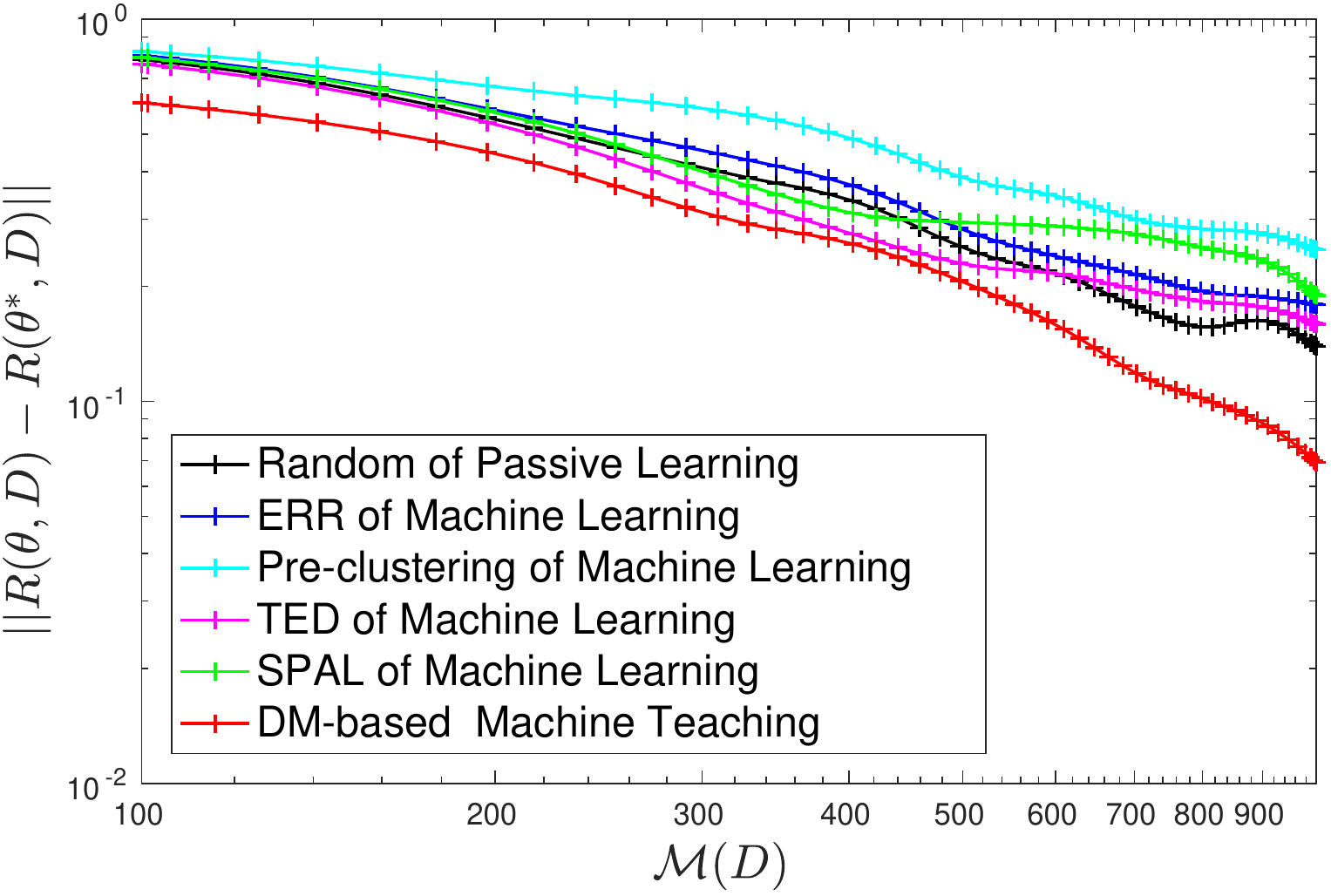}
\end{minipage}
}

\caption{ {Reducing  the search  space of  $\mathbbm{D}$ to minimize $\|R(\theta,D)-R(\theta^*,D)\|$  on \emph{Adult, Phishing, Satimage}, and \emph{MNIST} data sets.} } 
\end{figure*}
\subsection{Reducing the Search Space of $\mathbbm{D}$}
Reducing  the search space of $\mathbbm{D}$ can relieve the minimization costs of Eqs.~(1) and  (2) because estimating one data whether can be picked up as a teaching example needs to access the whole unlabeled data pool.   If the teacher needs to give feedback for the learner in a limited budge cost e.g. time and space, the teaching algorithm must help the teacher make a decision on  which example should be selected.

\par  Given an access budget of $\lceil n/2 \rceil$ to the unlabeled data pool one time, the   machine learning and teaching algorithms have to return one best candidate teaching example. Figure 5 draws the $\mathcal{M}(D)$ curves by progressively minimizing $\|R(\theta,D)-R(\theta^*,D) \|$. Compared to the  results in Figure 4 with a full access budget, the  $\mathcal{M}(D)$ of machine learning baselines arise rapidly due to the greedy updates on $\theta$ to $\hat \theta$ can not always be the optimal. This forces the learning algorithm  to request more data to reach an expected learning  risk. However, our machine teaching algorithm backward and iteratively halves  $\theta^*$ to $\hat \theta$ without greedy search in $D$, thereby 
 lower perturbations to the limited  access budget are presented. Therefore,  the generalized distribution matching-based machine teaching algorithm could trust  a black-box student learner with inestimable teaching loss in real teaching tasks.

\begin{figure*}[!t]
\subfloat[\emph{Adult}]{
\label{fig:improved_subfig_b}
\begin{minipage}[t]{0.49\textwidth}
\centering
\includegraphics[width=2.9in,height=2.31in]{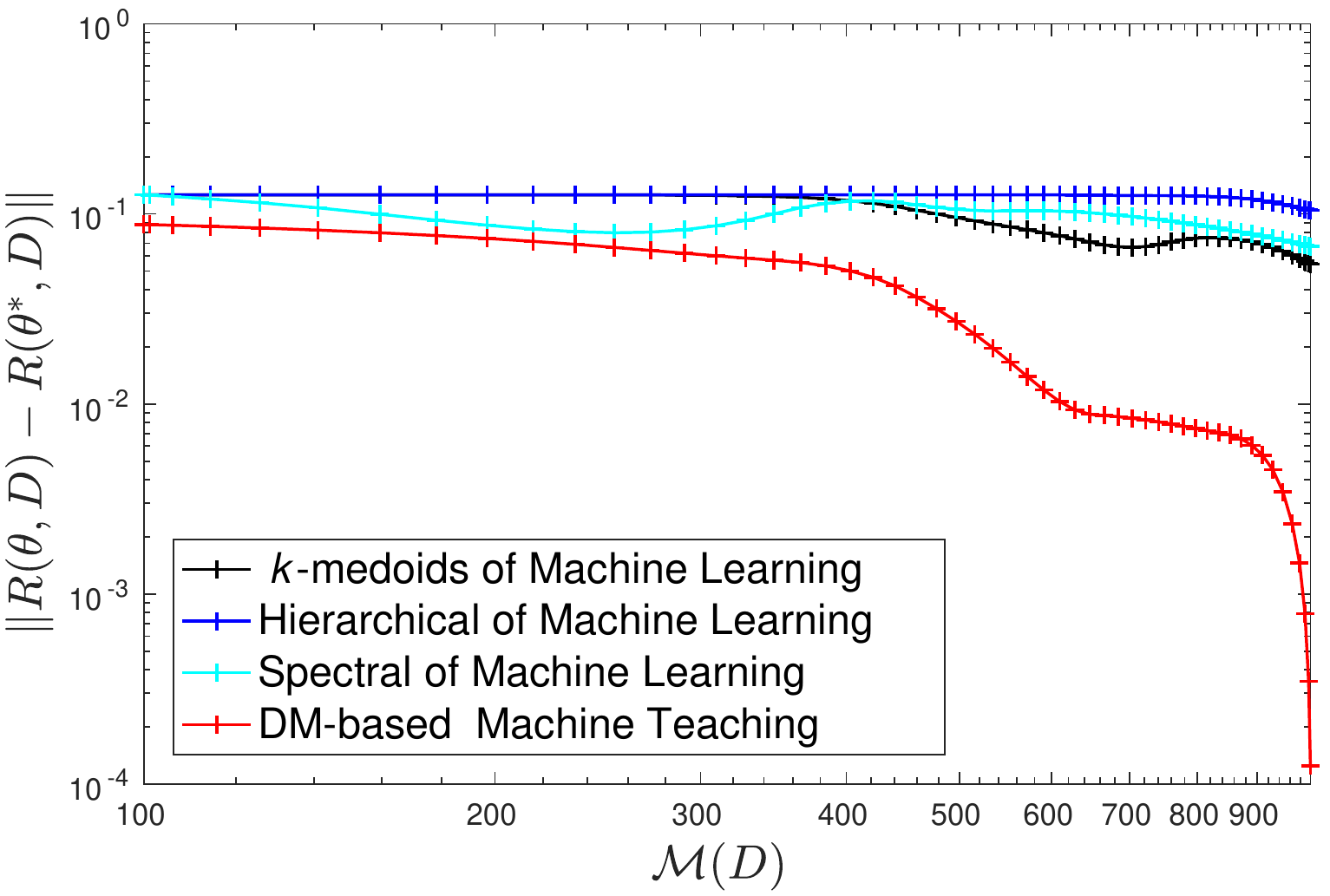}
\end{minipage}
}
\subfloat[ \emph{Phishing}  ]{
\label{fig:improved_subfig_b}
\begin{minipage}[t]{0.49\textwidth}
\centering
\includegraphics[width=2.9in,height=2.31in]{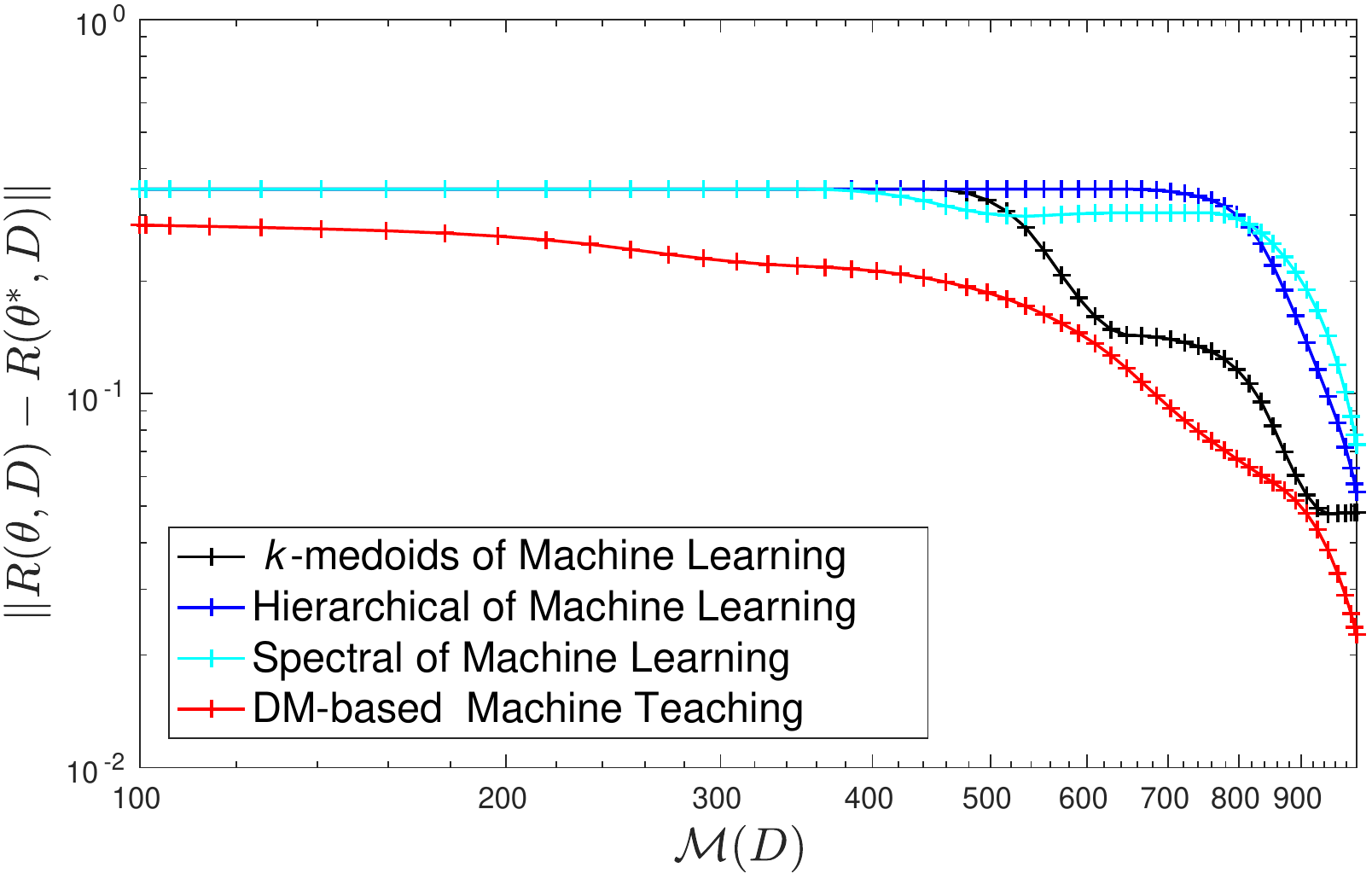}
\end{minipage}
}\\
\subfloat[\emph{Satimage}]{
\label{fig:improved_subfig_b}
\begin{minipage}[t]{0.49\textwidth}
\centering
\includegraphics[width=2.9in,height=2.31in]{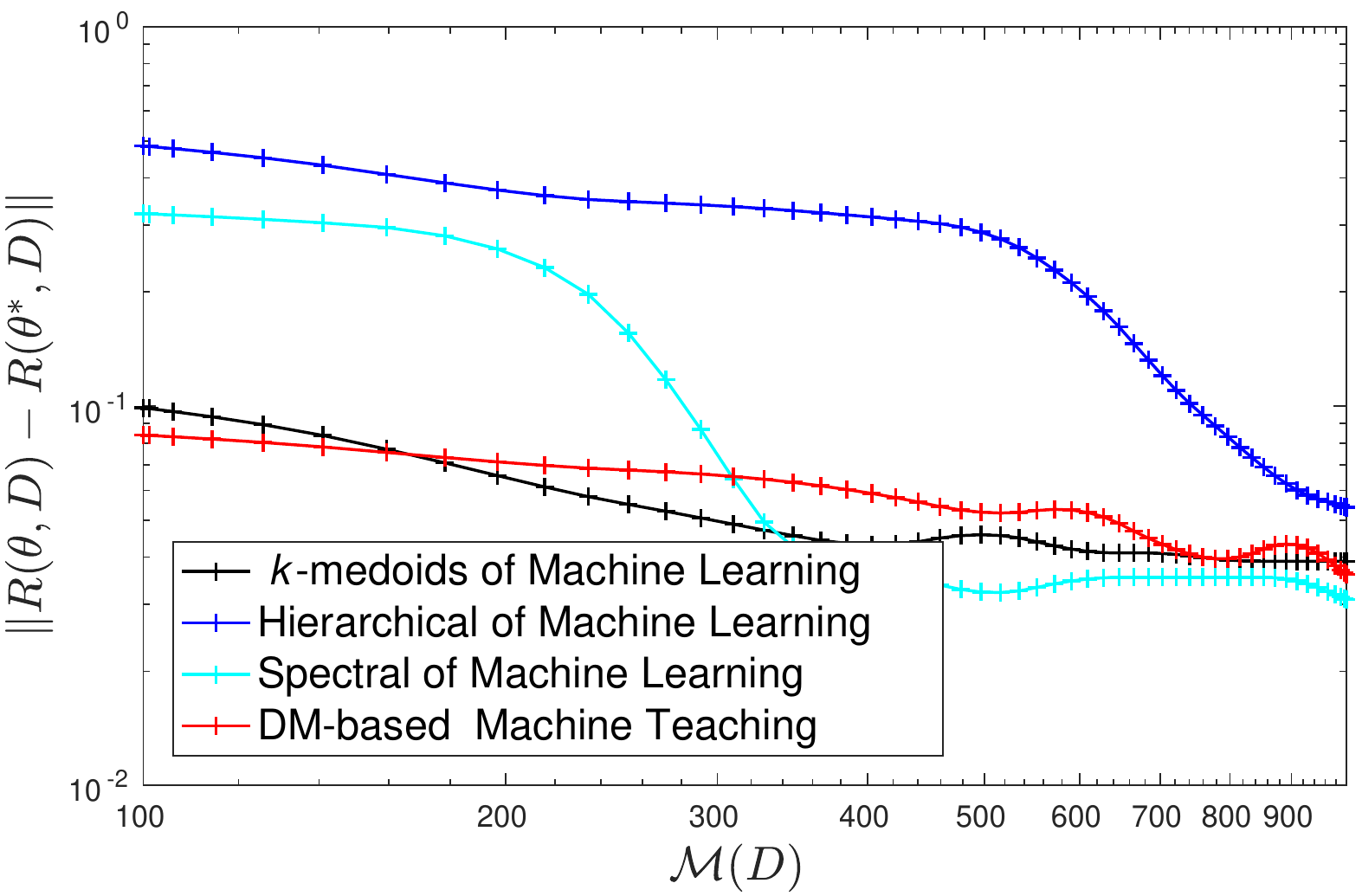}
\end{minipage}
}
\subfloat[\emph{MNIST}]{
\label{fig:improved_subfig_b}
\begin{minipage}[t]{0.49\textwidth}
\centering
\includegraphics[width=2.9in,height=2.31in]{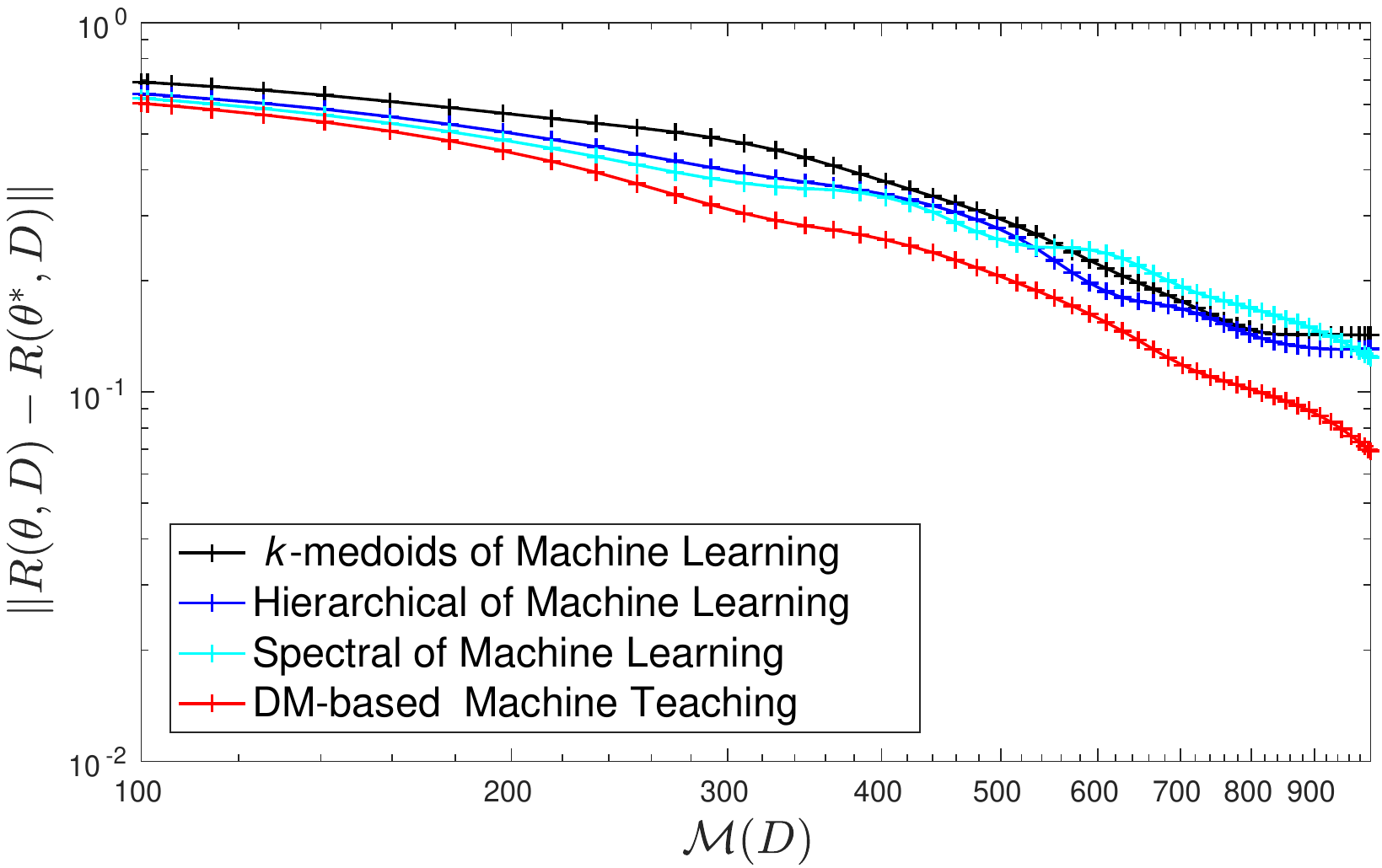}
\end{minipage}
}

\caption{{Minimizing $R(\theta^*,D)-R(\theta,D)$  with quantitative  $\mathcal{M}(D)$, i.e. unsupervised machine learning.  } }
\end{figure*}

\begin{figure*}[!t]
\subfloat[$\mathcal{M}(D')$=500]{
\label{fig:improved_subfig_b}
\begin{minipage}[t]{0.49\textwidth}
\centering
\includegraphics[width=2.9in,height=2.31in]{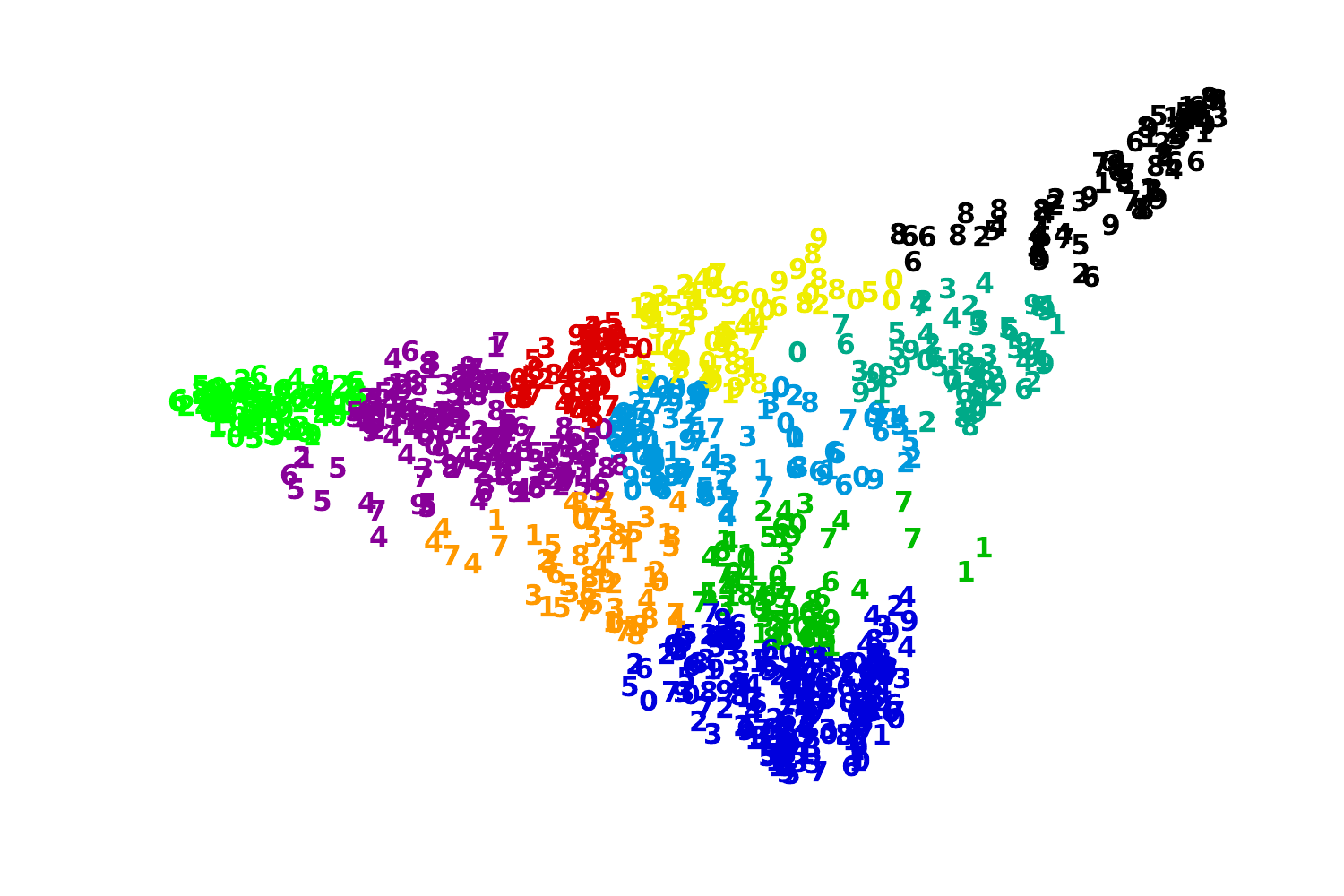}
\end{minipage}
}
\subfloat[$\mathcal{M}(D')$=1,000]{
\label{fig:improved_subfig_b}
\begin{minipage}[t]{0.49\textwidth}
\centering
\includegraphics[width=2.9in,height=2.31in]{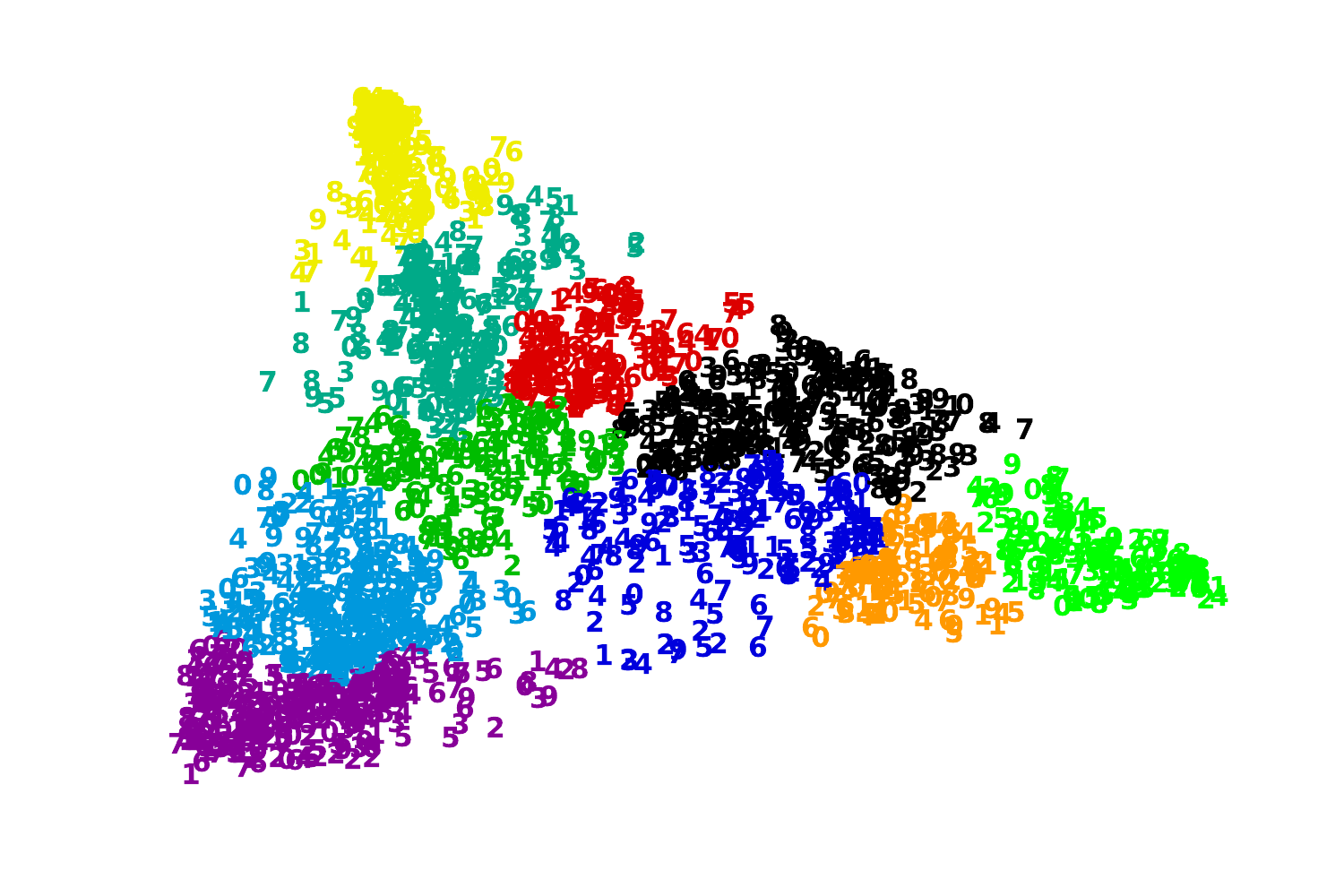}
\end{minipage}
}\\
\subfloat[$\mathcal{M}(D')$=2,000]{
\label{fig:improved_subfig_b}
\begin{minipage}[t]{0.49\textwidth}
\centering
\includegraphics[width=2.9in,height=2.31in]{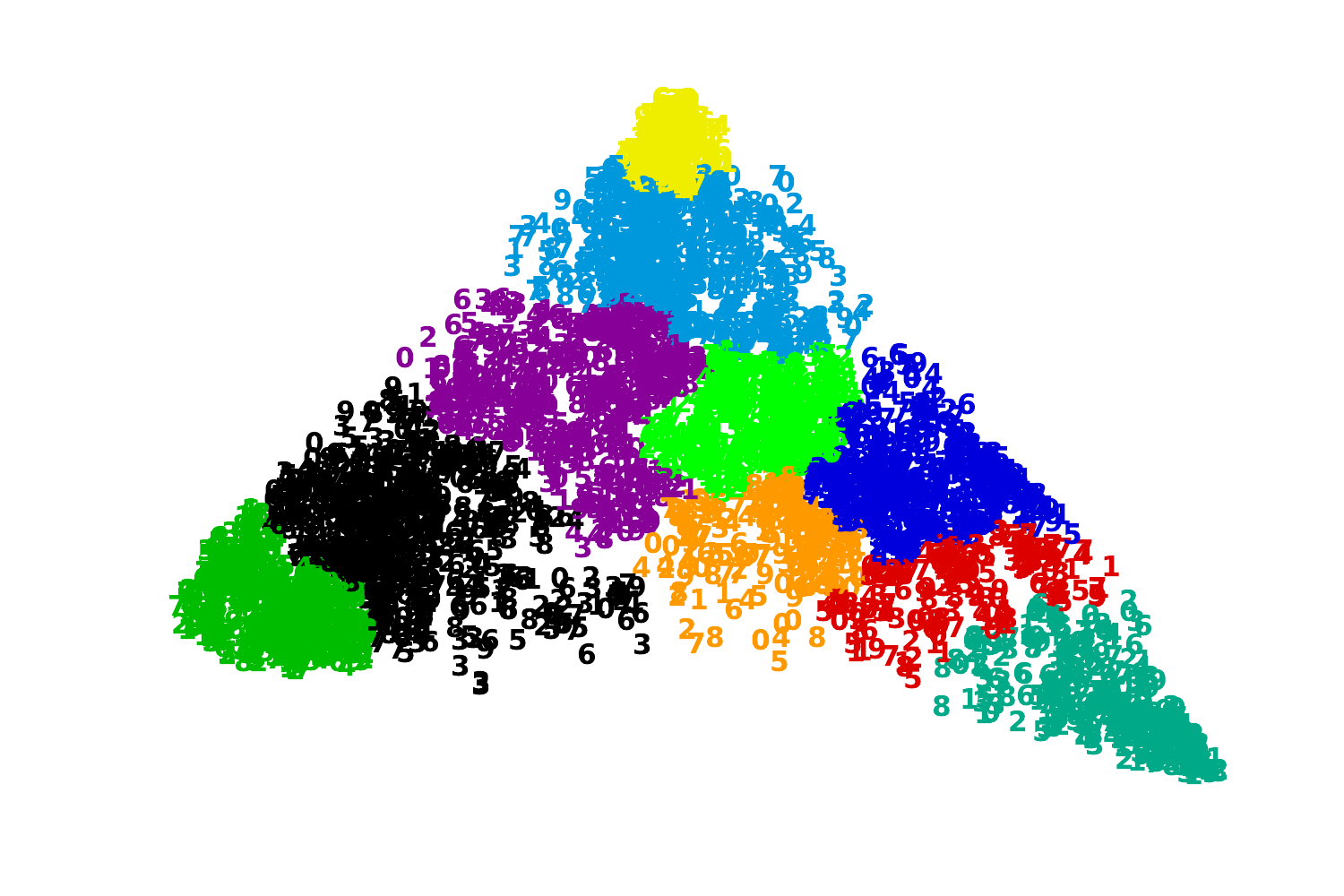}
\end{minipage}
}
\subfloat[$\mathcal{M}(D')$=3,000]{
\label{fig:improved_subfig_b}
\begin{minipage}[t]{0.49\textwidth}
\centering
\includegraphics[width=2.9in,height=2.31in]{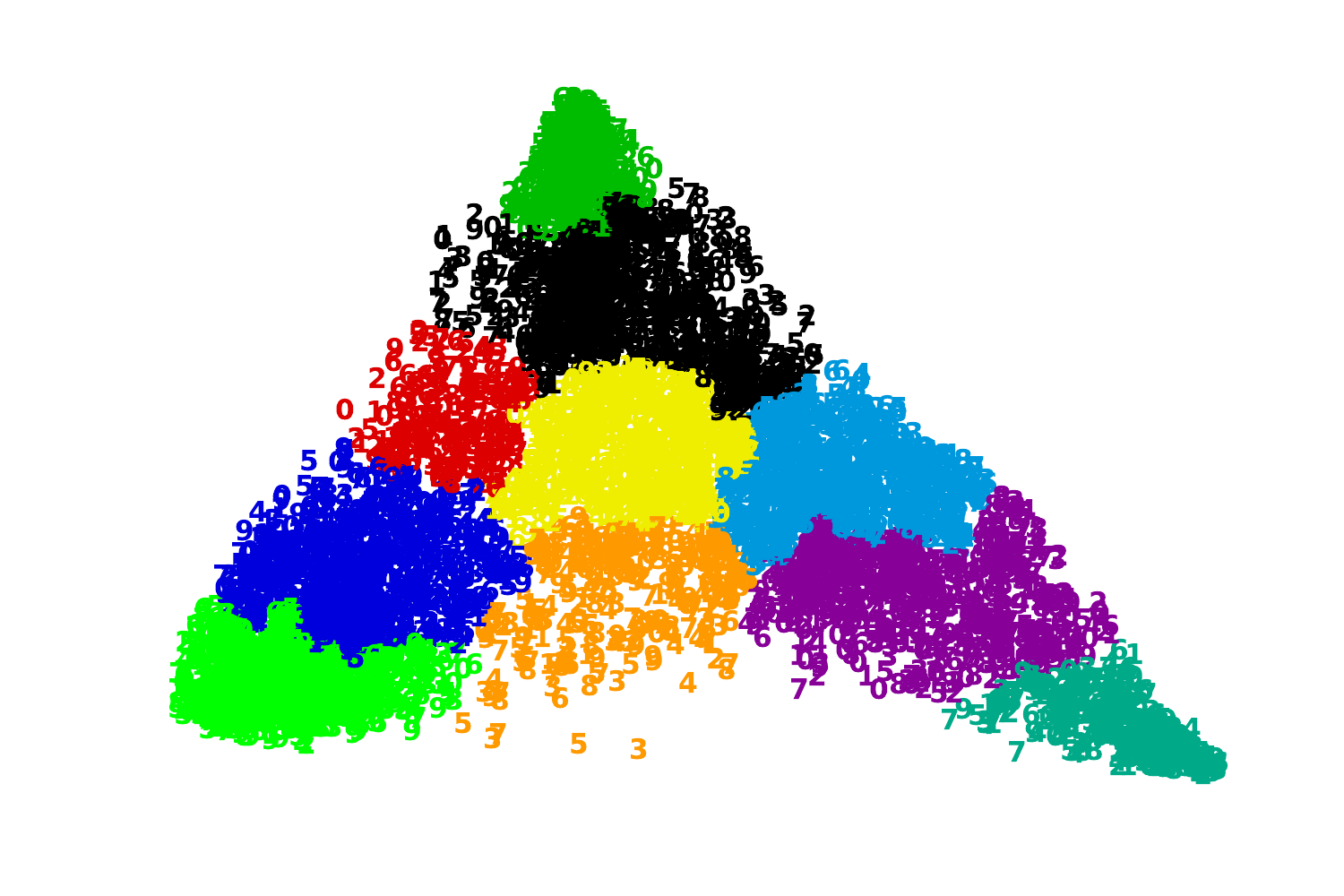}
\end{minipage}
}

\caption{2D embeddings of distribution matching-based machine teaching  sets on \emph{MNIST}, which properly draws the 10 classes.} 
\end{figure*}

\subsection{Minimizing $\|R(\theta,D)-R(\theta^*,D) \|$ with Quantitative  $\mathcal{M}(D)$}
Optimizing  Eqs.~(1) and (2) with a quantitative  $\mathcal{M}(D)$ is also a possible condition to simplify the minimization process. Then,  Eq.~(2) can be solved by an  unsupervised way. Therefore,    unsupervised machine learning algorithms such as clustering  can be deemed  as a special class of  candidate teaching methods with quantitative  $\mathcal{M}(D)$. 

\par In this group of experiments, we collect the learning risk change of $\|R(\theta,D)-R(\theta^*,D) \|$  with  quantitative  $\mathcal{M}(D)$ settings.
The compared three unsupervised algorithms are $k$-medoids, hierarchical, and spectral clustering, where $\mathcal{M}(D)$ is set as the clustering numbers.  Specifically, 
  kernel function of spectral clustering is set as  RBF, driving a kernel parameter as 0.1 to construct an affinity matrix, where a $k$-means clustering is used  to  assign labels in the embedding space of the kernel. The whole collected teaching results of the three baselines are drawn in Figure 6. We intuitively find the performance of all the clustering algorithms are very unstable.  They show sensitive  change on $\|R(\theta,D)-R(\theta^*,D) \|$ in term of the  test  set of \emph{Adult} and \emph{Phishing} since they are binary classification data sets without strong clustering structures. We also find Hierarchical clustering algorithm cannot decrease $\|R(\theta,D)-R(\theta^*,D) \|$ when setting $\mathcal{M}(D)$ be lower than 700. This is because that the two data sets have no intuitive tree structures.
 For the \emph{Satimage} and \emph{MNIST} data sets with clear clustering structure, unsupervised machine learning algorithms achieve better performance on 
minimizing $\|R(\theta,D)-R(\theta^*,D) \|$, even better than machine teaching on \emph{Satimage}.

Overall, the unsupervised machine learning approaches can be applied in  teaching a black-box learner, but show very unstable performance on minimizing learning risk due to their local convergence conditions.
A global strategy should be considered to minimize $\|R(\theta,D)-R(\theta^*,D) \|$ with quantitative  $\mathcal{M}(D)$.
This also is the inherent reason why our proposed distribution-based   algorithm can be adopted in    machine teaching with inestimable teaching risk.

To visualize the distribution of the output teaching set of distribution matching-based machine teaching algorithm, 
Figure 7 presents the 2D embeddings of   teaching sets of distribution matching-based machine teaching on \emph{MNIST} with different $\mathcal{M}(D')$. The results show those teaching sets can  properly draw the 10 separable classes.

\subsection{Teaching a Deep   Neural Network}
We compare the deep learning performance of our distribution matching-based machine teaching algorithm to the supervised and unsupervised machine learning models. Figure~8 presents the learning curves of regulating $\mathcal{M}(D)$ to minimize $\|R(\theta,D)-R(\theta^*,D) \|$ following the experiments of  Sections V.A and V.B. The deep neural network is ResNet20 and the tested datasets are CIFAR10 and CIFAR100. The hyperparameters of the network architecture
 are batch size=32, epochs=200, depth =20, learning rate=0.001, filter number=16, etc. The network architecture was implemented by Keras 2.2.3.
 The results show our machine teaching algorithm can still minimize $R(\theta^*,D)-R(\theta,D)$  faster than the compared supervised and unsupervised machine learning baselines, where $R(\theta^*,D)=0.9200$ over \emph{CIFAR10} and  $R(\theta^*,D)=0.6729$ over \emph{CIFAR100}. Figure 9 presents the 2D embeddings of  distribution matching-based machine teaching sets on \emph{CIFAR10} with different $\mathcal{M}(D')$.

\begin{figure}[!t]
\subfloat[ \emph{CIFAR10}  ]{
\label{fig:improved_subfig_b}
\begin{minipage}[t]{0.49\textwidth}
\centering
\includegraphics[width=2.9in,height=2.31in]{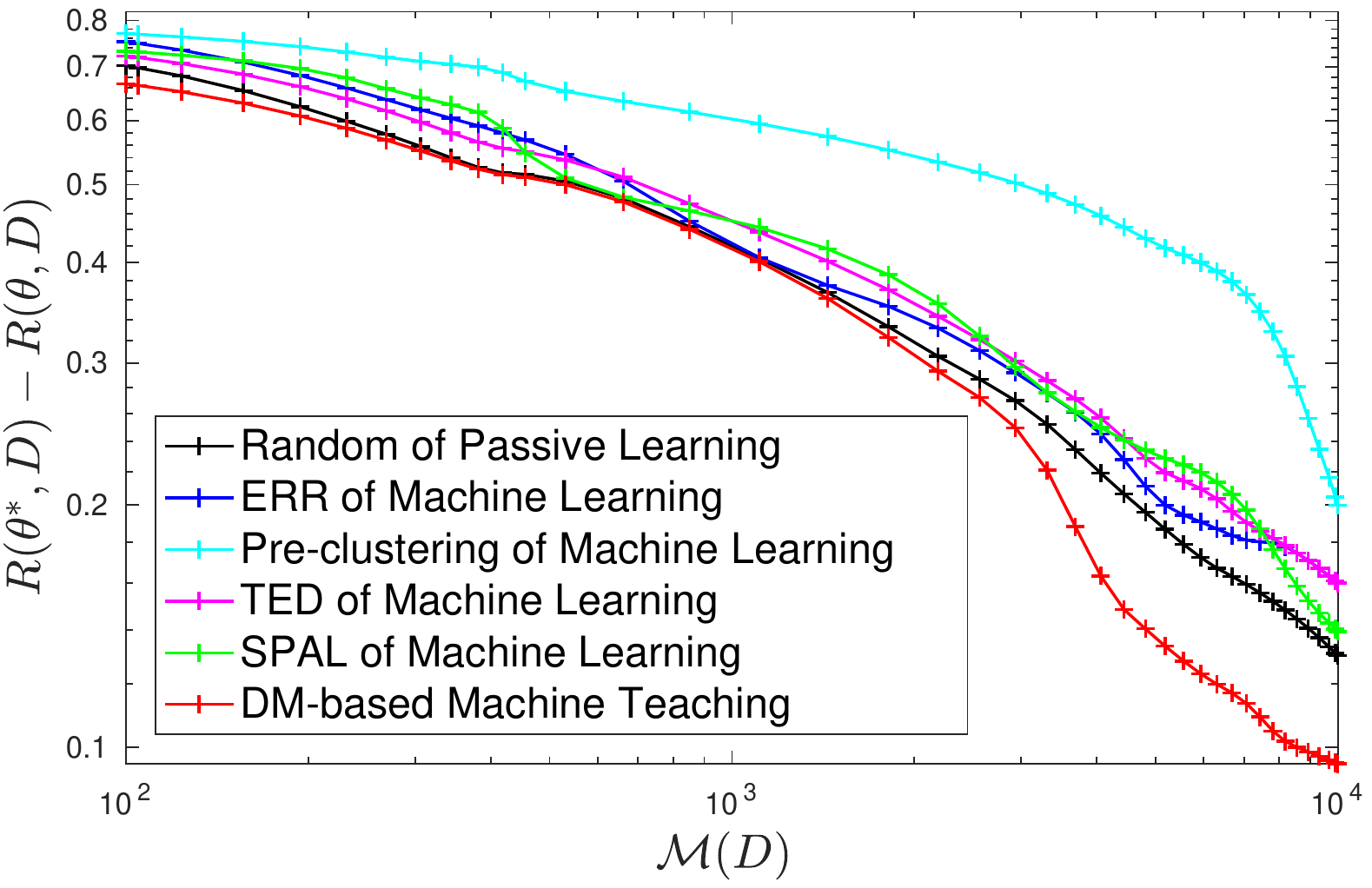}
\end{minipage}
}
\subfloat[\emph{CIFAR100}]{
\label{fig:improved_subfig_b}
\begin{minipage}[t]{0.49\textwidth}
\centering
\includegraphics[width=2.9in,height=2.31in]{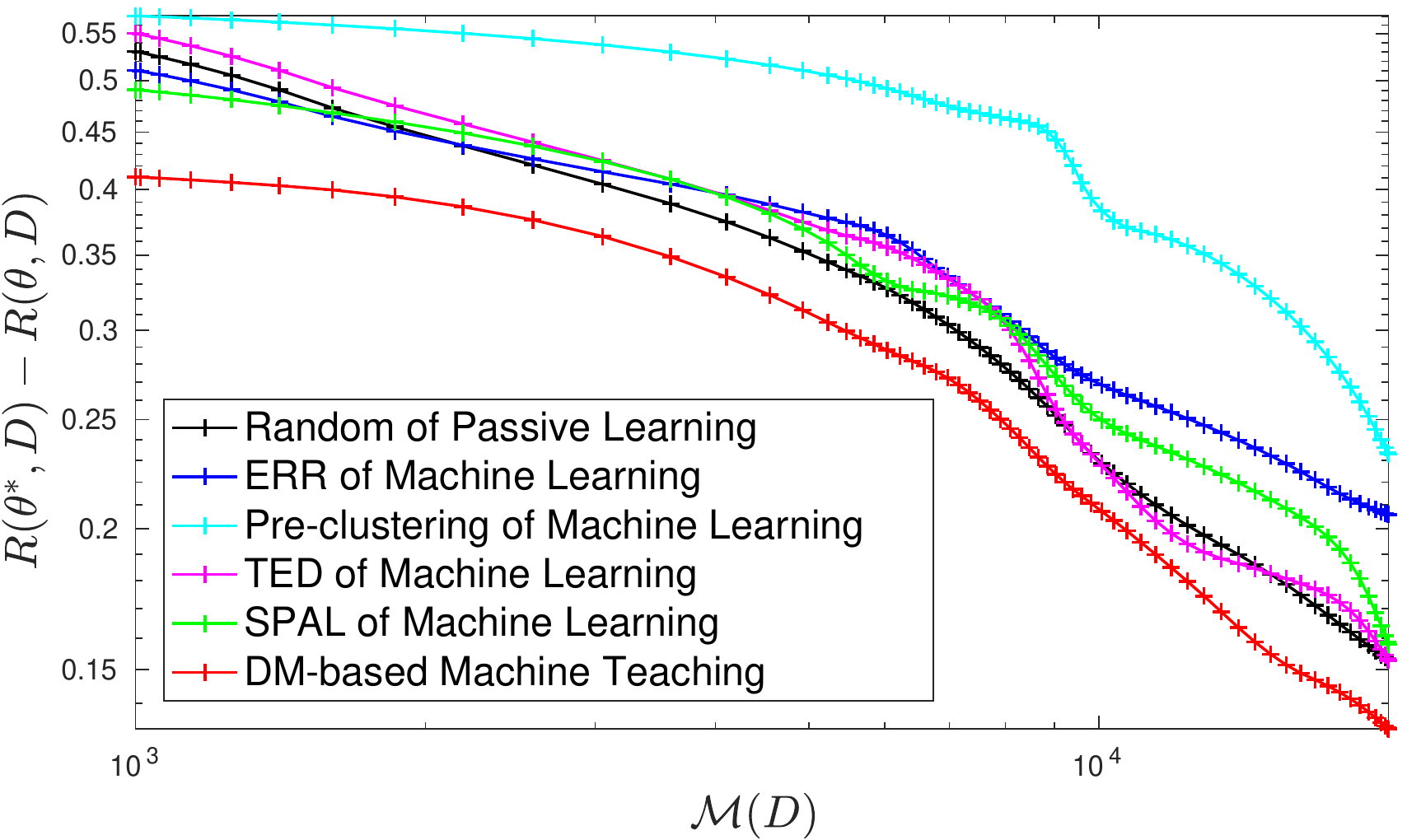}
\end{minipage}
}
\caption{{Regulating $\mathcal{M}(D)$ to minimize $\|R(\theta,D)-R(\theta^*,D) \|$ on data sets \emph{CIFAR10} and \emph{CIFAR100}.  } }
\end{figure}

\begin{figure}[!t]
\subfloat[ $\mathcal{M}(D')$=500  ]{
\label{fig:improved_subfig_b}
\begin{minipage}[t]{0.49\textwidth}
\centering
\includegraphics[width=2.9in,height=2.31in]{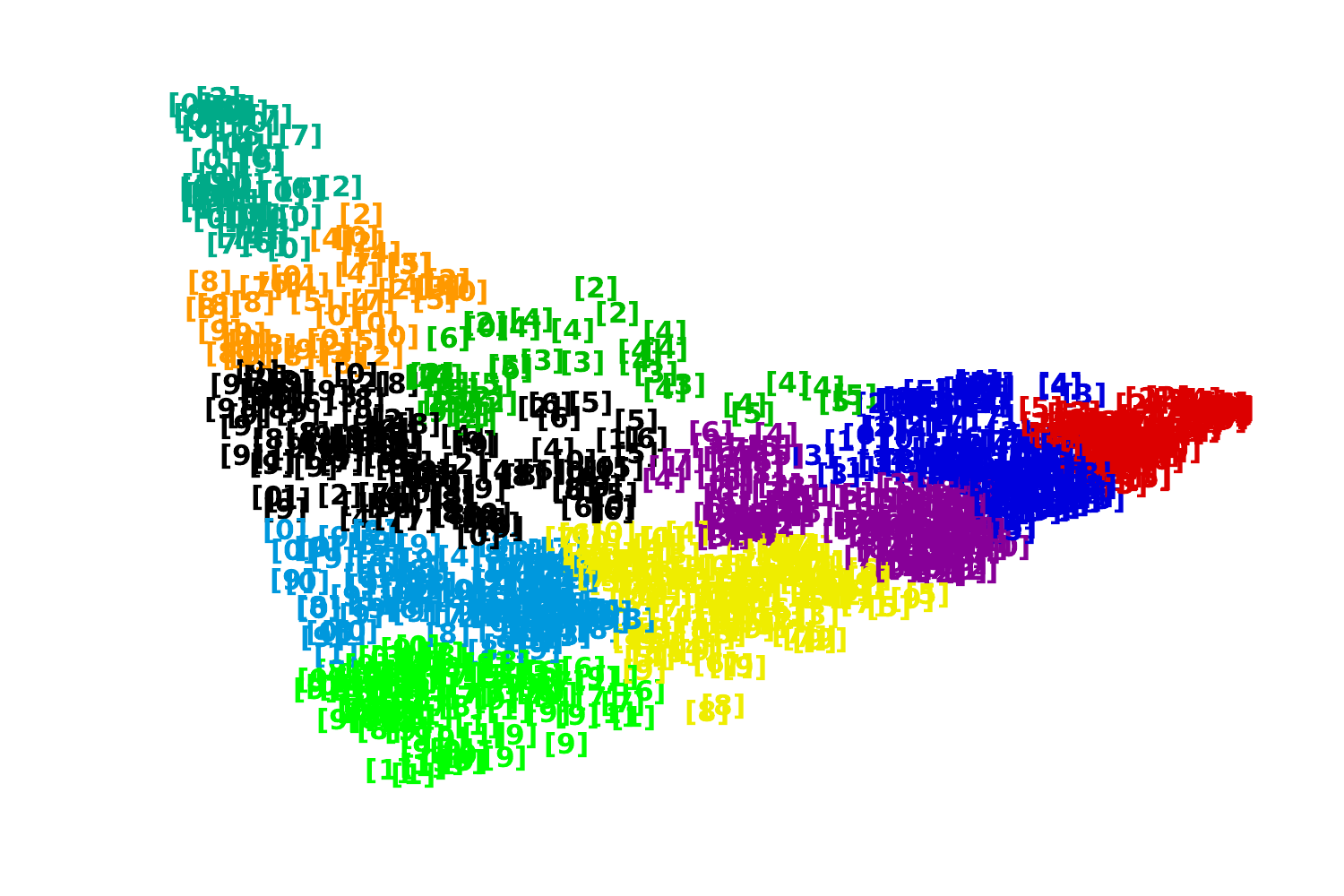}
\end{minipage}
}
\subfloat[$\mathcal{M}(D')$=1000]{
\label{fig:improved_subfig_b}
\begin{minipage}[t]{0.49\textwidth}
\centering
\includegraphics[width=2.9in,height=2.31in]{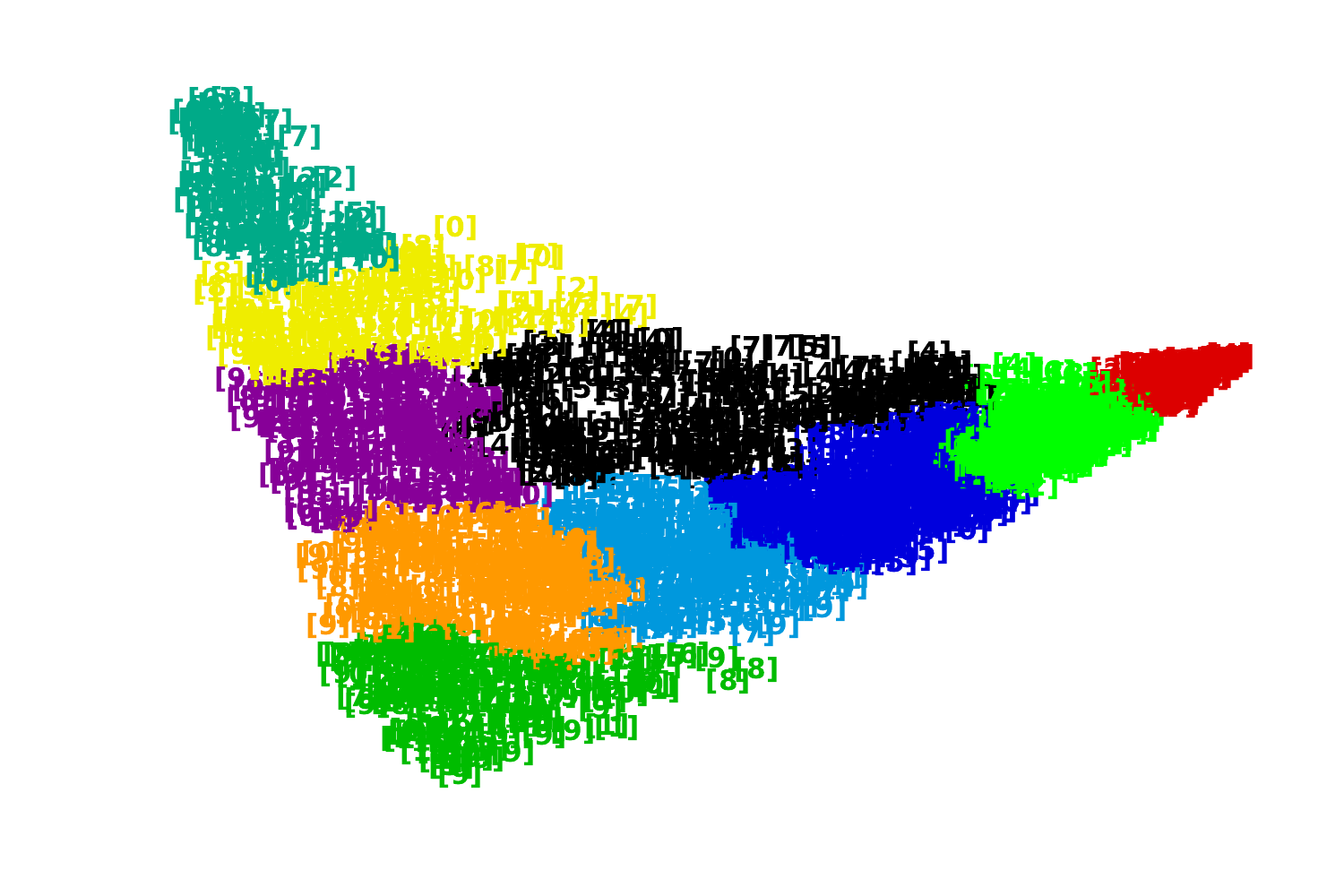}
\end{minipage}
}

\caption{{2D embeddings of distribution matching-based machine teaching  sets on \emph{CIFAR10}, which properly draws the 10 classes.  } }
\end{figure}

\begin{figure*}[!t]
\subfloat[KNR on \emph{digit}]{
\label{fig:improved_subfig_b}
\begin{minipage}[t]{0.32\textwidth}
\centering
\includegraphics[width=1.90in,height=1.51in]{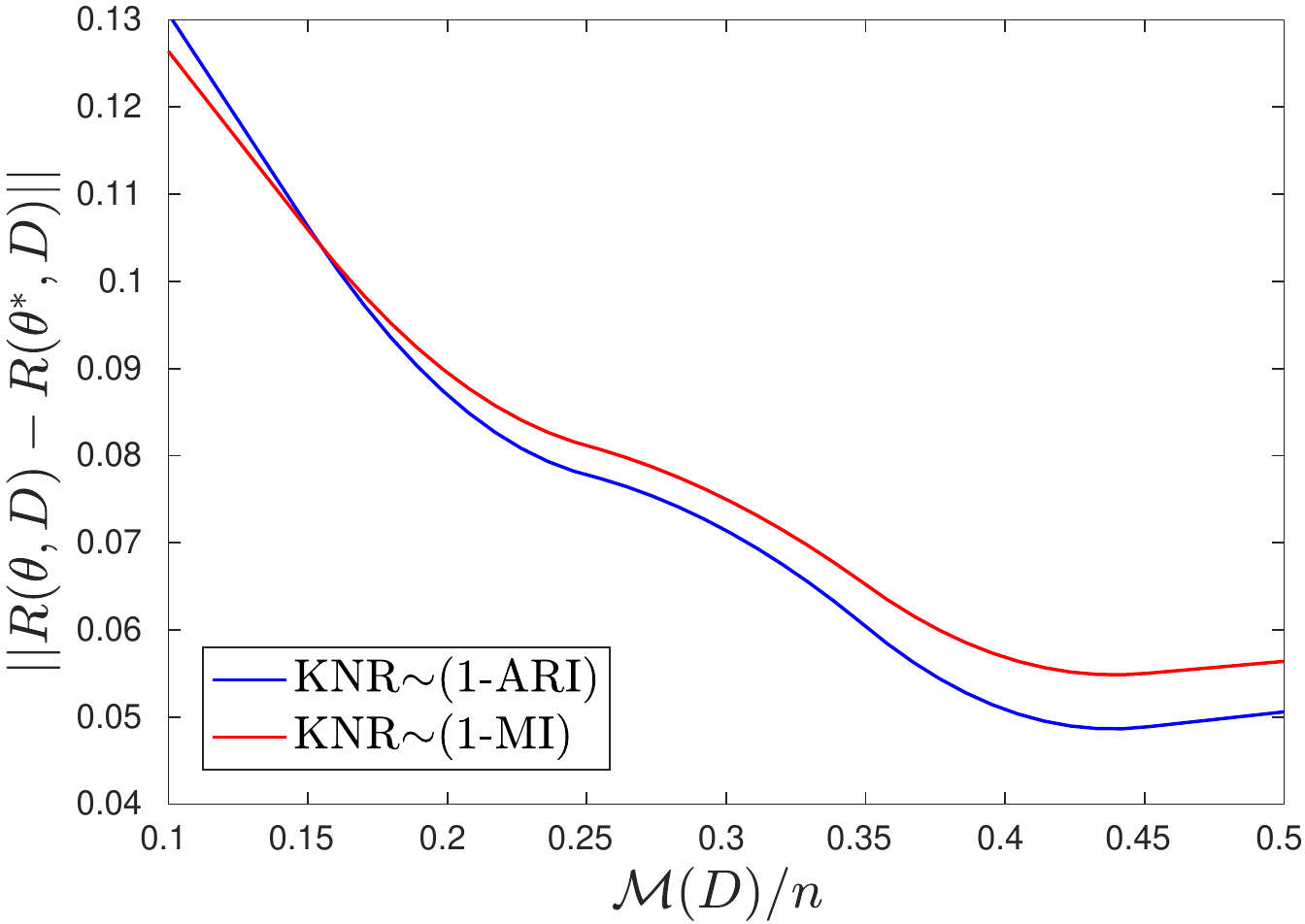}
\end{minipage}
}
\subfloat[ RandomForest on \emph{digit}  ]{
\label{fig:improved_subfig_b}
\begin{minipage}[t]{0.32\textwidth}
\centering
\includegraphics[width=1.90in,height=1.51in]{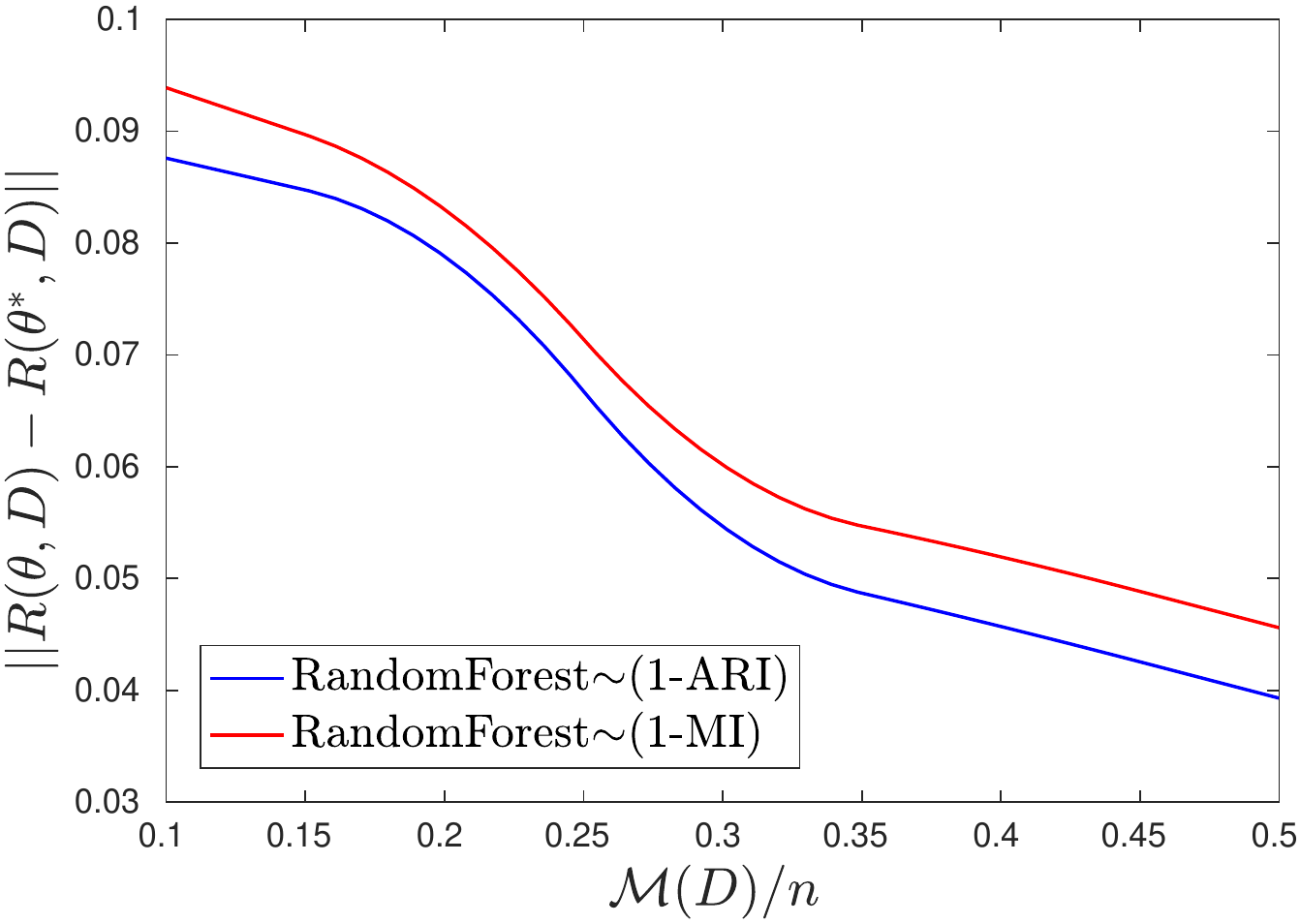}
\end{minipage}
}
\subfloat[MLPClassifier on \emph{digit}]{
\label{fig:improved_subfig_b}
\begin{minipage}[t]{0.32\textwidth}
\centering
\includegraphics[width=1.90in,height=1.51in]{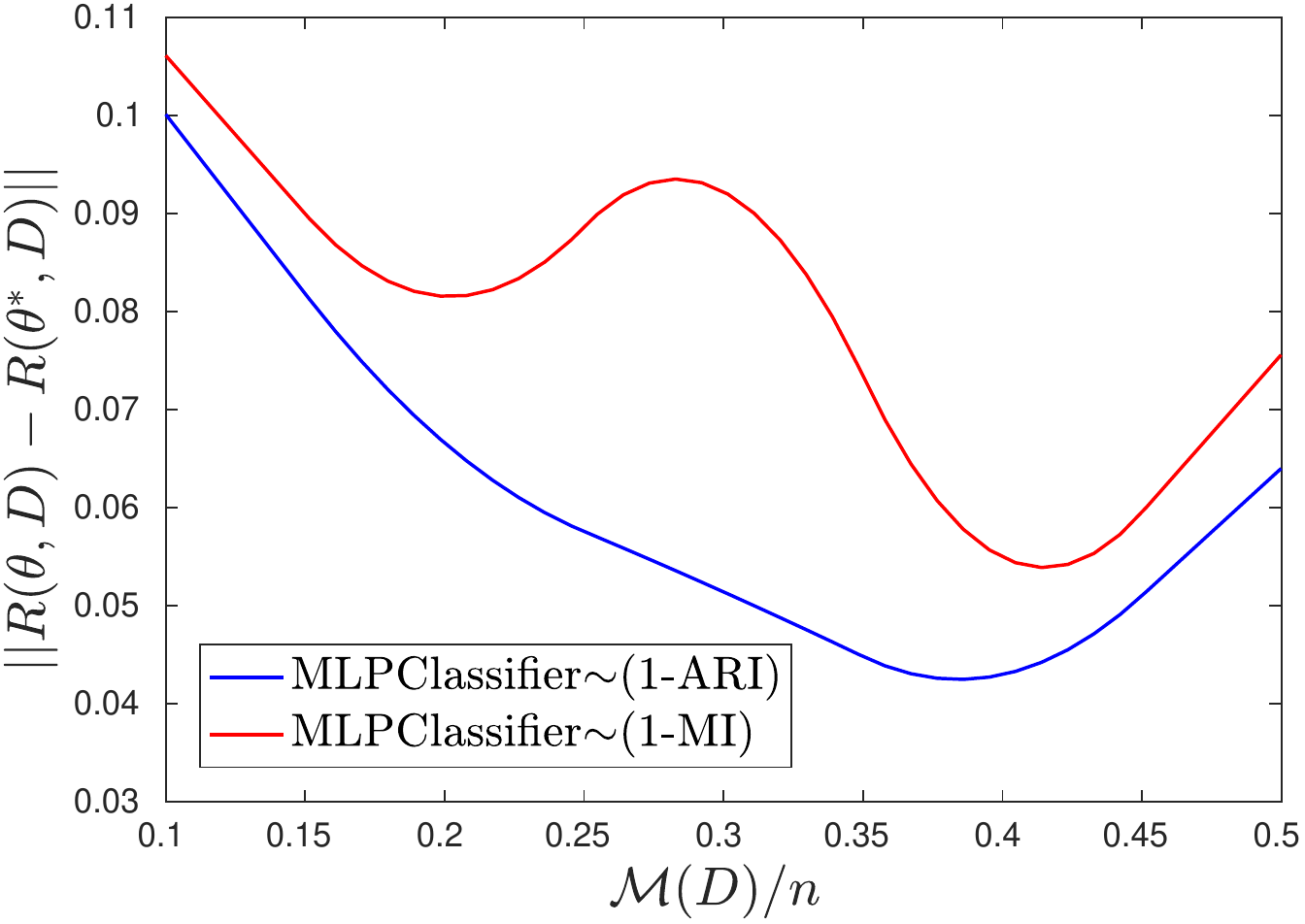}
\end{minipage}
}\\
\subfloat[SVM on \emph{digit}]{
\label{fig:improved_subfig_b}
\begin{minipage}[t]{0.32\textwidth}
\centering
\includegraphics[width=1.90in,height=1.51in]{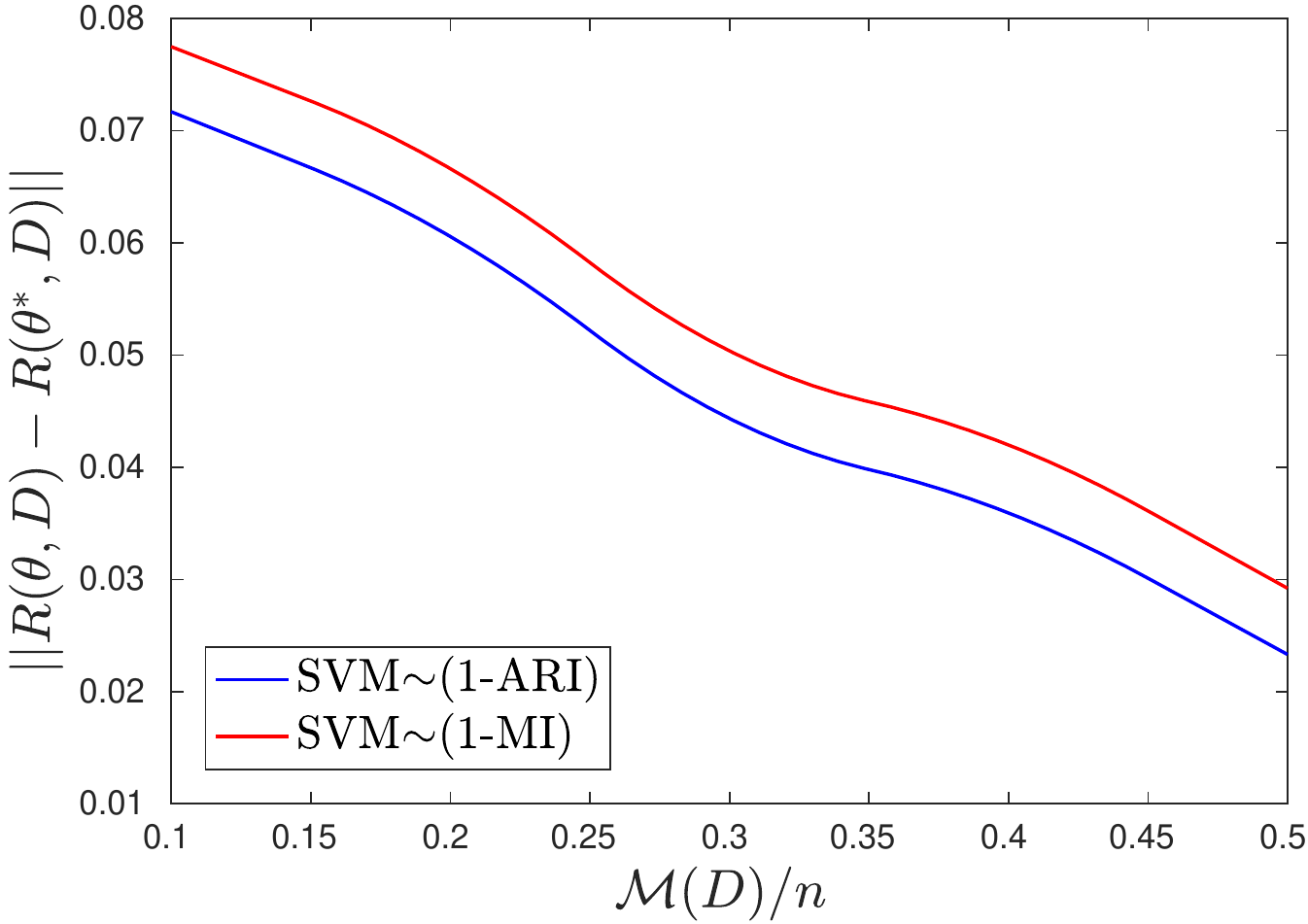}
\end{minipage}
}
\subfloat[KNR on \emph{USPS}]{
\label{fig:improved_subfig_b}
\begin{minipage}[t]{0.32\textwidth}
\centering
\includegraphics[width=1.90in,height=1.51in]{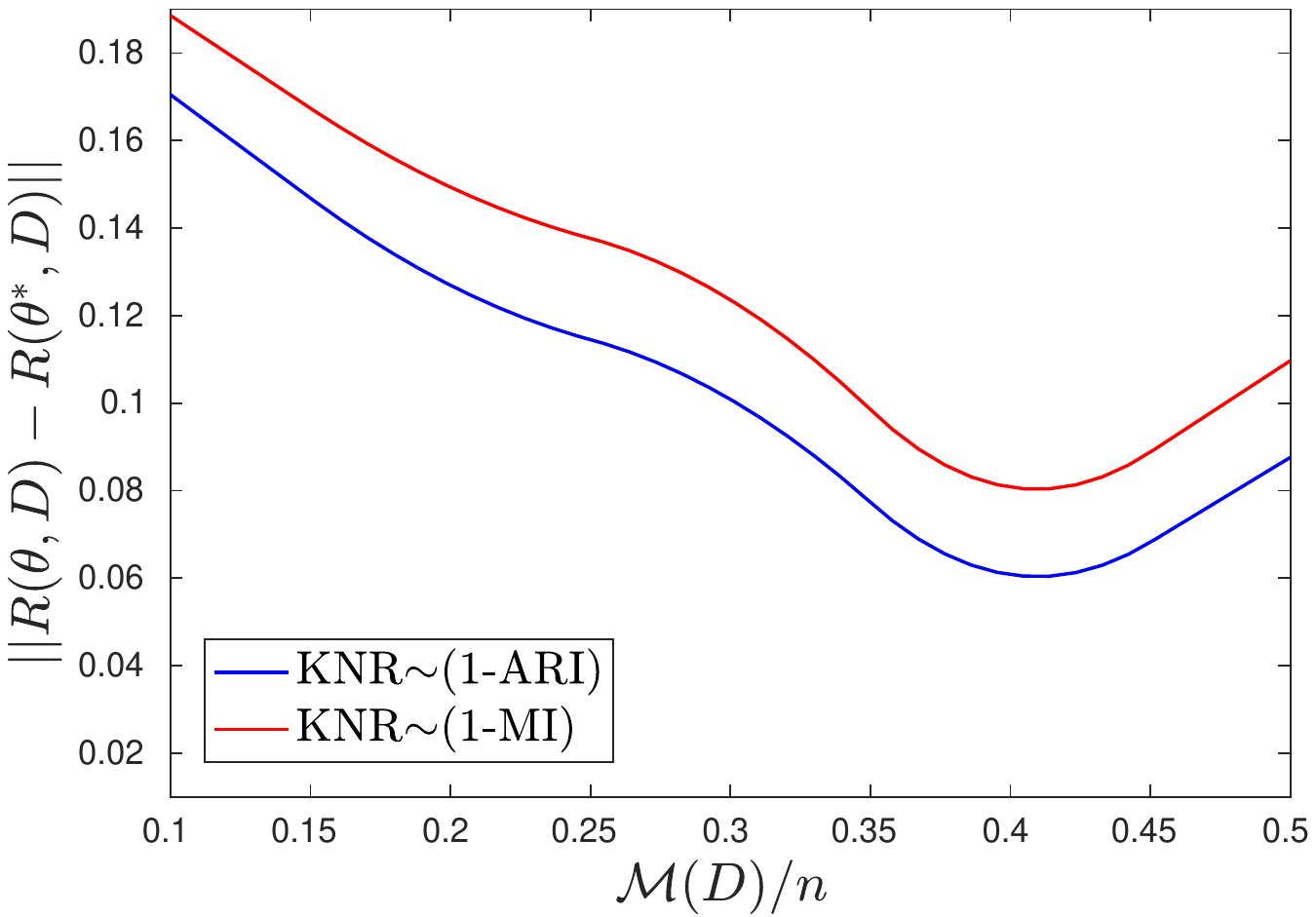}
\end{minipage}
}
\subfloat[ RandomForest on \emph{USPS} ]{
\label{fig:improved_subfig_b}
\begin{minipage}[t]{0.32\textwidth}
\centering
\includegraphics[width=1.90in,height=1.51in]{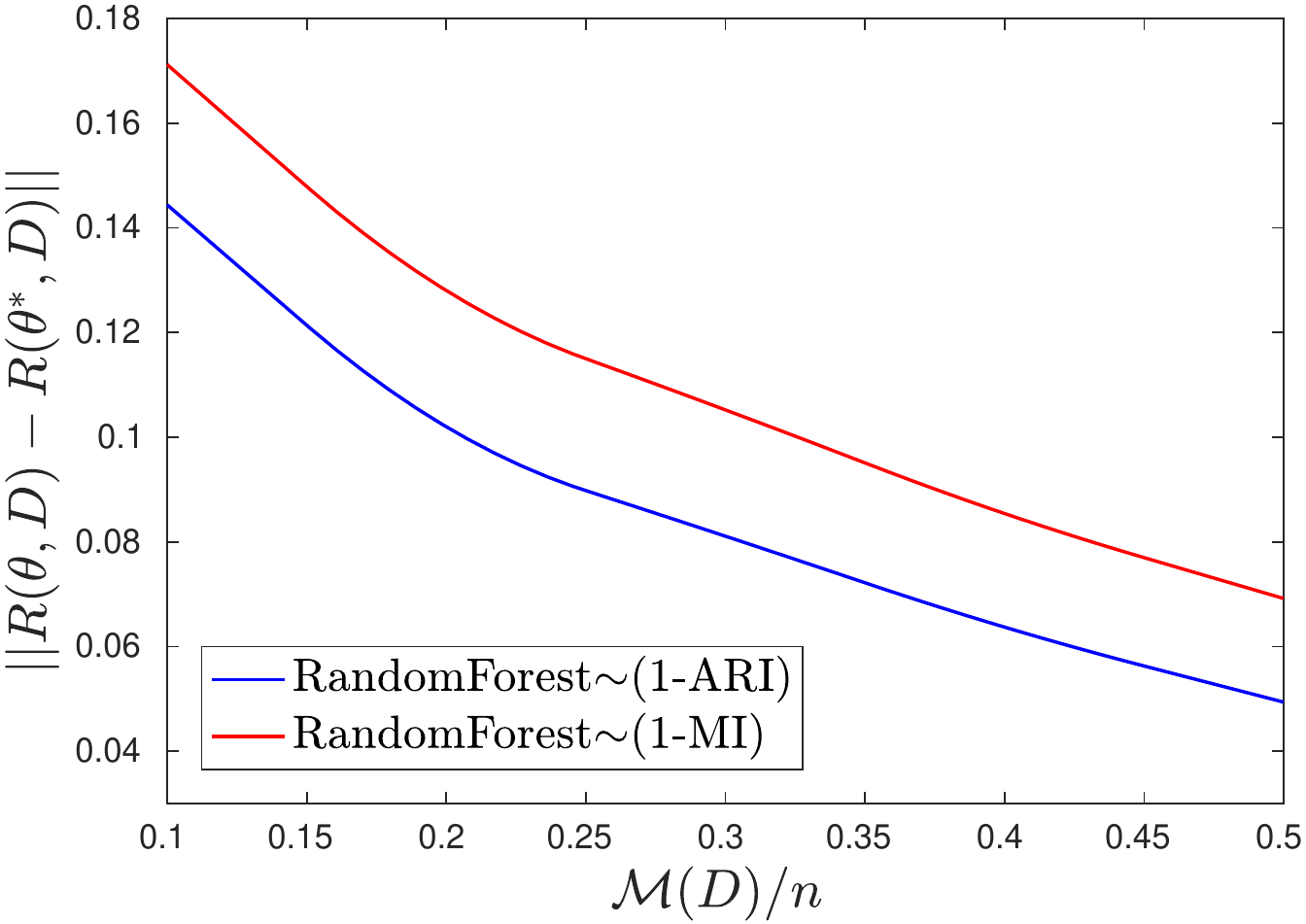}
\end{minipage}
}\\
\subfloat[MLPClassifier    on \emph{USPS}]{
\label{fig:improved_subfig_b}
\begin{minipage}[t]{0.32\textwidth}
\centering
\includegraphics[width=1.90in,height=1.51in]{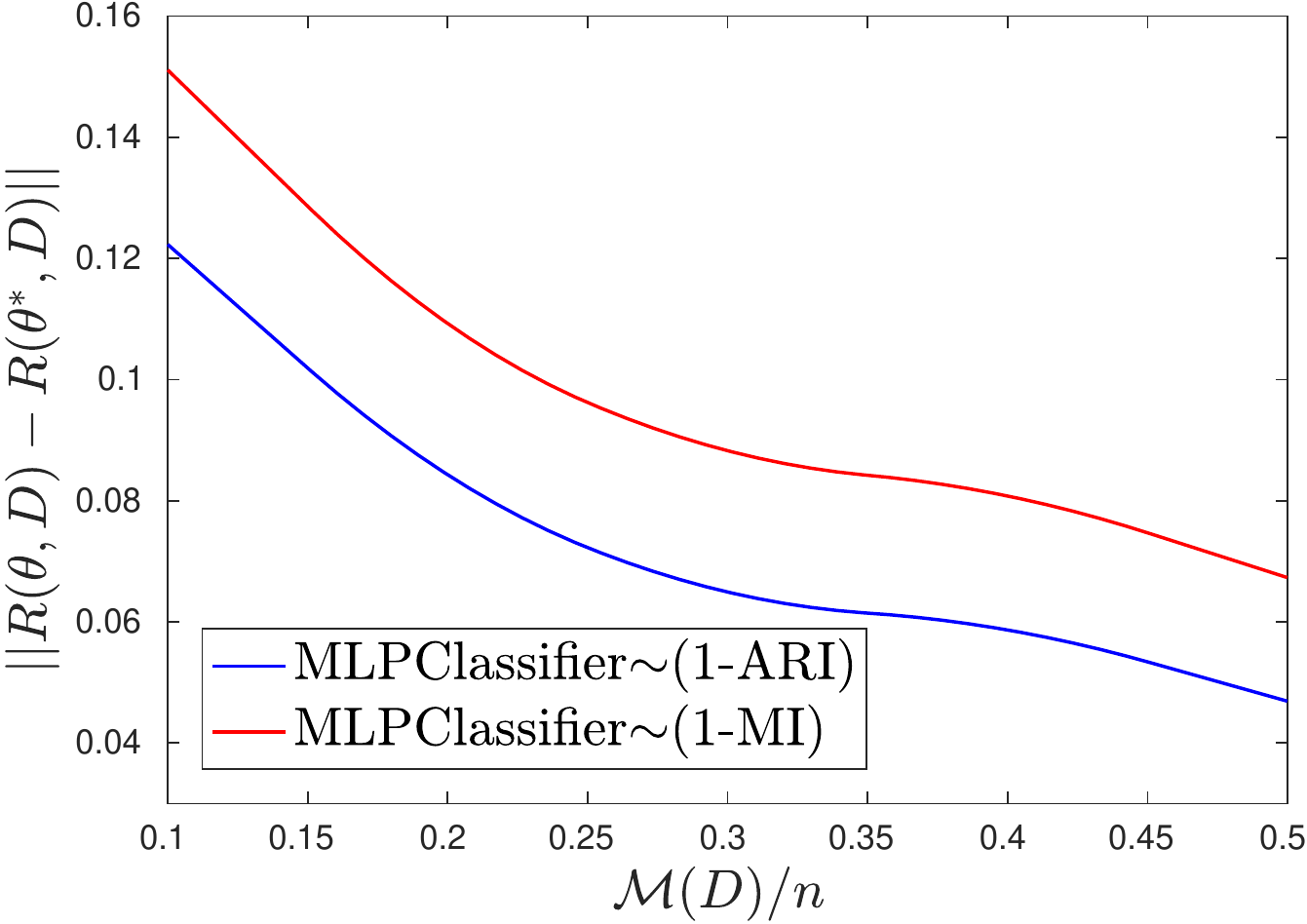}
\end{minipage}
}
\subfloat[ SVM on \emph{USPS}]{
\label{fig:improved_subfig_b}
\begin{minipage}[t]{0.32\textwidth}
\centering
\includegraphics[width=1.90in,height=1.51in]{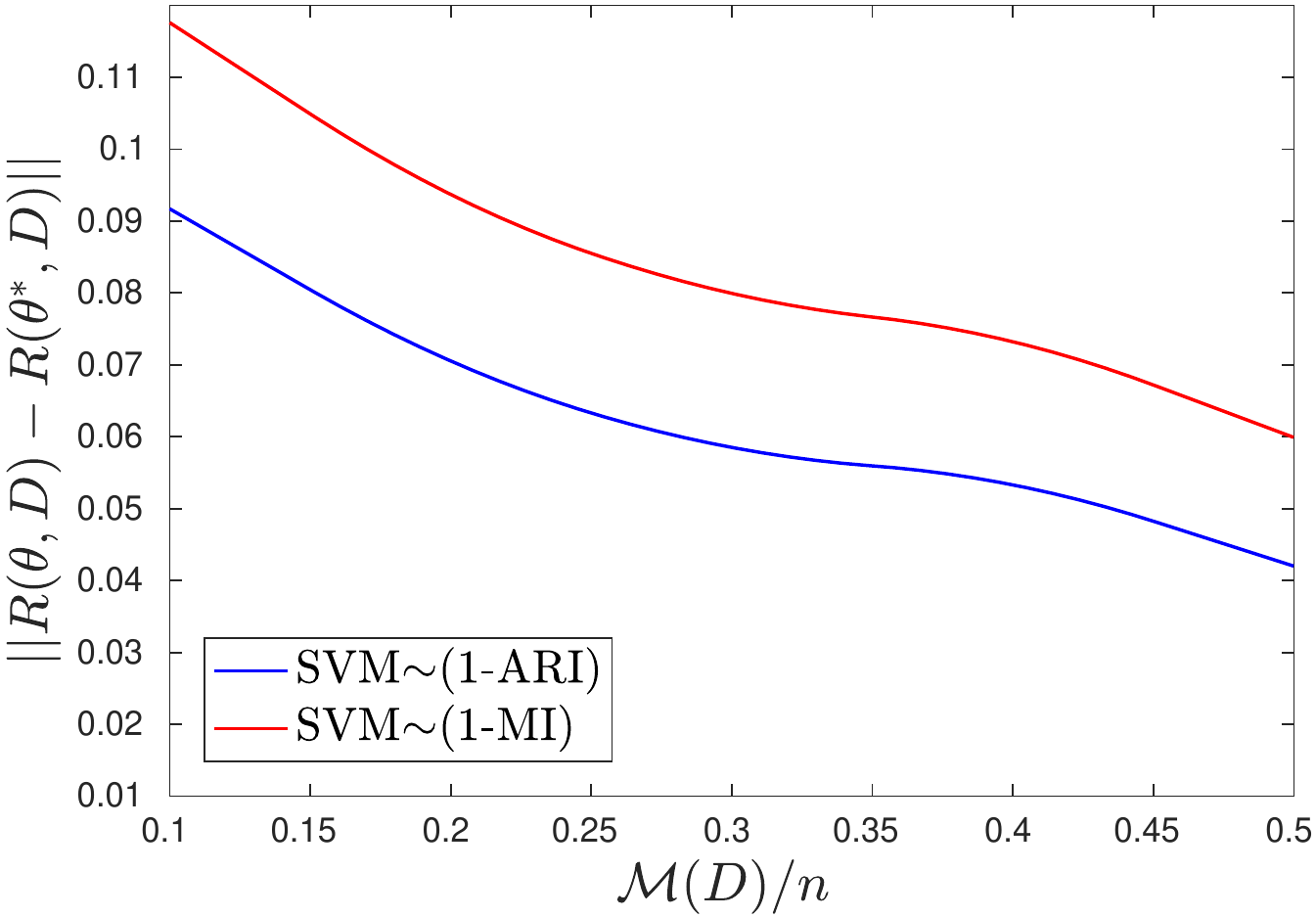}
\end{minipage}
}
\subfloat[KNR on \emph{FashionMnist}]{
\label{fig:improved_subfig_b}
\begin{minipage}[t]{0.32\textwidth}
\centering
\includegraphics[width=1.90in,height=1.51in]{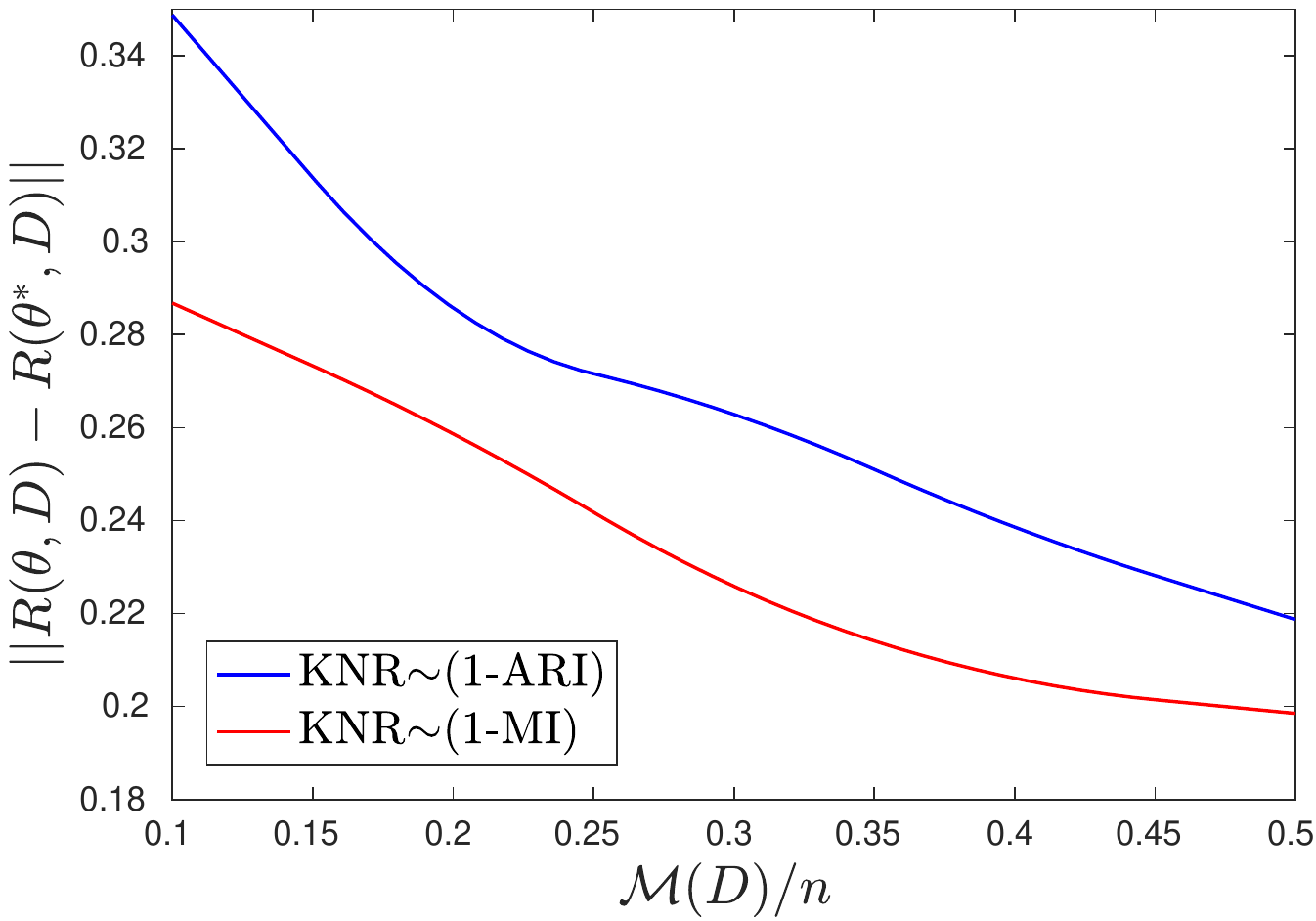}
\end{minipage}
}\\
\subfloat[RandomForest on \emph{FashionMnist} ]{
\label{fig:improved_subfig_b}
\begin{minipage}[t]{0.32\textwidth}
\centering
\includegraphics[width=1.90in,height=1.51in]{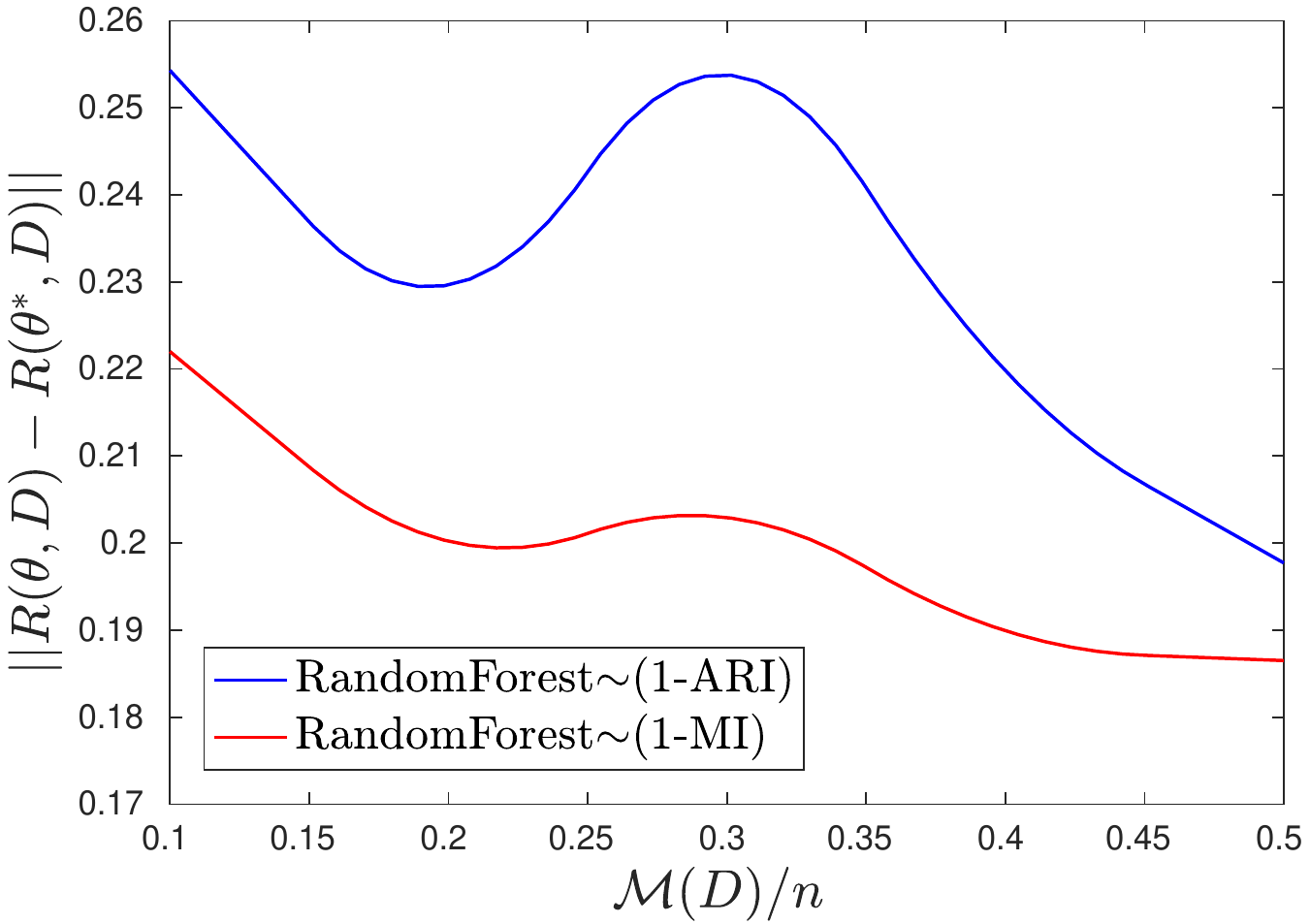}
\end{minipage}
}
\subfloat[MLPClassifier on \emph{FashionMnist}]{
\label{fig:improved_subfig_b}
\begin{minipage}[t]{0.32\textwidth}
\centering
\includegraphics[width=1.90in,height=1.51in]{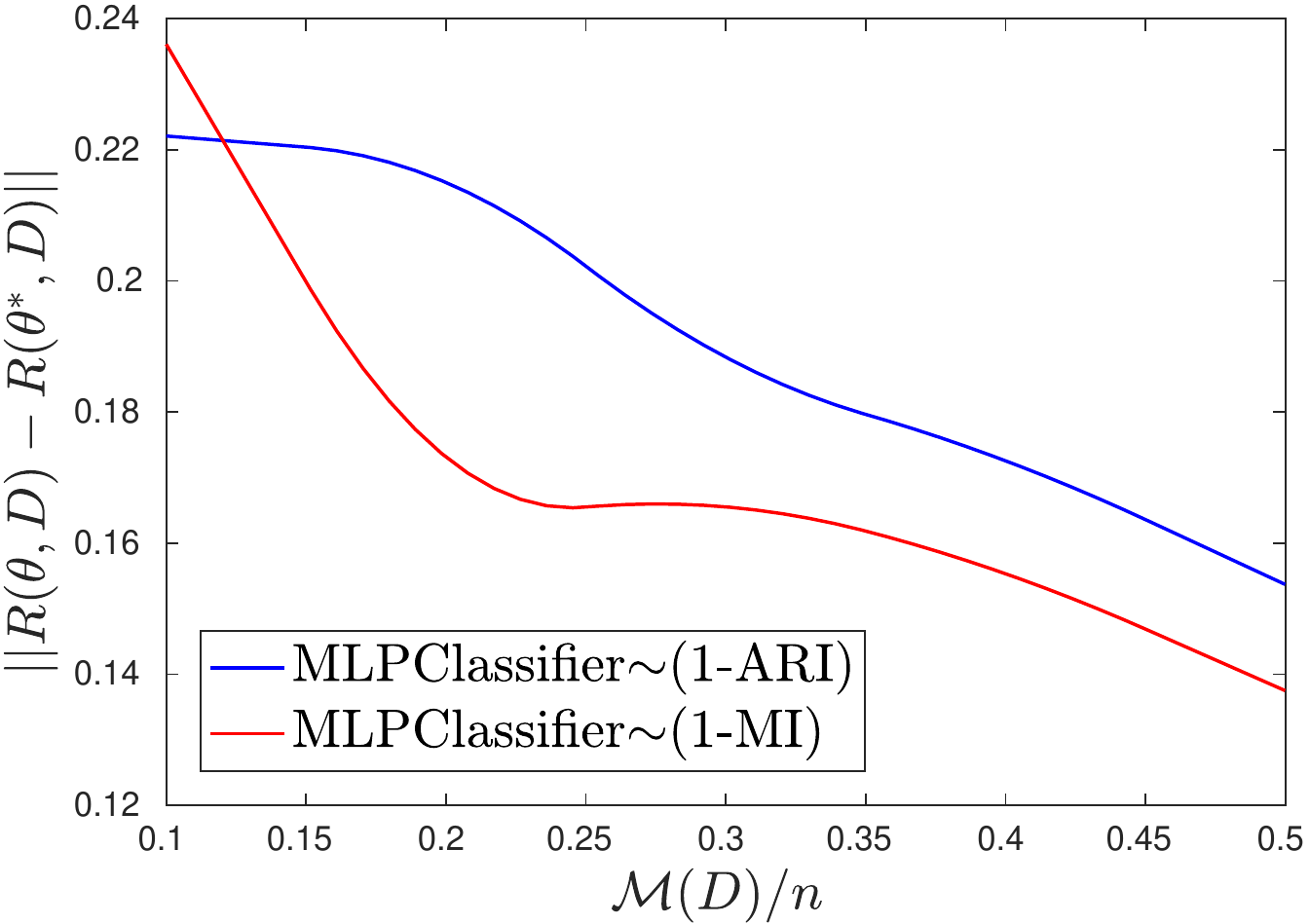}
\end{minipage}
}
\subfloat[SVM on \emph{FashionMnist}]{
\label{fig:improved_subfig_b}
\begin{minipage}[t]{0.32\textwidth}
\centering
\includegraphics[width=1.90in,height=1.51in]{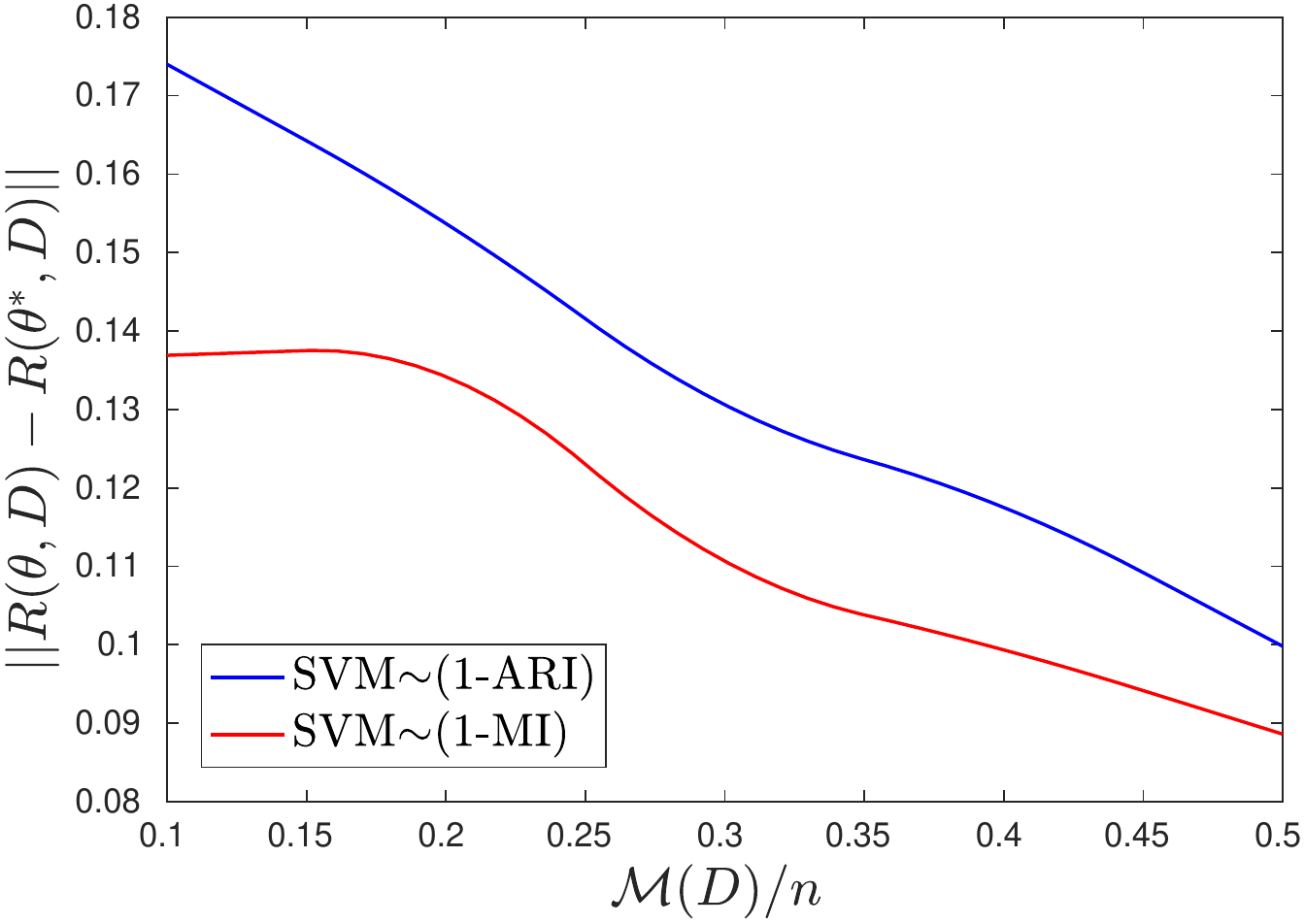}
\end{minipage}
}
\caption{ {Learning curves of regulating   $\mathcal{M}(\mathcal{D})$ to minimize $R(\theta^*,D)-R(\theta,D)$  on \emph{digit, USPS,} and {FashionMnist} data sets, where $\theta^*$ is respect to $h^*$ generalized from different classifiers.} } 
\end{figure*}

\section{Discussion on Our Assumption}
{This study is based on Assumption~3, which stipulates $R(\theta^*,D)\approx R(h^*,D)$. Therefore, how to derive a more general hypothesis $h$ may have perturbations to this assumption. In read-world, $h$  is usually generalized from different classifiers. In this section, we collect different classifiers to test the main technical steps of the iterative halving in distribution matching-based machine teaching.}

\par  {With this goal, we apply our machine teaching algorithm to derive  teaching sets   as the labeled data for the subsequent supervised classification. The candidate classifiers   are  $k$ Neighbors Regressor (KNR), Random Forest (RandomForest), Multi-layer Perceptron classifier (MLPClassifier), and 	Support Vector Machine (SVM). Figure~10 presents  the learning curves of regulating   $\mathcal{M}(D)$ to minimize $\|R(\theta,D)-R(\theta^*,D)\|$  on \emph{digit, USPS,} and {FashionMnist} data sets, where $\theta^*$ is respect to $h^*$ generalized from different classifiers.}

\par   The training parameters of the four classifiers are described as follows:  (1) we set   the $k$ nearest number as 10 for KNR; (2) we set the number  of trees in the RandomForest as 600; (3) for   MLPClassifier, we set the size of  the hidden layers as 100, the maximum iteration number as  1000, the $L_2$ penalty (regularization term) parameter as 0.0004, the optimization strategy as  stochastic gradient descent, and the learning rate as 0.001; (4) for the SVM classifier, we set the penalty parameter as 1.0, the kernel type as RBF, the degree of the   kernel function as 3, and the tolerance for stopping criterion as 0.003.

\par {Distribution matching-based machine teaching significantly  reduces the expected learning risks with the increase of $\mathcal{M}(D)$.
 In the  reported results of the four classifiers, perturbations of $\|R(\theta,D)-R(\theta^*,D)\|$,  i.e.  \[\Big(\max{R(\theta^*,D)-R(\theta,D)})-(\min{R(\theta^*,D)-R(\theta,D)}\Big),\] which yields an interval of [0.06, 0.14], where $\mathcal{M}(D)/n$ yields an interval of [0.1, 0.5]. Particularly, a part of learning curves don not keep consistent decreasing with the increase of $\mathcal{M}(D)/n$ such as MLPClassifier on \emph{digit}, RandomForest on \emph{FashionMnist}, etc. This explores that the iterative halving may delay the decrease of the expected learning risks, which further reduces the perturbations of having. Therefore, our machine teaching algorithm keeps an uniform decrease on expected learning risks, which also may  delay its decrease to a lower loss.
Besides this, SVM achieves the lowest $\|R(\theta,D)-R(\theta^*,D)\|$. Cooperating with a solid classifier may further delay the decrease of the learning risks.}

\section{Conclusion}
In this work, we proposed a distribution matching-based machine teaching algorithm with  regard to estimating  a teaching risk on distributions against a black-box learner.   The   analysis proved that the approximated  surrogate   had safety guarantee. Case study further presented support for this theoretical view  and demonstrated that Poincaré distance of  hyperbolic geometry could yield a smoother boundary for learning a surrogate  than Euclidean distance. We thus projected the subsequent iterative halving in this geometry. Experiments    demonstrated  distribution matching-based machine teaching outperformed
the  supervised and unsupervised machine learning algorithms   on minimizing expected learning risk disagreement. Finally, this work leads to an open question: can we co-teach the disagreement estimations on the distribution and model parameters?

\subsection*{Acknowledgments}
This work was supported  
 by Australian Research Council  under Grant   DP180100106 and DP200101328.

\section*{Appendix}
\textbf{Proof of Theorem 1.}
\vspace*{- 5pt}
\begin{proof}
Fix the input training data $D$, for any hypothesis $h_i \in \mathcal{H}$ over $D$, we use passive sampling to weight the learning process of the importance sampling algorithm. Let $ {\rm err}_T(h)$  denote  the important weighted error at $T$-time of sampling,  we define the risk of $T$ times of importance sampling as ${R(h, D, T):=} {\rm err}_T(h)=  \frac{1}{T}\sum_{t=1}^{T}\frac{q_t}{p_t}f(h(x_t), y_t)$, and the risk of full training on $D$ 
$R(h,D):={\rm err}_D(h)=\mathbb{E}_{(x,y)\sim{D}} f(h(x),y).$

\par Let $\mathcal{Y}$ be the label set of $D$, $f(\cdot)$ be a class of mapping function involved with error measure  from $D$ to $\mathcal{Y}$, such as the best-in-class error or all-in-class error. Given any sampled example  $x_t$ from $D$ which leads to a biased error at any $t$-th time of sampling, the upper bound of errors of the $T$ times of sampling satisfies $R(h,D,T)>R(h,D,t),  \forall  t<T$. That is to say, $R(h,D,T)>R(h,D,1), R(h,D,T)>R(h,D,2),$ $...,$ $R(h,D,T)>R(h,D,T\textit{-}1).$ Therefore,  the disagreement of the   surrogate and its full training data   in the $T$ times of sampling satisfies 
\begin{equation}
\begin{split}
\textbf{\rm R}&=|R(h, D, T)- R(h, D)| \leq R(h, D, T)\\
&=\frac{1}{T} \sum_{t=1}^{T}\frac{q_t}{\phi}f\left(h(x_t), y_t\right),\\
\end{split}
\end{equation}
then we have the following inequality about the upper bound of risk \textbf{\rm R}
\begin{equation}
\begin{split}
\textbf{\rm R} \leq \frac{1}{T} \sum_{t=1}^{T}\frac{q_t}{\phi}f\left(h(x_t), y_t\right),
\end{split}
\end{equation}
where $q_t<\phi$ for any $t\leq T$.To produce a   generalization on the    above  risk  disagreement, we need to estimate   a more general    \emph{upper bound} on  $\frac{1}{T} \sum_{t=1}^{T}\frac{q_t}{\phi}f\left(h(x_t), y_t\right)$. Following importance sampling,  we 
define  $f(\cdot)$ be a 0-1 loss. Then, a more general upper bound  of $\sum_{t=1}^{T}\frac{q_t}{\phi}f\left(h(x_t), y_t\right)$ is  $\sum_{t=1}^{T}\frac{q_t}{\phi}\times 1$. We here update  Eq. (21) into
\begin{equation}
\begin{split}
\textbf{\rm R} \leq  \frac{1}{\phi}.
\end{split}
\end{equation}

  Let $h_1, h_2,...,h_T$ be $T$  independent hypotheses over ${D}$ at different sampling times, where $h_t$ is the hypothesis generated at the $t$-time of sampling. We
   follow the result of Eq. (22) and know
\begin{equation}
\begin{split}
|R(h, D, T)- R(h, D)|\leq \frac{1}{\phi}, \forall t< T.
\end{split}
\end{equation}
Simply to say,  $\frac{1}{\phi}$ can be a hypothesis diameter that covers the hypothesis class of $\{h_1, h_2,...,h_T\}$. With this diameter constraint, for any pair hypotheses $\{h_t,h_{t'}\}$, we conclude 
\begin{equation}
\begin{split}
\Big|{\rm err}_T(h_{t'})-{\rm err}_{D}(h_t)\Big|\leq \frac{1}{\phi}, \forall t, t' < T.
\end{split}
\end{equation}
\par After applying the McDiarmid’s Inequality for Eq.~(24), with definition 2, the generalization probability bound of achieving a safety surrogate in the $T$ times of sampling is
\begin{equation}
\begin{split}
{\rm  Pr}\Bigg \{ \Big| |{\rm err}_T(h)-{\rm err}_D(h) |-  \mathbb{E} [{\rm err}_T(h)-{\rm err}_D(h) ] \Big | \leq  \lambda \Bigg \}  &\leq  {\rm exp} \left ({\frac{-2 \lambda^2}{\sum_{i=1}^T  \phi^{-2}}}  \right)\\
&= {\rm exp} \left ({\frac{-2 \lambda^2}{T  \phi^{-2}}}  \right).
\end{split}
\end{equation}
If we set $\lambda$ proportional to $\phi^{-1}$,  then there exists a maximum possible martingale value of  $\Big| |{\rm err}_T(h)-{\rm err}_D(h) |-   \mathbb{E} [{\rm err}_T(h)-{\rm err}_D(h) ] \Big |$. Following the choice of slack variable in \citep{beygelzimer2009importance}, we set  $\lambda=\sqrt{2({\rm In}\  d+{\rm In} ( 2/\delta)) } $ and $\hat{\textbf{\rm R}}=\mathbb{E}_{h\in \mathcal{H}} |R(h, D, T)- R(h, D)|$, then the bound  
 is as stated.
\end{proof}

\textbf{Proof of Theorem 2.}
\begin{proof}
Fix $\theta$, assume that ${D}$ is over a ball $B$ with radius $r$. Let $ h(x)\neq h^*(x)$,  $h^*=  \arg\inf\limits_{h\in \mathcal{H}}  {\rm err}_{{D}}(h),$ we define $\vartheta$ \citep{beygelzimer2010agnostic}  \citep{beygelzimer2008importance} as
\begin{equation}
\begin{split}
&\vartheta= \mathbb{E}_{x_t\in \mathcal{D}} {\rm sup}_{h\in B(h^*,r)} \left \{  \frac{ \ell(h(x_t),\mathcal{Y})-\ell(h^*(x_t),\mathcal{Y})   }{r}       \right\},
\end{split}
\end{equation}
where $\ell(h,\mathcal{Y})$ denotes the classification loss of a hypothesis $h$ regarded with the label space $\mathcal{Y}$, and $r$ denotes the hypothesis radius of $\mathcal{H}$.

To estimate the hypothesis radius $r$, we have $\ell(h,h^*)\leq r$. For any hypothesis $h$, we define the hypothesis distance $\ell(h,h^*)$ as:
$\ell(h,h^*)=\mathbb{E}_{x\in {D}} {\rm max}| \ell(h(x_t), \mathcal{Y})-\ell(h^*(x_t), \mathcal{Y})   | $. Then we have
\begin{equation}
\begin{split}
&\ell(h,h^*)=\mathbb{E}_{x\in {D}} {\rm max}| \ell(h(x_t), \mathcal{Y})-\ell(h^*(x_t), \mathcal{Y})   |\\
&\leq   \mathop{{\rm sup}}\limits_{x_t', x_t\in D'} \left |\frac{{\rm max} \ \ell(h(x_t), \mathcal{Y})-\ell(h(x_t'), \mathcal{Y}) }    {{\rm min} \ \ell(h(x_t), \mathcal{Y})-\ell(h(x_t'), \mathcal{Y}) } \right |  \times | \mathbb{E}_{x\in {D}} | \ell(h(x_t), \mathcal{Y})-\ell(h^*(x_t), \mathcal{Y})   |\\
&\leq K_f \left| \mathbb{E}_{x\in {D}}  \ell(h(x_t), \mathcal{Y}) \right| +\left | \mathbb{E}_{x\in {D}}   \ell(h^*(x_t), \mathcal{Y}) \right|\\
&\leq 2K_f \ell(h).\\
\end{split}
\end{equation}
By Theorem 2 of \citep{beygelzimer2008importance}, we know that the risk disagreement of $h$ and $h^*$ satisfies  $\ell(h)\leq \ell(h^*)+2\sqrt{\frac{8}{t-1} {\rm In} \left ( \frac{2(t^2-t) \frac{d}{2^{n-T}}}{\delta}\right )}$. Thus, 
\begin{equation}
 \begin{split}
 \ell(h,h^*)&   \leq 2K_f \ell(h)  \\
 &  \leq 
             2K_f \left(\ell(h^*)+2\sqrt{\frac{8}{t-1}{\rm In} \left( \frac{2(t^2-t) \frac{d}{2^{n-T}}}{\delta}\right ) }  \ \right).\\
\end{split}
\end{equation}
For any two hypotheses $h_i$, $h_j$, $\ell(h_i,h_j)\leq 2\ell(h,h^*)$ over $B(h^*, r)$.
 Therefore, updating the hypothesis $h$ into the optimal hypothesis $h^*$ over $D'$  costs at most $2\ell(h,h^*)$ at $t$-time. Here, we give a bound on $2\ell(h,h^*)$ as follows:
\begin{equation}
 \begin{split}
&2\ell(h,h^*) \\
&\leq 2\vartheta r  \leq 4\vartheta K_f \left ( \ell(h^*)+2\sqrt{\frac{8}{t-1}{\rm In}\left ( \frac{2(t^2-t) \frac{d}{2^{n-T}}}{\delta}\right ) }\ \right).\\
\end{split}
\end{equation}
When we associate  the loss function $\ell(h,\mathcal{Y})$ with   ${\rm err}_D'(h)$ as a generalization, $\ell(h^*)$ equals $R(h^*,D')$.  Theorem~2 then holds.

\end{proof}

\bibliographystyle{apacite}
\bibliography{example_paper}
\end{document}